\documentclass[10pt,journal,compsoc]{IEEEtran}
\ifCLASSINFOpdf
   \usepackage[pdftex]{graphicx}
\else
\fi
\usepackage{amsmath}
\usepackage[cmintegrals]{newtxmath}

\newtheorem{theorem}{Theorem}
\newtheorem{definition}{Definition}

\usepackage{bm}
\usepackage{algorithm}  
\usepackage{algorithmicx}  
\usepackage{algpseudocode}  
\usepackage{amsmath}  
\usepackage[table]{xcolor}
\usepackage{float}  
\usepackage{lipsum}
\makeatletter
\usepackage{array}
\usepackage{multirow}
\usepackage{longtable}
\usepackage{rotating}
\usepackage{stfloats}
\usepackage{graphicx}  
\ifCLASSOPTIONcompsoc
\usepackage[caption=false,font=normalsize,labelfon
t=sf,textfont=sf]{subfig}
\else
\usepackage[caption=false,font=footnotesize]{subfig}
\fi
\begin{document}
%
\title{Tensor Full Feature Measure and Its Nonconvex Relaxation Applications to Tensor Recovery}
%
%
%
%

\author{Hongbing Zhang,
	Xinyi Liu, Hongtao Fan, Yajing Li, Yinlin Ye
\IEEEcompsocitemizethanks{\IEEEcompsocthanksitem This work was supported by the National Natural Science Foundation of
	China (Nos. 11701456, 11801452), Fundamental Research Project of Natural Science in Shaanxi Province General Project (Youth) (Nos. 2019JQ-415, 2019JQ-196), the Initial Foundation for Scientific Research of Northwest A\&F University (Nos.2452017219, 2452018017), and Innovation and Entrepreneurship Training Program for College Students of Shaanxi Province (S201910712132).\protect\\
\IEEEcompsocthanksitem H. Zhang, X. Liu, H. Fan, Y. Li and Y. Ye are with the College of Science, Northwest A\&F University, Yangling, Shaanxi 712100, China(e-mail: zhanghb@nwafu.edu.cn; Lxy6x1@163.com; fanht17@nwafu.edu.cn; hliyajing@163.com; 13314910376@163.com).}
}

%
%

\markboth{ }%
{Shell \MakeLowercase{\textit{et al.}}: Bare Demo of IEEEtran.cls for Computer Society Journals}
%



\IEEEtitleabstractindextext{%
\begin{abstract}
Tensor sparse modeling, as a promising method, has achieved great success in the whole field of science and engineering. As is known to all, various data in practical application are often generated by multiple factors, so the use of tensors to represent the data containing the internal structure of multiple factors came into being. However, different from the matrix case, constructing reasonable sparse measure of tensor is a relatively difficult and very important task. In this paper, we propose a novel tensor sparsity measure called Tensor Full Feature Measure (FFM). It comprehensively delineates the sparse features of tensors by simultaneously integrating the feature information of each dimension of tensors and the relevant feature information between any two different dimensions, and for the first time relates Tucker rank to tensor tube rank, which further advances the connection between the definitions of tensor rank. On this basis, we establish its non-convex relaxation, and apply FFM to two typical problems of tensor recovery: low-rank tensor completion (LRTC) and tensor robust principal component analysis (TRPCA). LRTC and TRPCA models based on FFM are proposed, and two efficient Alternating Direction Multiplier Method (ADMM) algorithms are developed to solve the proposed model. A variety of real numerical experiments substantiate the superiority of the proposed methods beyond state-of-the-arts.
\end{abstract}

}

\maketitle

\IEEEdisplaynontitleabstractindextext

%
\IEEEpeerreviewmaketitle

\IEEEraisesectionheading{\section{INTRODUCTION}\label{sec:introduction}}

%
%
%
%
\IEEEPARstart{T}{ensor} play an increasingly important role in many applications, such as hyperspectral/multispectral image (HSI/MSI) processing \cite{8359412}, \cite{7467446}, \cite{8657368}, \cite{8941238}, color image/video (CI/CV) processing \cite{1472018163}, \cite{3762018t397}, \cite{7562018767}, \cite{1372020149}, magnetic resonance imaging (MRI) data recovery \cite{341202131}, \cite{1242020783}, \cite{4032018417}, \cite{9412019964}, background subtraction \cite{5032019109}, \cite{8454775}, \cite{8319458}, \cite{7488247}, video rain stripe removal \cite{8237537}, \cite{8578793} and signal reconstruction \cite{7010937}, \cite{7676397}. These applications can often be expressed as tensor recovery problems, that is, recovering underlying tensors from damaged observations. In particular, as two typical examples, tensor robust principal component analysis aims to eliminate sparse outliers, and tensor completeness aims to complete missing elements. The key of tensor recovery is to explore the sparse feature of underlying tensors. Therefore, in the process of tensor recovery, it is very important to consider an appropriate quantitative measure of tensor sparsity. Specifically, the tensor recovery model based on sparsity can usually be expressed as follows:
\begin{eqnarray}
\min_{\mathcal{X}}S(\mathcal{X})+\gamma L(\mathcal{X},\mathcal{Y}),
\label{lossfunction}\end{eqnarray}
where $\mathcal{Y}\in\mathbb{R}^{\mathit{I}_{1}\times\mathit{I}_{2}\times\cdots\times\mathit{I}_{N}}$ is the obesrvation, $L(\mathcal{X},\mathcal{Y})$ is the loss function between $\mathcal{X}$ and $\mathcal{Y}$, $S(\mathcal{X})$ defines the tensor sparsity measure of $\mathcal{X}$ and $\gamma$ is the compromise parameter. It is easy to see that the key problem in constructing (\ref{lossfunction}) is to design an appropriate tensor sparsity measure on data.

As we all know, tensors are high-order generalizations of matrices. Matrix can be reasonably measured by the quantity (ranks) of nonzero singular values. Its relaxed form (nuclear norm) has been proven to be helpful for fine coding of data sparseness in applications, and has inspired various low-rank models and algorithms for different practical problems \cite{137201114}, \cite{8772008905}, \cite{1172003142}, \cite{28622014}, \cite{6751276}, \cite{4550564}, \cite{2892009298}, \cite{55201463}. But the sparsity of tensors cannot be quantitatively measured according to the rank of matrices. The main reason is that there is no unique definition of the rank of a tensor. In the past few decades, the most popular rank definitions are CANDECOMP/PARAFAC(CP) rank based on CP decomposition \cite{41201156}, \cite{36232017}, Tucker rank based on Tucker decomposition \cite{64201881}, \cite{7460200} and tubal-rank and multi-rank \cite{6909886} based on t-svd. Calculating CP rank of a given tensor is NP hard \cite{139201360}, which makes its application quite inconvenient. Tucker rank is often directly extended to higher order by simply finding the rank (or its relaxation) of all tensor mode unfolding matrices. However, such simple rank-sum terms do not fully represent the feature of tensors. To be specific, the properties of tensor sparse measures should surpass the low-rank properties of all unfolding matrices according to modes, rather than simply considering the low-rank properties of each mode. The t-SVD was first proposed by Braman et al. \cite{124120101253} and Kilmer et al. \cite{6412011658}, based on the tensor-tensor product, in which the third-order tensor is integrated to avoid the loss of information inherent in the matrixization or flattening of the tensor \cite{1482013172}. However, the limitation of its tensor-tensor product makes it impossible to apply to higher-order situations. In \cite{2020170} Zheng et al. proposed a new rank (N-tubal rank), in the form of the higher-order tensor using the new way will unfold into multiple three-order tensor, in this way t-SVD is applied to the situation of the higher order. However, based on the tubal rank obtained by t-SVD, the feature shown tend to be more inclined to the relevant feature of the first and the second dimensions of the three-order tensor. N-tube rank adopts t-SVD of multiple three-order tensors, and obtains the correlation feature between any two dimensions. But it still has certain defects in the sparse features of tensors, i.e., it lacks a feature information of each dimension itself. Actually, color video data contains four dimensions, one of which is time, and time has continuity and a feature information of its own. Considering only about the relationship between two dimensions misses the character of each dimension itself. Therefore, it is necessary to find a sparse measure with comprehensiveness. 
\subsection{OUR CONTRIBUTION}
In response to the above problems, this article mainly made four contributions.

Firstly, we propose for the first time a tensor sparse measure that can represent all the feature information of a tensor. In order to describe its unique feature aptly, we name it as full feature measure (FFM). This new measure not only contains the feature information of each dimension and the relevant information features between any two different dimensions, but also integrates Tucker rank and tube rank for the first time and draws on the advantages of the two kinds of rank, thus further advancing the connection between different rank definitions.

Secondly, in response to the above problem, a type of sparsity measure minimization model, which is an NP-hard problem, is established. To address such problem, a relaxed form based on FFM is developed. In the FFM relaxation process, since the sparse measure we proposed is a non-convex function, the accuracy of our method will be greatly reduced if only the traditional kernel norm convex relaxation is adopted. In order to further sharply improve the accuracy and efficiency of such model, we propose a type of non-convex relaxation form of FFM.

Thirdly, by applying FFM to two typical tensor recovery problems, a tensor complete model based on FFM (FFMTC) and a tensor robust principal component analysis model based on FFM (FFMTRPCA) are proposed respectively. Meanwhile, two efficient Alternate Direction Multiplier Method (ADMM) algorithms are designed and used to solve the FFM-based tensor complete minimization problem and the FFM-based tensor robust principal component analysis problem respectively. On this basis, the closed solution of each parameter update is obtained, which facilitates the effective implementation of the presented algorithm.

Fourth, for the low-rank tensor completeness problem, we apply the proposed FFMTC model to a variety of real data tasks, including multispectral image, magnetic resonance imaging, color video and hyperspectral video. In these four experiments with real data, our method is the optimal method compared with state-of-the-art methods, and has obvious improvement compared with the suboptimal method. In particular, the model can still achieve excellent results when the sampling rate is low. For tensor robust principal component analysis, we apply the proposed FFMTRPCA model to the task of restoring hyperspectral data damaged by salt and pepper noise. As we all know, salt and pepper noise is quite serious for image damage. However, when the noise level reaches 0.4, our method can still achieve excellent results in retaining original information and removing noise. A variety of experiments prove that our method is more comprehensive and efficient than other state-of-the-art methods. Through experiments on various real data, the convergence behavior of the algorithm in ADMM framework is obtained. Experiments show that our algorithm can converge quickly and ensure the effectiveness of the algorithm. In summary, the proposed model behaves extremely effective and fairly efficient.

The rest of this paper is structured as follows. Section 2
gives some notations and preliminaries. Section 3 introduces some related work on tensor sparse measure. Section 4 provides our approach and its relaxation. Section 5 supplys two models and algorithms for solving LRTC and TRPCA problems. Section 6 presents numerical experiments conducted on real data. We conclude this work in Section 7.

\section{NOTATIONS and PRELIMINARIES}
In this section, we give some basic notations and briefly introduce some definitions used throughout the paper. Generally, a lowercase letter and an uppercase letter denote a vector x and a marix $X$, respectively. An $N$th-order tensor is denoted by a calligraphic upper case letter $\mathcal{X}\in \mathbb{R}^{\mathit{I}_{1}\times\mathit{I}_{2}\times\cdots\times\mathit{I}_{N}}$ and $\mathit{x}_{i_{1},i_{2},\cdots,i_{N}}$ is its $(i_{1},i_{2},\cdots,i_{N})$-th element. The Frobenius norm of a tensor is defined as $\|\mathcal{X}\|_{F}=(\sum_{i_{1},i_{2},\cdots,i_{N}}\mathit{x}_{i_{1},i_{2},\cdots,i_{N}}^{2})^{1/2}$. For a three order tensor $\mathcal{X}\in\mathbb{R}^{\mathit{I}_{1}\times\mathit{I}_{2}\times\mathit{I}_{3}}$. We use $\bar{\mathcal{X}}$ to denote the tensor generated by performing discrete Fourier transformation (DFT) along each tube of $\mathcal{X}$, i.e., $\bar{\mathcal{X}}=fft(\mathcal{X},[],3)$. The inverse DFT is computed by command ifft satisfying $\mathcal{X}=ifft(\bar{\mathcal{X}},[],3)$. More often, the frontal slice $\mathcal{X}(:,:,i)$ is denoted compactly as $\mathcal{X}^{(i)}$.
\begin{definition}[Tensor Mode-$n$ Unfolding and Folding \cite{12345152009}]
	The mode-$n$ unfolding of a tensor $\mathcal{X}\in \mathbb{R}^{\mathit{I}_{1}\times\mathit{I}_{2}\times\cdots\times\mathit{I}_{N}}$ is denoted as a matrix $\mathcal{X}_{(n)}\in\mathbb{R}^{\mathit{I}_{n}\times\mathit{I}_{1}\cdots\mathit{I}_{n-1}\mathit{I}_{n+1}\cdots\mathit{I}_{N}} $. Tensor element $(i_{1}, i_{2},...,i_{N} )$ maps to matrix element $(i_{n}, j)$, where
	\begin{equation}
	j=1+\sum_{k=1,k\neq n}^{N}(i_{k}-1)\mathit{J}_{k}\quad with\quad \mathit{J}_{k}=\prod_{m=1,m\neq n}^{k-1}\mathit{I}_{m}. 
	\end{equation}
	The mode-$n$ unfolding operator and its inverse are respectively denoted as $unfold_{n}$ and $fold_{n}$, and they satisfy $\mathcal{X}=fold_{n}(\mathcal{X}_{(n)})=fold_{n}(unfold_{n}(\mathcal{X}))$.
\end{definition}
\begin{definition}[The mode-$n$ product of tensor \cite{12345152009}]
	The mode-$n$ product of tensor $\mathcal{X}\in \mathbb{R}^{\mathit{I}_{1}\times\mathit{I}_{2}\times\cdots\times\mathit{I}_{N}}$ with matrix $U\in\mathbb{R}^{\mathit{J}_{n}\times \mathit{I}_{n}}$ is denoted by $\mathcal{Y}=\mathcal{X}\times_{n}U$, where $\mathcal{Y}\in\mathbb{R}^{\mathit{I}_{1}\times\mathit{I}_{2}\times\cdots\mathit{I}_{n-1}\mathit{J}_{n}\mathit{I}_{n+1}\cdots\mathit{I}_{N}}$. Elementwise, we have
	\begin{equation}
	\mathcal{Y}=\mathcal{X}\times_{n}U\quad\Leftrightarrow\quad \mathbf{Y}_{(n)}=U\cdotp unfold_{n}(\mathcal{X}_{(n)}). 
	\end{equation} 
\end{definition}
\begin{definition}[Mode-$k_{1}k_{2}$ slices \cite{2020170}]
	For an $N$th-order tensor $\mathcal{X}\in \mathbb{R}^{\mathit{I}_{1}\times\mathit{I}_{2}\times\cdots\times\mathit{I}_{N}}$, its mode-$k_{1}k_{2}$ slices ($X^{(k_{1}k_{2})},1\leqslant k_{1} <k_{2}\leqslant N,k_{1},k_{2}\in\mathbb{Z}$) are two-dimensional sections, defined by fixing all but the mode-$k_{1}$ and the mode-$k_{2}$ indexes.
\end{definition}
\begin{definition}[Tensor Mode-$k_{1},k_{2}$ Unfolding and Folding \cite{2020170}]
	For an $N$th-order tensor $\mathcal{X}\in \mathbb{R}^{\mathit{I}_{1}\times\mathit{I}_{2}\times\cdots\times\mathit{I}_{N}}$, its mode-$k_{1}k_{2}$ unfolding is a three order tensor denoted by $\mathcal{X}_{(k_{1}k_{2})}\in\mathbb{R}^{\mathit{I}_{k_{1}}\times\mathit{I}_{k_{2}}\times\prod_{s\neq k_{1},k_{2}}\mathit{I}_{s}}$, the frontal slices of which are the lexicographic orderings of the mode-$k_{1}k_{2}$ slices of $\mathcal{X}$. Mathematically, the  $(i_{1}, i_{2},...,i_{N} )$-th element of $\mathcal{X}$ maps to the $(i_{k_{1}},i_{k_{2}},j)$-th element of $\mathcal{X}_{(k_{1}k_{2})}$, where
	\begin{equation}
	j=1+\sum_{s=1,s\neq k_{1},s\neq k_{2}}^{N}(i_{s}-1)\mathit{J}_{s}\quad with\quad \mathit{J}_{s}=\prod_{m=1,m\neq k_{1},m\neq k_{2}}^{s-1}\mathit{I}_{m}. 
	\end{equation}
	The mode-$k_{1}k_{2}$ unfolding operator and its inverse operation are respectively denoted as $\mathcal{X}_{(k_{1}k_{2})}:=t-unfold(\mathcal{X},k_{1},k_{2})$ and $\mathcal{X}:=t-fold(\mathcal{X}_{(k_{1}k_{2})},k_{1},k_{2})$.
\end{definition}

For a three order tensor $\mathcal{X}\in\mathbb{R}^{\mathit{I}_{1}\times\mathit{I}_{2}\times\mathit{I}_{3}}$, the block circulation operation is defined as
\begin{equation}
bcirc(\mathcal{X}):=
\begin{pmatrix}
X^{(1)}& X^{(\mathit{I}_{3})}&\dots& X^{(2)}&\\
X^{(2)}& X^{(1)}&\dots& X^{(3)}&\\
\vdots&\vdots&\ddots&\vdots&\\
X^{(\mathit{I}_{3})}& X^{(\mathit{I}_{3}-1)}&\dots& X^{(1)}&
\end{pmatrix}\in\mathbb{R}^{\mathit{I}_{1}\mathit{I}_{3}\times\mathit{I}_{2}\mathit{I}_{3}}.\nonumber
\end{equation}

The block diagonalization operation and its inverse operation are defined as 
\begin{eqnarray}
&&bdiag(\mathcal{X}):=\begin{pmatrix}
X^{(1)} & & &\\
& X^{(2)} & &\\
& & \ddots &\\
& & & X^{(\mathit{I}_{3})}
\end{pmatrix} \in\mathbb{R}^{\mathit{I}_{1}\mathit{I}_{3}\times\mathit{I}_{2}\mathit{I}_{3}},\nonumber\\
&&bdfold(bdiag(\mathcal{X})):=\mathcal{X}.\nonumber
\end{eqnarray}

The block vectorization operation and its inverse operation are defined as 
\begin{eqnarray}
bvec(\mathcal{X}):=\begin{pmatrix}
X^{(1)}\\X^{(2)}\\\vdots\\X^{(\mathit{I}_{3})}
\end{pmatrix}\in\mathbb{R}^{\mathit{I}_{1}\mathit{I}_{3}\times\mathit{I}_{2}},\quad bvfold(bvec(\mathcal{X})):=\mathcal{X}.\nonumber
\end{eqnarray}
\begin{definition}[T-product \cite{6416568}]
	Let $\mathcal{A}\in\mathbb{R}^{\mathit{I}_{1}\times\mathit{I}_{2}\times\mathit{I}_{3}}$ and $\mathcal{B}\in\mathbb{R}^{\mathit{I}_{2}\times\mathit{J}\times\mathit{I}_{3}}$. Then the t-product $\mathcal{A}\ast\mathcal{B}$ is defined to be a tensor of size $\mathit{I}_{1}\times\mathit{J}\times\mathit{I}_{3}$,
	\begin{eqnarray}
	\mathcal{A}\ast\mathcal{B}:=bvfold(bcirc(\mathcal{A})bvec(\mathcal{B})).\nonumber
	\end{eqnarray}
	
	Since that circular convolution in the spatial domain is equivalent to multiplication in the Fourier domain, the T-product between two tensors $\mathcal{C}=\mathcal{A}\ast\mathcal{B}$ is equivalent to
	\begin{eqnarray}
	\bar{\mathcal{C}}=bdfold(bdiag(\bar{\mathcal{A}})bdiag(\bar{\mathcal{B}})).\nonumber
	\end{eqnarray}
\end{definition}
\begin{figure}
	\centering
	\includegraphics[width=0.7\linewidth]{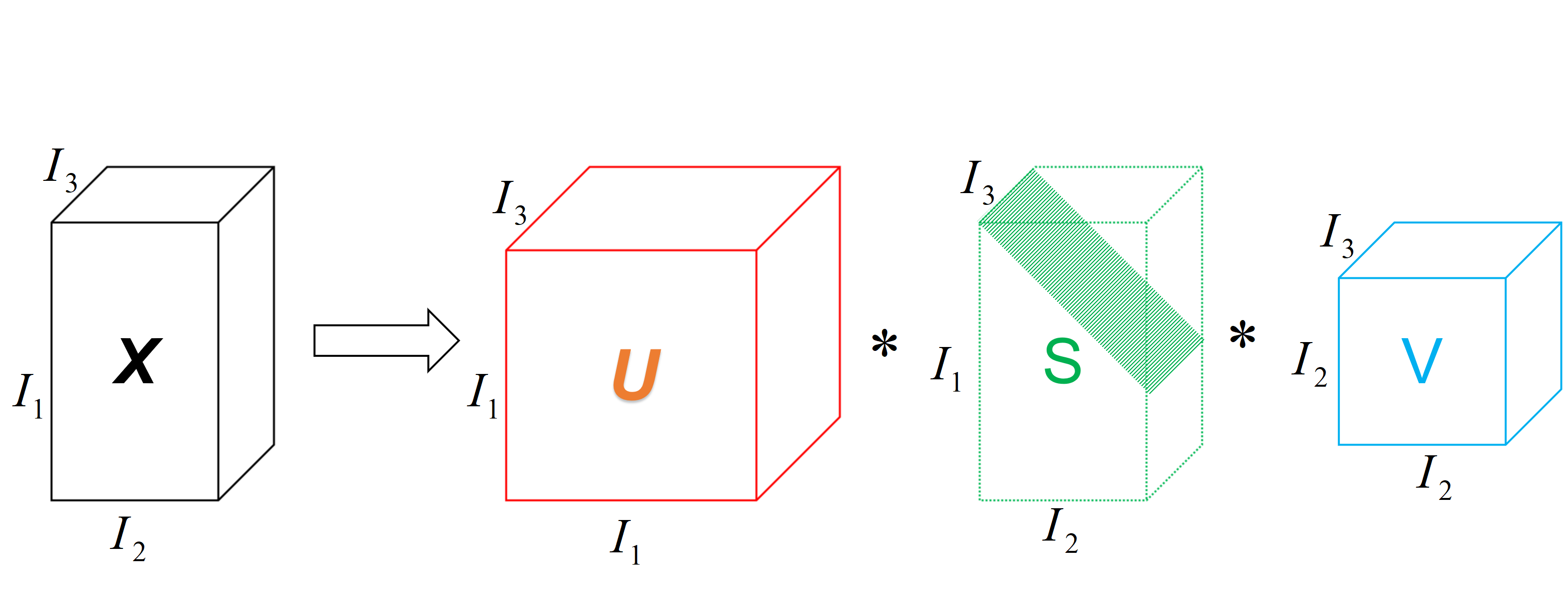}
	\caption{An illustration of the t-SVD of an $\mathit{I}_{1}\times\mathit{I}_{2}\times\mathit{I}_{3}$ tensor}
	\label{tsvd}
\end{figure}
\begin{definition}[Tensor conjugate transpose \cite{6416568}]
	The conjugate transpose of a tensor $\mathcal{A}\in\mathbb{C}^{\mathit{I}_{1}\times\mathit{I}_{2}\times\mathit{I}_{3}}$ is the tensor $\mathcal{A}^{H}\in\mathbb{C}^{\mathit{I}_{2}\times\mathit{I}_{1}\times\mathit{I}_{3}}$ obtained by conjugate transposing each of the frontal slices and then reversing the order of transposed frontal slices 2 through $\mathit{I}_{3}$.
\end{definition}
\begin{definition}[identity tensor \cite{6416568}]
	The identity tensor $\mathcal{I}\in\mathbb{R}^{\mathit{I}_{1}\times\mathit{I}_{1}\times\mathit{I}_{3}}$ is the tensor whose first frontal slice is the $\mathit{I}_{1}\times\mathit{I}_{1}$ identity matrix, and whose other frontal slices are all zeros.
\end{definition}

It is clear that $bcirc(\mathcal{I})$ is the  $\mathit{I}_{1}\mathit{I}_{3}\times\mathit{I}_{1}\mathit{I}_{3}$ identity matrix. So it is easy to get $\mathcal{A}\ast\mathcal{I}=\mathcal{A}$ and $\mathcal{I}\ast\mathcal{A}=\mathcal{A}$. 
\begin{definition}[orthogonal tensor \cite{6416568}]
	A tensor $\mathcal{Q}\in\mathbb{R}^{\mathit{I}_{1}\times\mathit{I}_{1}\times\mathit{I}_{3}}$ is orthogonal if it satisfied
	\begin{eqnarray}
		\mathcal{Q}\ast\mathcal{Q}^{H}=\mathcal{Q}^{H}\ast\mathcal{Q}=\mathcal{I}.\nonumber
	\end{eqnarray}
\end{definition}
\begin{definition}[F-diagonal Tensor \cite{6416568}]
	A tensor is called f-diagonal if each of its frontal slices is a diagonal matrix.
\end{definition}
\begin{theorem}[t-SVD \cite{8606166}]
	Let $\mathcal{X}\in\mathbb{R}^{\mathit{I}_{1}\times\mathit{I}_{2}\times\mathit{I}_{3}}$ be a three order tensor, then it can be factored as 
	\begin{eqnarray}
	\mathcal{X}=\mathcal{U}\ast\mathcal{S}\ast\mathcal{V}^{H},\nonumber
	\end{eqnarray}
	where $\mathcal{U}\in\mathbb{R}^{\mathit{I}_{1}\times\mathit{I}_{1}\times\mathit{I}_{3}}$ and $\mathcal{V}\in\mathbb{R}^{\mathit{I}_{2}\times\mathit{I}_{2}\times\mathit{I}_{3}}$ are orthogonal tensors, and $\mathcal{S}\in\mathbb{R}^{\mathit{I}_{1}\times\mathit{I}_{2}\times\mathit{I}_{3}}$ is an f-diagonal tensor. The t-SVD scheme is illustrated in Fig.\ref{tsvd}, and its computation is given in Algorithm \ref{TSVD1}.
\end{theorem}
\begin{algorithm}[t]
	\caption{t-SVD \cite{8606166}} 
	\hspace*{0.02in} {\bf Input:} 
	$\mathcal{X}\in\mathbb{R}^{\mathit{I}_{1}\times\mathit{I}_{2}\times\mathit{I}_{3}}$ \\
	\hspace*{0.02in} {\bf Output:} 
	t-SVD components $\mathcal{U}$, $\mathcal{S}$ and $\mathcal{V}$ of $\mathcal{A}$.
	\begin{algorithmic}[1]
		\State Compute $\bar{\mathcal{A}}=fft(\mathcal{A},[],3)$.
		\State Compute each frontal slice of $\bar{\mathcal{U}}$, $\bar{\mathcal{S}}$, and $\bar{\mathcal{V}}$ from $\bar{\mathcal{A}}$ by
		\For {$i=1,\dots,[\frac{\mathit{I}_{3}+1}{2}]$} 
		\State $[\bar{\mathcal{U}}^{(i)}, \bar{\mathcal{S}}^{(i)}, \bar{\mathcal{V}}^{(i)}]=SVD(\bar{\mathcal{A}}^{(i)})$;
		\EndFor
		\For {$i=[\frac{\mathit{I}_{3}+1}{2}]+1,\dots,\mathit{I}_{3}$} 
		\State $[\bar{\mathcal{U}}^{(i)}, \bar{\mathcal{S}}^{(i)}, \bar{\mathcal{V}}^{(i)}]=SVD(\bar{\mathcal{A}}^{(i)})$;
		\State $\bar{\mathcal{U}}^{(i)}=conj(\bar{\mathcal{U}}^{(\mathit{I}_{3}-i+2)})$;
		\State $\bar{\mathcal{S}}^{(i)}=\bar{\mathcal{S}}^{(\mathit{I}_{3}-i+2)}$;
		\State $\bar{\mathcal{V}}^{(i)}=conj(\bar{\mathcal{V}}^{(\mathit{I}_{3}-i+2)})$;
		\EndFor
	\end{algorithmic}
	\hspace*{0.02in}  
	Compute $\mathcal{U}=ifft(\bar{\mathcal{U}},[],3)$, $\mathcal{S}=ifft(\bar{\mathcal{S}},[],3)$, and $\mathcal{V}=ifft(\bar{\mathcal{V}},[],3)$. 
\label{TSVD1}\end{algorithm}
\begin{definition}[tensor tubal-rank and multi-rank \cite{6909886}]
	The tubal-rank of a tensor $\mathcal{X}\in\mathbb{R}^{\mathit{I}_{1}\times\mathit{I}_{2}\times\mathit{I}_{3}}$, denoted as $rank_{t}(\mathcal{X})$, is defined to be the number of non-zero singular tubes of $\mathcal{S}$, where $\mathcal{S}$ comes from the t-SVD of $\mathcal{X}:\mathcal{X}=\mathcal{U}\ast\mathcal{S}\ast\mathcal{V}^{H}$. That is 
	\begin{eqnarray}
	rank_{t}(\mathcal{X})=\#\{i:\mathcal{S}(i,:,:)\neq0\}.
	\end{eqnarray}
	The tensor multi-rank of $\mathcal{X}\in\mathbb{R}^{\mathit{I}_{1}\times\mathit{I}_{2}\times\mathit{I}_{3}}$ is a vector, denoted as $rank_{r}(\mathcal{X})\in\mathbb{R}^{\mathit{I}_{3}}$, with the $i$-th element equals to the rank of $i$-th frontal slice of $\mathcal{X}$.
\end{definition}
\section{Related Work}
In this section, we first review some sparsity-based tensor
recovery methods proposed in previous literatures, and then
briefly review two particular sparse measure, both
containing insightful understanding of tensor sparsity.
\subsection{Tucker rank}
The Tucker rank is defined as a vector, the $i$-th element of which is the rank of the mode-i unfolding matrix \cite{4552009500}, i.e.,
\begin{eqnarray}
rank_{tc}:=(rank(\mathcal{X}_{(1)}), rank(\mathcal{X}_{(2)}),\cdots, rank(\mathcal{X}_{(N)}),)
\end{eqnarray}
where $\mathcal{X}$ is an $N$-order tensor and $\mathcal{X}_{(i)}(i=1,2,\dots,N)$ is the mode-$i$ unfolding of $\mathcal{X}$. In order to facilitate the minimization of the Tucker rank, Liu et al. \cite{6138863} considered its convex relaxation, defined as the sum of the nuclear norm (SNN) of unfolding matrices, i.e.,
\begin{eqnarray}
\|\mathcal{X}\|_{SNN}=\sum_{i=1}^{N}\alpha_{i}\|\mathcal{X}_{(i)}\|_{\ast},
\end{eqnarray}
where $\mathcal{X}\in\mathbb{R}^{\mathit{I}_{1}\times\mathit{I}_{2}\times\cdots\times\mathit{I}_{N}}$, $\mathcal{X}_{(i)}$ is unfolding the tensor along the $i$-th dimension, and $\|\cdot\|_{\ast}$ is the nuclear norm of a matrix, i.e., sum of singular values. This simple calculation algorithm makes SNN widely used \cite{341202131}, \cite{7460141}, \cite{101212016}, \cite{51602016}. Although SNN can flexibly make use of features along different modes by adjusting the weights $\alpha_{i}$ \cite{73812014}, such models that expand directly along each dimension are often inadequate in describing the sparsity of tensors. Because unfolding tensor into matrices will inevitably destroy the internal structure of the tensor, which will cause loss of information. For example, unfolding a CV along its temporal dimension damages the spatial information of each frame in the CV. And each unfolding matrix only contains the feature information of each dimension, ignoring the intrinsic correlation between tensor dimensions. 
\subsection{tensor nuclear norm}
To avoid information loss in SNN, Kilmer and Martin \cite{6416568} propose a tensor decomposition named t-SVD with a Fourier transform matrix. Zhang et al. \cite{6909886} proposed tensor tubal-rank and tensor multi-rank based on t-SVD. Directly minimizing the tensor tubal/multi-rank is NP-hard \cite{139201360}, so Zhang et al. \cite{6909886} give a 
definition of the tensor nuclear norm on $\mathcal{X}\in\mathbb{R}^{\mathit{I}_{1}\times\mathit{I}_{2}\times\mathit{I}_{3}}$ corresponding to t-SVD, i.e., Tensor Nuclear Norm (TNN):
\begin{eqnarray}
\|\mathcal{X}\|_{TNN}:=\sum_{i=1}^{\mathit{I}_{3}}\|\bar{\mathcal{X}}^{(i)}\|_{\ast},
\end{eqnarray}
where $\bar{\mathcal{X}}^{(i)}$ is the $i$-th frontal of $\bar{\mathcal{X}}$, with $\bar{\mathcal{X}}=fft(\mathcal{X},[],3)$. The operation of Fourier transform along the third dimension makes TNN based models have a natural computing advantage for video and other data with strong time continuity along a certain dimension. Although TNN is very effective in maintaining the internal structure of tensors, It's easy to find obvious two shortcomings. One is that it cannot be applied to higher-order ($N>3$) applications, because t-SVD limits its application to a three-order tensor. The other is that it lacks different correlations under different modes. In the t-SVD framework, for the three-order tensor, the correlations of the first and second dimensions are represented by matrix singular values, while the features of the third dimension are generated by embedded cyclic convolution. Therefore, TNN tends to highlight the correlation between the first and second dimensions. In view of this, Zheng et al. \cite{2020170} proposed n-tubal rank to remedy the above defects.
\begin{definition}[n-tubal rank \cite{2020170}]
	The tensor n-tubal rank is a vector consisting of the tubal ranks of all mode-$k_{1}k_{2}$ ($k_{1}\neq k_{2}$) unfolding tensors, i.e.,
	\begin{eqnarray}
	N-rank_{t}(\mathcal{X}):=(rank_{t}(\mathcal{X}_{12}),rank_{t}(\mathcal{X}_{13}),\cdots,\nonumber\\rank_{t}(\mathcal{X}_{1N}),rank_{t}(\mathcal{X}_{23}),\cdots,\nonumber rank_{t}(\mathcal{X}_{2N}),\cdots,\\rank_{t}(\mathcal{X}_{N-1N}))\in\mathbb{R}^{N(N-1)/2}.
	\end{eqnarray}
\end{definition}
It is not difficult to find that n-tube rank makes up for the defect of TNN, and its feature is the correlation between any two dimensions. However, it is still not sufficient to express the correlation between any two dimensions of a tensor, and the feature of each dimension are of vital importance. For example, the time dimension in the video has a certain correlation with the space dimension, but its unique feature in time cannot be ignored. The specific experimental results can be found in the experimental section, and there is a lot of room for improvement from its observation.

To be inspired by this and to solve this very difficult problem right now, we attempt to propose a measure for more reasonable and comprehensive assessing tensor sparsity.
\section{Tensor Full Feature Measure and Relaxation}
\subsection{Tensor Full Feature Measure}
Through more fully consider the properties of Tucker rank and N-tubal rank comprehensively, the full feature measure (FFM) of the proposed tensor $\mathcal{X}$ has the following expression:
\begin{eqnarray}
\mathit{S}(\mathcal{X})=\sum_{1\leqslant k_{1}\leqslant k_{2}\leqslant N}\beta_{k_{1}k_{2}}rank(\mathcal{X}_{(k_{1}k_{2})})
\label{FFM1}\end{eqnarray}
where parameter $\beta_{k_{1}k_{2}}\geqslant0(1\leqslant k_{1}\leqslant k_{2}\leqslant N,k_{1},k_{2}\in\mathbb{Z})$ and $\sum_{1\leqslant k_{1}\leqslant k_{2}\leqslant N}\beta_{k_{1}k_{2}}=1$ and $\mathcal{X}_{(k_{1}k_{2})}=unfold_{k_{1}}(\mathcal{X})(k_{1}=k_{2})$, $\mathcal{X}_{(k_{1}k_{2})}=t-unfold(\mathcal{X},k_{1},k_{2})(k_{1}\neq k_{2})$.
It is not difficult to find that (\ref{FFM1}) can be divided into two parts. The first part is $k_{1}=k_{2}$, which mainly delivers the feature information of each dimension. The second part is $k_{1}\neq k_{2}$, which can convey the feature information of the correlation between any two different dimensions. The weight parameter $\beta_{k_{1}k_{2}}$ is used to balance the size difference between the different ranks, so that the different ranks of different representations are equal. This comprehensive consideration makes the tensor not only has the internal correlation of any two dimensions, but also contain the low-rank property of the tensor space along each mode, so that the tensor sparsity can be described more fully.
\subsection{Relaxation}
It should be noted that the rank term in (\ref{FFM1}) can only take discrete values, which will produce optimization problems that are difficult to solve in practical applications \cite{8000407}. However, directly using the nuclear norm as its loose approximation will still have a considerable gap with the original rank minimization problem, which usually leads to excessive punishment of the optimization problem and suboptimality of the original problem solution. To remedy this shortcoming, we therefore exploit non-convex relaxation of the FFM, i.e., a logarithmic function is applied to each singular value. The effectiveness of this relaxation has been confirmed in previous research \cite{8772008905}, \cite{2920151}, \cite{286220142869}, \cite{7202003727}, \cite{9340243}, and this can improve the accuracy of the method. Therefore, the non-convex relaxation form of FFM proposed by us is as follows:
\begin{eqnarray}
\mathit{S}^{\ast}(\mathcal{X})=\sum_{1\leqslant k_{1}\leqslant k_{2}\leqslant N}\beta_{k_{1}k_{2}}Log(\mathcal{X}_{(k_{1}k_{2})}),
\end{eqnarray}
where 
\begin{eqnarray}
Log(\mathcal{X}_{(k_{1}k_{2})})=\sum_{m}(log(\sigma_{m}(\mathcal{X}_{(k_{1})})+\varepsilon)),\quad k_{1}=k_{2},\nonumber
\\Log(\mathcal{X}_{(k_{1}k_{2})})=\sum_{i=1}^{J_{k_{1}k_{2}}}\sum_{m}(log(\sigma_{m}(\mathcal{X}_{(k_{1}k_{2})}^{(i)})+\varepsilon)),\quad k_{1}\neq k_{2},\nonumber
\end{eqnarray}
are both with log-sum forms, $\varepsilon$ is a small positive number, and $\sigma_{m}(\mathcal{X}_{(k_{1})})$ and $ \sigma_{m}(\mathcal{X}_{(k_{1}k_{2})}^{(i)})$ respectively defines the
$m$-th singular value of $\mathcal{X}_{(k_{1})}$ and $\mathcal{X}_{(k_{1}k_{2})}^{(i)}$, and $J_{k_{1}k_{2}}=\prod_{s\neq k_{1},k_{2}}\mathit{I}_{s}$. In the later section, we will use this relaxation form of FFM to establish FFM-based models.
\section{FFM-based models and solving algorithms}

In this section, we apply the FFM to low rank tensor complete (LRTC) and tensor robust principal component analysis (TRPCA) and propose the FFM-based models with ADMM-based solving algorithms.
\subsection{FFM-based LRTC model}
\begin{algorithm}[t]
	\caption{FFMTC} 
	\hspace*{0.02in} {\bf Input:} 
	an incomplete tensor $\mathcal{Z}$, the index set of the known elements $\Omega$, convergence criteria $\epsilon$, maximum iteration number $K$. \\
	\hspace*{0.02in} {\bf Initialization:} 
	$\mathcal{X}^{0}=\mathcal{Z}_{\Omega}$, $\mathcal{M}_{k_{1}k_{2}}^{0}=\mathcal{X}^{0}$, $\mu_{k_{1}k_{2}}^{0}>0$, $\rho>1$.
	\begin{algorithmic}[1]
		\While{not converged and $k<K$} 
		\State Updating $\mathcal{M}_{k_{1}k_{2}}^{k}$ via (\ref{UPM});
		\State Updating $\mathcal{X}^{k}$ via (\ref{UPX});
		\State Updating the multipliers $\mathcal{Q}_{k_{1}k_{2}}^{k}$ via (\ref{UPQ});
		\State $\mu_{k_{1}k_{2}}^{k}=\rho\mu_{k_{1}k_{2}}^{k-1}$, $k=k+1$;
		\State Check the convergence conditions $\|\mathcal{X}^{k+1}-\mathcal{X}^{k}\|_{\infty}\leq\epsilon$.
		\EndWhile
		\State \Return $\mathcal{X}^{k+1}$.
	\end{algorithmic}
	\hspace*{0.02in} {\bf Output:} 
	Completed tensor $\mathcal{X}=\mathcal{X}^{k+1}$.
	\label{TC}\end{algorithm}
Tensor completion aims at estimating the missing elements from an incomplete observation tensor. Considering an $N$-order tensor $\mathcal{X}\in\mathbb{R}^{\mathit{I}_{1}\times\mathit{I}_{2}\times\cdots\times\mathit{I}_{N}}$, the proposed FFM-based LRTC model is formulated as follow
\begin{eqnarray}
\min_{\mathcal{X}}\mathit{S}^{\ast}(\mathcal{X})\quad s.t.\quad \mathcal{P}_{\Omega}(\mathcal{X}-\mathcal{Z})=0,\label{FTC1}
\end{eqnarray}
where $\mathcal{X}$ is the reconstructed tensor and $\mathcal{Z}$ is the observed tensor, $\Omega$ is the index set for the known entries, and $\mathcal{P}_{\Omega}(\mathcal{X})$ is a projection
operator that keeps the entries of $\mathcal{X}$ in $\Omega$ and sets all others to zero. Let
\begin{eqnarray}
\Phi_{\mathbb{G}}(\mathcal{X}):=\left\{\begin{array}{l}
0,\qquad if\quad\mathcal{X}\in\mathbb{G},
\\\infty,\qquad otherwise
\end{array}
\right.
\end{eqnarray}
where $\mathbb{G}:=\{\mathcal{X}\in\mathbb{R}^{\mathit{I}_{1}\times\mathit{I}_{2}\times\cdots\times\mathit{I}_{N}},\mathcal{P}_{\Omega}(\mathcal{X}-\mathcal{Z})=0\}$. Then (\ref{FTC1}) can be expressed equivalently as
\begin{eqnarray}
\min_{\mathcal{X}}\sum_{1\leqslant k_{1}\leqslant k_{2}\leqslant N}\beta_{k_{1}k_{2}}Log(\mathcal{X}_{(k_{1}k_{2})})+\Phi_{\mathbb{G}}(\mathcal{X})\label{FTC2}
\end{eqnarray}
where $\beta_{k_{1}k_{2}}\geqslant0(1\leqslant k_{1}\leqslant k_{2}\leqslant N,k_{1},k_{2}\in\mathbb{Z})$ and $\sum_{1\leqslant k_{1}\leqslant k_{2}\leqslant N}\beta_{k_{1}k_{2}}=1$.
Next, we use the ADMM to solve (\ref{FTC2}). We first introduce auxiliary variables $\mathcal{M}_{k_{1}k_{2}}$, and then rewrite (\ref{FTC2}) as the following equivalent constrained problem:
\begin{eqnarray}
\min_{\mathcal{X}}\sum_{1\leqslant k_{1}\leqslant k_{2}\leqslant N}\beta_{k_{1}k_{2}}Log((\mathcal{M}_{k_{1}k_{2}})_{(k_{1}k_{2})})+\Phi_{\mathbb{G}}(\mathcal{X})\label{FTC3}
\\s.t.\quad \mathcal{X}=\mathcal{M}_{k_{1}k_{2}},1\leqslant k_{1}\leqslant k_{2}\leqslant N,k_{1},k_{2}\in\mathbb{Z}.\nonumber
\end{eqnarray}
The augmented Lagrangian function of (\ref{FTC3}) can be expressed in the following concise form:
\begin{eqnarray}
L_{\mu_{k_{1}k_{2}}}(\mathcal{X},\mathcal{M}_{k_{1}k_{2}},\mathcal{Q}_{k_{1}k_{2}})=\sum_{1\leqslant k_{1}\leqslant k_{2}\leqslant N}\beta_{k_{1}k_{2}}Log((\mathcal{M}_{k_{1}k_{2}})_{(k_{1}k_{2})})\nonumber
\\+\Phi_{\mathbb{G}}(\mathcal{X})+\frac{\mu_{k_{1}k_{2}}}{2}\|\mathcal{X}-\mathcal{M}_{k_{1}k_{2}}+\frac{\mathcal{Q}_{k_{1}k_{2}}}{\mu_{k_{1}k_{2}}}\|_{F}^{2},
\end{eqnarray}
where $\mathcal{Q}_{k_{1}k_{2}}(1\leqslant k_{1}\leqslant k_{2}\leqslant N)$ are the Lagrange multipliers, $\mu_{k_{1}k_{2}}$ are positive scalars. Then we can solve the problem under the ADMM framework. For the sake of convenience, we denote the variable updated by the iteration as $(\cdot)^{+}$, and omit the specific number of iterations. 

Fixed $\mathcal{X}$ and $\mathcal{Q}_{k_{1}k_{2}}$, then $\mathcal{M}_{k_{1}k_{2}}$ can be solved by the following minimization problem:
\begin{eqnarray}
\min_{\mathcal{M}_{k_{1}k_{2}}}&&\sum_{1\leqslant k_{1}\leqslant k_{2}\leqslant N}\beta_{k_{1}k_{2}}Log((\mathcal{M}_{k_{1}k_{2}})_{(k_{1}k_{2})})\nonumber
\\&&\qquad+\frac{\mu_{k_{1}k_{2}}}{2}\|\mathcal{X}-\mathcal{M}_{k_{1}k_{2}}+\frac{\mathcal{Q}_{k_{1}k_{2}}}{\mu_{k_{1}k_{2}}}\|_{F}^{2}.\label{FFMM}
\end{eqnarray}
Since the relationship between $k_{1}$ and $k_{2}$ will produce different minimization problems, we divide (\ref{FFMM}) into two cases to consider.

In case 1: $k_{1}=k_{2}$, $(\mathcal{M}_{k_{1}k_{2}})_{(k_{1}k_{2})}$ is a matrix. In this case the problem (\ref{FFMM}) can be obtained as follows:
\begin{eqnarray}
\min_{\mathcal{M}_{k_{1}k_{2}}}&&\sum_{1\leqslant k_{1}\leqslant k_{2}\leqslant N}\beta_{k_{1}k_{2}}\sum_{m}(log(\sigma_{m}(\mathcal{M}_{k_{1}k_{2}})_{(k_{1}k_{2})}+\varepsilon))\nonumber
\\&&\qquad+\frac{\mu_{k_{1}k_{2}}}{2}\|\mathcal{X}-\mathcal{M}_{k_{1}k_{2}}+\frac{\mathcal{Q}_{k_{1}k_{2}}}{\mu_{k_{1}k_{2}}}\|_{F}^{2}.\label{FFMM1}
\end{eqnarray}
To solve (\ref{FFMM1}), we introduce the following theorem \cite{8000407}.
\begin{theorem}[\cite{8000407}]
	Assuming that $Y\in\mathbb{R}^{\mathit{I}_{1}\times\mathit{I}_{2}}$ is a matirx, a minimizer to
	\begin{eqnarray}
	\min_{X} \alpha Log(X)+\frac{1}{2}\|X-Y\|_{F}^{2},\nonumber
	\end{eqnarray}
	is given by
	\begin{eqnarray}
	X=U\Sigma_{\alpha}V^{T},\nonumber
	\end{eqnarray}
	where $\Sigma_{\alpha}=diag(D_{\alpha,\varepsilon}(\sigma_{1}),D_{\alpha,\varepsilon}(\sigma_{2}),\cdots,D_{\alpha,\varepsilon}(\sigma_{m}))$ and $Y=Udiag(\sigma_{1},\sigma_{2},\cdots,\sigma_{m})V^{T}$. Here, $D_{\alpha,\varepsilon}(\cdot)$ is the thresholding operator defined as:
	\begin{eqnarray}
	D_{\alpha,\varepsilon}(x)=\left\{\begin{array}{l}
	 0,\qquad\qquad\qquad\qquad if\quad|x|\leqslant 2\sqrt{\alpha}-\varepsilon,
	\\sign(x)(\frac{l_{1}(x)+l_{2}(x)}{2}),\quad if\quad|x|> 2\sqrt{\alpha}-\varepsilon
	\end{array}
	\right.
	\end{eqnarray}
	where $l_{1}(x)=|x|-\varepsilon,l_{2}(x)=\sqrt{(|x|+\varepsilon)^{2}-4\alpha}$.\label{theorem2}
\end{theorem}

From Theorem $\ref{theorem2}$, the update of $\mathcal{M}_{k_{1}k_{2}}$ can be obtained as follows:
\begin{eqnarray}
\mathcal{M}_{k_{1}k_{2}}^{+}=fold_{k_{1}}(U_{1}\Sigma_{\frac{\beta_{k_{1}k_{2}}}{\mu_{k_{1}k_{2}}}}V_{1}^{T}),
\end{eqnarray}
where $U_{1}diag(\sigma_{1},\sigma_{2},\cdots,\sigma_{m})V_{1}^{T}$ is the SVD of $unfold_{k_{1}}(\mathcal{X}+\frac{\mathcal{Q}_{k_{1}k_{2}}}{\mu_{k_{1}k_{2}}})$, and let $\nu_{k_{1}k_{2}}=\frac{\beta_{k_{1}k_{2}}}{\mu_{k_{1}k_{2}}}$, then $\Sigma_{\nu_{k_{1}k_{2}}}=diag(D_{\nu_{k_{1}k_{2}},\varepsilon}(\sigma_{1}),D_{\nu_{k_{1}k_{2},\varepsilon}}(\sigma_{2}),\cdots,D_{\nu_{k_{1}k_{2}},\varepsilon}(\sigma_{m}))$.

In case 2: $k_{1}\neq k_{2}$, $(\mathcal{M}_{k_{1}k_{2}})_{(k_{1}k_{2})}$ is a three-order tensor. We transformed the question (\ref{FFMM}) into the following form:
\begin{eqnarray}
\min_{\mathcal{M}_{k_{1}k_{2}}}&&\sum_{1\leqslant k_{1}\leqslant k_{2}\leqslant N}\beta_{k_{1}k_{2}}\sum_{i=1}^{J_{k_{1}k_{2}}}\sum_{m}(log(\sigma_{m}(\mathcal{X}_{(k_{1}k_{2})}^{(i)})+\varepsilon))\nonumber
\\&&\qquad+\frac{\mu_{k_{1}k_{2}}}{2}\|\mathcal{X}-\mathcal{M}_{k_{1}k_{2}}+\frac{\mathcal{Q}_{k_{1}k_{2}}}{\mu_{k_{1}k_{2}}}\|_{F}^{2}.\label{FFMM2}
\end{eqnarray}
To solve (\ref{FFMM2}), we introduce the following theorem \cite{9200799}.
\begin{theorem}[\cite{9200799}]
	For any $\mathcal{A}\in\mathbb{R}^{\mathit{I}_{1}\times\mathit{I}_{2}\times\mathit{I}_{3}}$ and $\lambda>0$, the local optimal solution of the following minimization 
	\begin{eqnarray}
	\min_{\mathcal{B}} \frac{1}{2}\|\mathcal{B}-\mathcal{A}\|_{F}^{2}+\lambda Log(\mathcal{X})
	\end{eqnarray} 
	is given by $\mathcal{B}=\mathcal{U}\ast\mathcal{S}_{1}\ast\mathcal{V}^{H}$, where $\mathcal{U}$ and $\mathcal{V}$ derive from the t-SVD of $\mathcal{A}=\mathcal{U}\ast\mathcal{S}_{2}\ast\mathcal{V}^{H}$. More importantly, the ith frontal slice of DFT of $\mathcal{S}_{1}$  and $\mathcal{S}_{2}$, i.e., $\bar{\mathcal{S}_{1}}^{(i)}=diag(\sigma_{1}(\bar{\mathcal{B}}^{(i)}),\sigma_{2}(\bar{\mathcal{B}}^{(i)}),\cdots,\sigma_{m}(\bar{\mathcal{B}}^{(i)}))$ and $\bar{\mathcal{S}_{2}}^{(i)}=diag(\sigma_{1}(\bar{\mathcal{A}}^{(i)}),\sigma_{2}(\bar{\mathcal{A}}^{(i)}),\cdots,\sigma_{m}(\bar{\mathcal{A}}^{(i)}))$, has the following relationship
	\begin{eqnarray}
	\sigma_{j}(\bar{\mathcal{B}}^{(i)})=\left\{\begin{array}{l}
	\frac{1}{2}(\sqrt{\Delta_{ij}}+\sigma_{j}(\bar{\mathcal{A}}^{(i)})-\varepsilon),if\quad\Delta_{ij}>0,\\
	\sqrt{\Delta_{ij}}>\varepsilon-\sigma_{j}(\bar{\mathcal{A}}^{(i)}),\sigma_{j}^{2}(\bar{\mathcal{A}}^{(i)})>2\phi;
	\\0,\quad otherwise. 
	\end{array}
	\right.
	\end{eqnarray}
where
	\begin{eqnarray}
	&\Delta_{ij}=(\sigma_{j}(\bar{\mathcal{A}}^{(i)})+\varepsilon)^{2}-4\lambda,\nonumber
	\\&\phi=\frac{1}{8}(\sqrt{\Delta_{ij}}-\sigma_{j}(\bar{\mathcal{A}}^{(i)})-\varepsilon)^{2}+\lambda log(\frac{\sigma_{j}(\bar{\mathcal{A}}^{(i)})+\sqrt{\Delta_{ij}}+\varepsilon}{2\varepsilon}).\nonumber
	\end{eqnarray} 
\label{theorem3}\end{theorem}

From Theorem $\ref{theorem3}$, the update of $\mathcal{M}_{k_{1}k_{2}}$ can be obtained as follows:
\begin{eqnarray}
\mathcal{M}_{k_{1}k_{2}}^{+}=t-fold(\mathcal{U}\ast\mathcal{S}_{1}\ast\mathcal{V}^{H},k_{1},k_{2}),
\end{eqnarray}
where $\mathcal{S}_{1}$ is given by Theorem \ref{theorem3} and    $\mathcal{U}\ast\mathcal{S}_{2}\ast\mathcal{V}^{H}$ is the t-SVD of $t-unfold(\mathcal{X}+\frac{\mathcal{Q}_{k_{1}k_{2}}}{\mu_{k_{1}k_{2}}},k_{1},k_{2})$.

In summary, the update of $\mathcal{M}_{k_{1}k_{2}}$ is expressed in the following form:
\begin{eqnarray}
\mathcal{M}_{k_{1}k_{2}}^{+}=\left\{\begin{array}{l}
fold_{k_{1}}(U_{1}\Sigma_{\frac{\beta_{k_{1}k_{2}}}{\mu_{k_{1}k_{2}}}}V_{1}^{T}),\quad if\quad k_{1}=k_{2},
\\t-fold(\mathcal{U}\ast\mathcal{S}_{1}\ast\mathcal{V}^{H},k_{1},k_{2}),\quad if\quad k_{1}\neq k_{2}
\end{array}
\right.\label{UPM}
\end{eqnarray}

Fixed $\mathcal{M}_{k_{1}k_{2}}$ and $\mathcal{Q}_{k_{1}k_{2}}$, the minimization problem of $\mathcal{X}$ is as follows:
\begin{eqnarray}
\min_{\mathcal{X}}\Phi_{\mathbb{G}}(\mathcal{X})+\sum_{1\leqslant k_{1}\leqslant k_{2}\leqslant N}\frac{\mu_{k_{1}k_{2}}}{2}\|\mathcal{X}-\mathcal{M}_{k_{1}k_{2}}+\frac{\mathcal{Q}_{k_{1}k_{2}}}{\mu_{k_{1}k_{2}}}\|_{F}^{2}.\label{FFXX1}
\end{eqnarray}
The closed form of $\mathcal{X}$ can be derived by setting the derivative of (\ref{FFXX1}) to zero. We can now update $\mathcal{X}$ by the following equation:
\begin{eqnarray}
\mathcal{X}^{+}=\mathcal{P}_{\Omega^{c}}(\frac{\sum_{1\leqslant k_{1}\leqslant k_{2}\leqslant N}\mu_{k_{1}k_{2}}(\mathcal{M}_{k_{1}k_{2}}-\frac{\mathcal{Q}_{k_{1}k_{2}}}{\mu_{k_{1}k_{2}}})}{\sum_{1\leqslant k_{1}\leqslant k_{2}\leqslant N}\mu_{k_{1}k_{2}}})+\mathcal{P}_{\Omega}(\mathcal{Z})\label{UPX}
\end{eqnarray}
Finally, multipliers $\mathcal{Q}_{k_{1}k_{2}}$ are updated as follows:
\begin{eqnarray}
	\mathcal{Q}_{k_{1}k_{2}}^{+}=\mathcal{Q}_{k_{1}k_{2}}+\mu_{k_{1}k_{2}}(\mathcal{X}-\mathcal{M}_{K_{1}K_{2}}).\label{UPQ}
\end{eqnarray}

FFM-based LRTC model computation is given in Algorithm \ref{TC}. The main per-iteration cost lies in the update of $\mathcal{M}_{k_{1}k_{2}}$, which requires computing SVD and t-SVD . The per-iteration complexity is $O(LE(\sum_{1\leqslant k_{1}< k_{2}\leqslant N}[log(le_{k_{1}k_{2}})+\min(\mathit{I}_{k_{1}},\mathit{I}_{k_{2}})]+\sum_{1\leqslant k_{1} \leqslant N}\mathit{I}_{k_{1}}))$, where $LE=\prod_{i=1}^{N}\mathit{I}_{i}$ and $le_{k_{1}k_{2}}=LE/(\mathit{I}_{k_{1}}\mathit{I}_{k_{2}})$.

\subsection{FFM-based TRPCA model}
\begin{algorithm}[t]
	\caption{FFMTPRCA} 
	\hspace*{0.02in} {\bf Input:} 
	The corrupted observation tensor $\mathcal{T}$, convergence criteria $\epsilon$, maximum iteration number $K$. \\
	\hspace*{0.02in} {\bf Initialization:} 
	$\mathcal{L}^{0}=\mathcal{T}$, $\mathcal{G}_{l_{1}l_{2}}^{0}=\mathcal{L}^{0}$, $\mu_{l_{1}l_{2}}^{0}>0$, $\rho>0$, $\tau>1$.
	\begin{algorithmic}[1]
		\While{not converged and $k<K$} 
		\State Updating $\mathcal{G}_{l_{1}l_{2}}^{k}$ via (\ref{UPG});
		\State Updating $\mathcal{L}^{k}$ via (\ref{UPL});
		\State Updating $\mathcal{E}^{k}$ via (\ref{UPE});
		\State Updating the multipliers $\mathcal{R}_{k_{1}k_{2}}^{k}$ and $\mathcal{F}^{k}$ via (\ref{UPM21});
		\State $\mu_{k_{1}k_{2}}^{k}=\tau\mu_{k_{1}k_{2}}^{k-1}$, $\rho^{k}=\tau\rho^{k-1}$ $k=k+1$;
		\State Check the convergence conditions $\|\mathcal{L}^{k+1}-\mathcal{L}^{k}\|_{\infty}\leq\epsilon$.
		\EndWhile
		\State \Return $\mathcal{L}^{k+1}$ and $\mathcal{E}^{k+1}$.
	\end{algorithmic}
	\hspace*{0.02in} {\bf Output:} 
	$\mathcal{L}$ and $\mathcal{E}$.
	\label{TPRCA}\end{algorithm}
Tensor robust PCA (TRPCA) aims to recover the tensor from
grossly corrupted observations. Using the proposed FFM
, we can get the following FFM-based TRPCA model:
\begin{eqnarray}
\min_{\mathcal{L},\mathcal{E}}\qquad\mathit{S}^{\ast}(\mathcal{L})+\tau\|\mathcal{E}\|_{1}\quad s.t.\quad \mathcal{T}=\mathcal{L}+\mathcal{E},\label{FPCA1}
\end{eqnarray}
where $\mathcal{T}$ is the corrupted observation tensor, $\mathcal{L}$ is the low-rank component, $\mathcal{E}$ is the sparse component, and $\tau$ is a tuning parameter compromising $\mathcal{L}$ and $\mathcal{E}$. We use the ADMM to solve (\ref{FPCA1}). Firstly, we introduce auxiliary variables $\mathcal{G}_{l_{1}l_{2}}$, and then rewrite (\ref{FPCA1}) as the following equivalent constrained problem:
\begin{eqnarray}
\min_{\mathcal{L},\mathcal{E}}&&\sum_{1\leqslant l_{1}\leqslant l_{2}\leqslant N}\beta_{l_{1}l_{2}}Log((\mathcal{G}_{l_{1}l_{2}})_{(l_{1}l_{2})})+\tau\|\mathcal{E}\|_{1}\nonumber
\\&&s.t.\mathcal{T}=\mathcal{L}+\mathcal{E},\nonumber
\\&&\quad \mathcal{L}=\mathcal{G}_{l_{1}l_{2}},1\leqslant l_{1}\leqslant l_{2}\leqslant N,l_{1},l_{2}\in\mathbb{Z}.
\label{FPCA2}\end{eqnarray}
The augmented Lagrangian function of (\ref{FPCA2}) can be expressed in the following concise form:
\begin{eqnarray}
&&L_{\mu_{l_{1}l_{2}},\rho}(\mathcal{L}, \mathcal{G}_{l_{1}l_{2}}, \mathcal{E}, \mathcal{R}_{k_{1}k_{2}},\mathcal{F})=\tau\|\mathcal{E}\|_{1}\nonumber\\&&+\sum_{1\leqslant l_{1}\leqslant l_{2}\leqslant N}\beta_{l_{1}l_{2}}Log((\mathcal{G}_{l_{1}l_{2}})_{(l_{1}l_{2})})\\&&+\frac{\mu_{l_{1}l_{2}}}{2}\|\mathcal{L}-\mathcal{G}_{l_{1}l_{2}}+\frac{\mathcal{R}_{l_{1}l_{2}}}{\mu_{l_{1}l_{2}}}\|_{F}^{2}+\frac{\rho}{2}\|\mathcal{T}-\mathcal{L}-\mathcal{E}+\frac{\mathcal{F}}{\rho}\|_{F}^{2}\nonumber.
\end{eqnarray}
where $\mathcal{F}$ and ${R}_{l_{1}l_{2}}(1\leqslant k_{1}\leqslant k_{2}\leqslant N)$ are the Lagrange multipliers, $\mu_{k_{1}k_{2}}$ and $\rho$ are positive scalar. Then we can solve the problem under the ADMM framework. Similar to FFM-based LRTC model, we denote the variable updated by the iteration as $(\cdot)^{+}$, and omit the specific number of iterations. 

Fixed $\mathcal{L}$, $\mathcal{E}$, $\mathcal{R}_{k_{1}k_{2}}$ and $\mathcal{F}$, then $\mathcal{G}_{l_{1}l_{2}}$ can be solved by the following minimization problem:
\begin{eqnarray}
\min_{\mathcal{G}_{l_{1}l_{2}}} \sum_{1\leqslant l_{1}\leqslant l_{2}\leqslant N}\beta_{l_{1}l_{2}}Log((\mathcal{G}_{l_{1}l_{2}})_{(l_{1}l_{2})})\nonumber\\+\frac{\mu_{l_{1}l_{2}}}{2}\|\mathcal{L}-\mathcal{G}_{l_{1}l_{2}}+\frac{\mathcal{R}_{l_{1}l_{2}}}{\mu_{l_{1}l_{2}}}\|_{F}^{2}
\end{eqnarray}
From the solution process of $\mathcal{M}_{k_{1}k_{2}}$ in FFM-based LRTC model, the update of $\mathcal{G}_{l_{1}l_{2}}$ is as follows:
\begin{eqnarray}
\mathcal{G}_{l_{1}l_{2}}^{+}=\left\{\begin{array}{l}
fold_{l_{1}}(U_{1}\Sigma_{\frac{\beta_{l_{1}l_{2}}}{\mu_{l_{1}l_{2}}}}V_{1}^{T}),\quad if\quad l_{1}=l_{2},
\\t-fold(\mathcal{U}\ast\mathcal{S}_{1}\ast\mathcal{V}^{H},l_{1},l_{2}),\quad if\quad l_{1}\neq l_{2}
\end{array}
\right.\label{UPG}
\end{eqnarray}
where $U_{1}diag(\sigma_{1},\sigma_{2},\dots,\sigma_{m})V_{1}^{T}$ is the SVD of $unfold_{l_{1}}(\mathcal{L}+\frac{\mathcal{R}_{l_{1}l_{2}}}{\mu_{l_{1}l_{2}}})$ and $\mathcal{S}_{1}$ is given by Theorem \ref{theorem3} and    $\mathcal{U}\ast\mathcal{S}_{2}\ast\mathcal{V}^{H}$ is the t-SVD of $t-unfold(\mathcal{L}+\frac{\mathcal{R}_{l_{1}l_{2}}}{\mu_{l_{1}l_{2}}},l_{1},l_{2})$.

Then fixed $\mathcal{G}_{l_{1}l_{2}}$, $\mathcal{E}$, $\mathcal{R}_{k_{1}k_{2}}$ and $\mathcal{F}$, the minimization problem $\mathcal{L}$ is converted into the following form:
\begin{eqnarray}
\min_{\mathcal{L}}&&\sum_{1\leqslant l_{1}\leqslant l_{2}\leqslant N}\beta_{l_{1}l_{2}}\frac{\mu_{l_{1}l_{2}}}{2}\|\mathcal{L}-\mathcal{G}_{l_{1}l_{2}}+\frac{\mathcal{R}_{l_{1}l_{2}}}{\mu_{l_{1}l_{2}}}\|_{F}^{2}\nonumber\\&&+\frac{\rho}{2}\|\mathcal{T}-\mathcal{L}-\mathcal{E}+\frac{\mathcal{F}}{\rho}\|_{F}^{2}.\label{FFLL1}
\end{eqnarray}
The closed form of $\mathcal{L}$ can be derived by setting the derivative of (\ref{FFLL1}) to zero. We can now update $\mathcal{L}$ by the following equation:
\begin{eqnarray}
\mathcal{L}^{+}=\dfrac{\sum_{1\leqslant l_{1}\leqslant l_{2}\leqslant N}\mu_{l_{1}l_{2}}(\mathcal{G}_{l_{1}l_{2}}-\frac{\mathcal{R}_{l_{1}l_{2}}}{\mu_{l_{1}l_{2}}})+\rho(\mathcal{T}-\mathcal{E}+\frac{\mathcal{F}}{\rho})}{\sum_{1\leqslant l_{1}\leqslant l_{2}\leqslant N}\mu_{l_{1}l_{2}}+\rho}.\label{UPL}
\end{eqnarray}

Now, let's solve $\mathcal{E}$. The minimization problem of $\mathcal{E}$ is as follows:
\begin{eqnarray}
\min_{\mathcal{E}}\tau\|\mathcal{E}\|_{1}+\frac{\rho}{2}\|\mathcal{T}-\mathcal{L}-\mathcal{E}+\frac{\mathcal{F}}{\rho}\|_{F}^{2}.\label{FFEE1}
\end{eqnarray}
Problem (\ref{FFEE1}) has the following closed-form solution:
\begin{eqnarray}
\mathcal{E}^{+}=S_{\frac{\tau}{\rho}}(\mathcal{T}-\mathcal{L}+\frac{\mathcal{F}}{\rho}),\label{UPE}
\end{eqnarray}
where $S_{\lambda}(\cdot)$ is the soft thresholding operator \cite{2011733}: 
\begin{eqnarray}
S_{\lambda}(x)=\left\{\begin{array}{c}
0,\quad if\quad|x|\leqslant \lambda,
\\sign(x)(|x|-\lambda),\quad if\quad|x|> \lambda
\end{array}
\right.
\end{eqnarray}

Finally, multipliers $\mathcal{R}_{l_{1}l_{2}}$ and $\mathcal{F}$ are updated according to the following formula:
\begin{eqnarray}
\left\{\begin{array}{l}
\mathcal{R}_{l_{1}l_{2}}^{+}=\mathcal{R}_{k_{1}k_{2}}+\mu_{l_{1}l_{2}}(\mathcal{L}-\mathcal{G}_{l_{1}l_{2}});
\\\mathcal{F}^{+}=\mathcal{F}+\rho(\mathcal{T}-\mathcal{L}-\mathcal{E}).
\end{array}
\right.
\label{UPM21}\end{eqnarray}
FFM-based TPRCA model computation is given in Algorithm \ref{TPRCA}. The main per-iteration cost lies in the update of $\mathcal{G}_{l_{1}l_{2}}$, which requires computing SVD and t-SVD . The per-iteration complexity is $O(LE(\sum_{1\leqslant l_{1}< l_{2}\leqslant N}[log(le_{l_{1}l_{2}})+\min(\mathit{I}_{l_{1}},\mathit{I}_{l_{2}})]+\sum_{1\leqslant l_{1} \leqslant N}\mathit{I}_{l_{1}}))$, where $LE=\prod_{i=1}^{N}\mathit{I}_{i}$ and $le_{l_{1}l_{2}}=LE/(\mathit{I}_{l_{1}}\mathit{I}_{l_{2}})$.

\section{Numerical experiments}
We evaluate the performance of the proposed FFM-based LRTC and TRPCA methods. All methods are tested on real-world data. We employ the peak signal-to-noise rate (PSNR), the structural similarity (SSIM) \cite{1284395}, the feature similarity (FSIM) \cite{5705575}, and erreur relative globale adimensionnelle de synth$\grave{e}$se (ERGAS) \cite{2432002352} to measure the quality of the recovered results. The higher the value of PSNR, SSIM and FSIM is, the better the quality of the recovered image will be. Meanwhile, the smaller the ERGAS value is, the better the quality of the recovered image will be. All tests are implemented on the Windows 10 platform and MATLAB (R2019a) with an Intel Core i7-10875H 2.30 GHz and 32 GB of RAM.

\subsection{Low-rank tensor completion}
In this section, we test four kinds of real-world data: MSI, MRI, color video (CV), and hyperspectral video (HSV). The methodology for sampling the data is purely random sampling. The compared LRTC methods are as follows: HaLRTC \cite{6138863}, LRTCTV-I \cite{3120171}, KBRTC \cite{8000407} and ESPTC \cite{8941238}, representing state-of-the-art for the Tucker-decomposition-based method; and TNN \cite{7782758}, PSTNN \cite{1122020112680}, FTNN \cite{9115254}, WSTNN \cite{2020170}, nonconvex WSTNN \cite{7342019749}, representing state-of-the-art for the t-SVD-based method; and minmax concave plus penalty-based TC method (McpTC) \cite{2612015273} and a novel low-rank tensor completion model using smooth matrix factorization 
(SMFLRTC)\cite{6772019695}. In view of the fact that the four of them, i.e., the TNN, the PSTNN, the FTNN, and the SMFLRTC apply only to three-order tensors, in all four-order tensor tests, we first reshape the four-order tensors into three-order tensors and then test the performances of these methods.

\subsubsection{MSI completion}

\begin{table}[]
	\caption{The names and corresponding labels of individual MSI in the database CAVE.}
	\begin{tabular}{|c|c|c|c|}
		\hline
		\textbf{ID} & \textbf{MSI name}        & \textbf{ID} & \textbf{MSI name}              \\ \hline
		1           & watercolors              & 17          & fake\_and\_real\_tomatoes      \\ \hline
		2           & thread\_spools           & 18          & fake\_and\_real\_sushi         \\ \hline
		3           & superballs               & 19          & fake\_and\_real\_strawberries  \\ \hline
		4           & stuffed\_toys            & 20          & fake\_and\_real\_peppers       \\ \hline
		5           & sponges                  & 21          & fake\_and\_real\_lemons        \\ \hline
		6           & real\_and\_fake\_peppers & 22          & fake\_and\_real\_lemon\_slices \\ \hline
		7           & real\_and\_fake\_apples  & 23          & fake\_and\_real\_food          \\ \hline
		8           & pompoms                  & 24          & fake\_and\_real\_beers         \\ \hline
		9           & photo\_and\_face         & 25          & face                           \\ \hline
		10          & paints                   & 26          & egyptian\_statue               \\ \hline
		11          & oil\_painting            & 27          & cloth                          \\ \hline
		12          & jelly\_beans             & 28          & clay                           \\ \hline
		13          & hairs                    & 29          & chart\_and\_stuffed\_toy       \\ \hline
		14          & glass\_tiles             & 30          & beads                          \\ \hline
		15          & flowers                  & 31          & balloons                       \\ \hline
		16          & feathers                 & 32          & cd                             \\ \hline
	\end{tabular}
\label{MSINAME}
\end{table}
\begin{figure*}[!h] 
	\centering  
	\vspace{0cm} 
	\subfloat[]{
		\begin{minipage}[b]{0.06\linewidth}
			\includegraphics[width=1\linewidth]{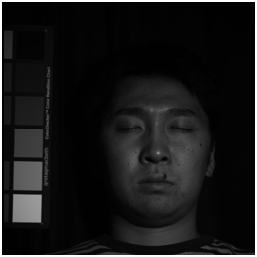}
			\includegraphics[width=1\linewidth]{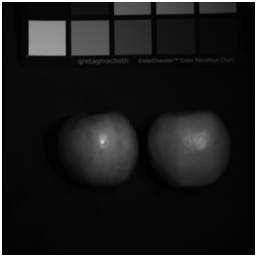}
			\includegraphics[width=1\linewidth]{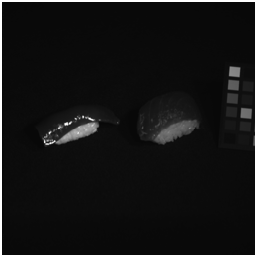}
			\includegraphics[width=1\linewidth]{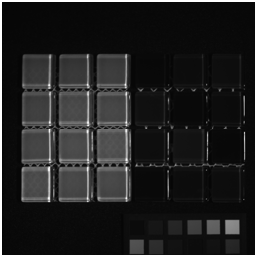}
			\includegraphics[width=1\linewidth]{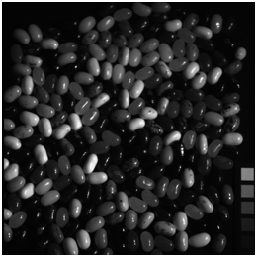}
			\includegraphics[width=1\linewidth]{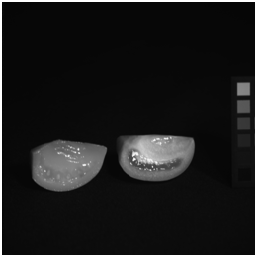}
			\includegraphics[width=1\linewidth]{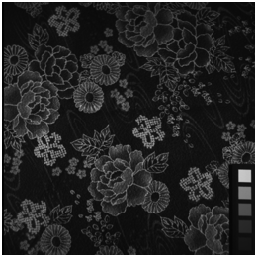}
			\includegraphics[width=1\linewidth]{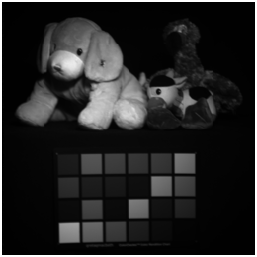}
			\includegraphics[width=1\linewidth]{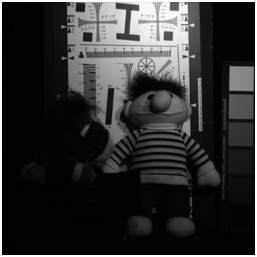}
	\end{minipage}}
	\subfloat[]{
		\begin{minipage}[b]{0.06\linewidth}
			\includegraphics[width=1\linewidth]{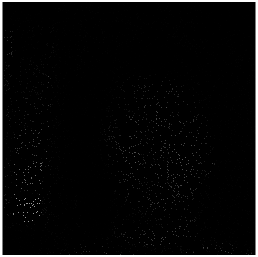}
			\includegraphics[width=1\linewidth]{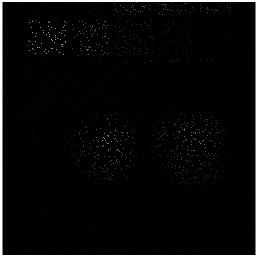}
			\includegraphics[width=1\linewidth]{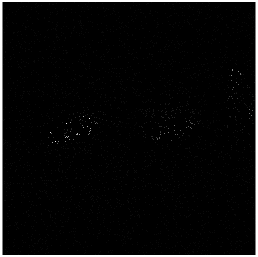}
			\includegraphics[width=1\linewidth]{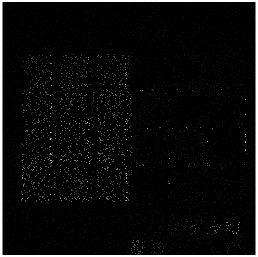}
			\includegraphics[width=1\linewidth]{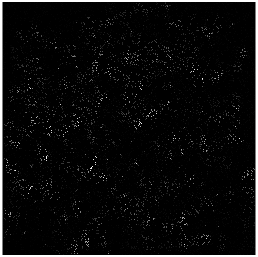}
			\includegraphics[width=1\linewidth]{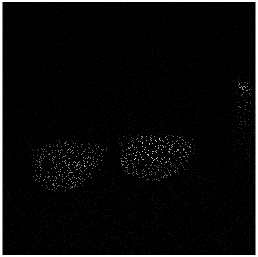}
			\includegraphics[width=1\linewidth]{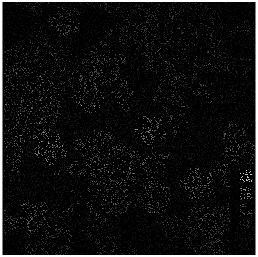}
			\includegraphics[width=1\linewidth]{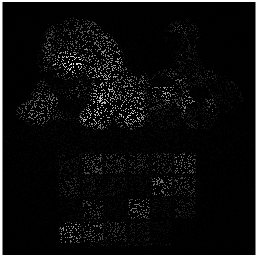}
			\includegraphics[width=1\linewidth]{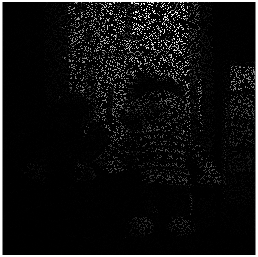}
	\end{minipage}}
	\subfloat[]{
		\begin{minipage}[b]{0.06\linewidth}
			\includegraphics[width=1\linewidth]{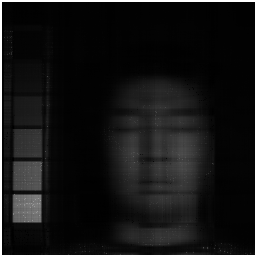}
			\includegraphics[width=1\linewidth]{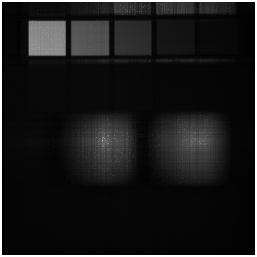}
			\includegraphics[width=1\linewidth]{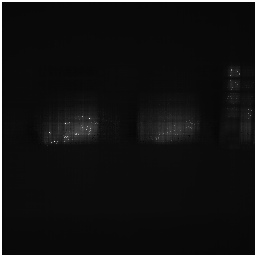}
			\includegraphics[width=1\linewidth]{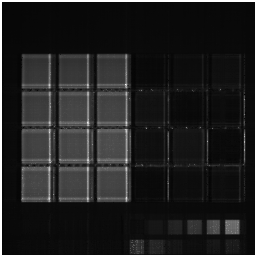}
			\includegraphics[width=1\linewidth]{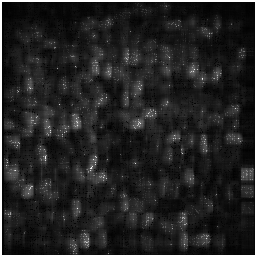}
			\includegraphics[width=1\linewidth]{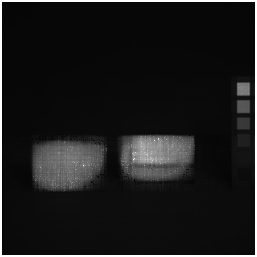}
			\includegraphics[width=1\linewidth]{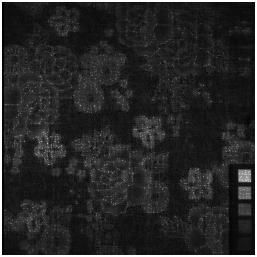}
			\includegraphics[width=1\linewidth]{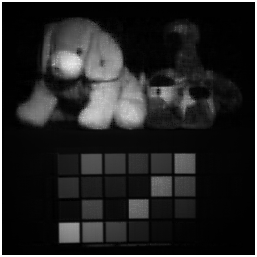}
			\includegraphics[width=1\linewidth]{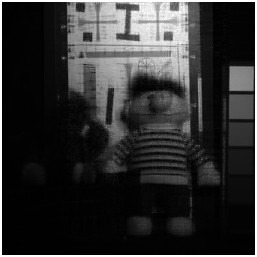}
	\end{minipage}}
	\subfloat[]{
		\begin{minipage}[b]{0.06\linewidth}
			\includegraphics[width=1\linewidth]{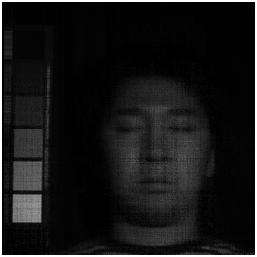}
			\includegraphics[width=1\linewidth]{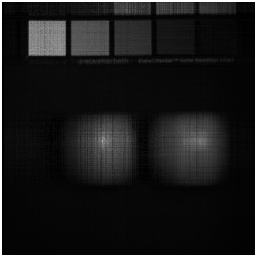}
			\includegraphics[width=1\linewidth]{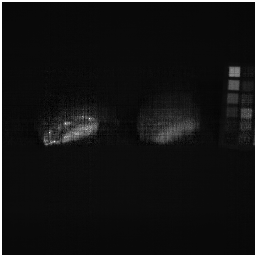}
			\includegraphics[width=1\linewidth]{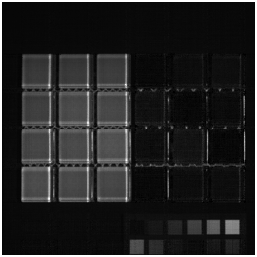}
			\includegraphics[width=1\linewidth]{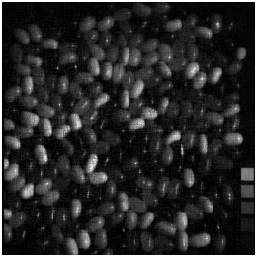}
			\includegraphics[width=1\linewidth]{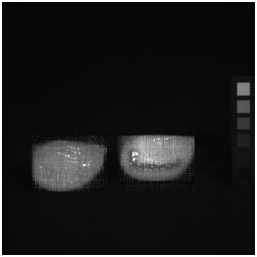}
			\includegraphics[width=1\linewidth]{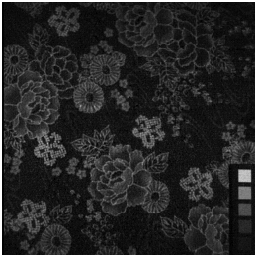}
			\includegraphics[width=1\linewidth]{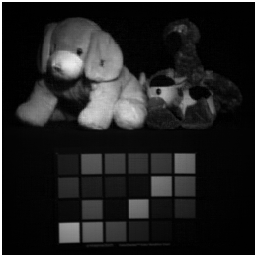}
			\includegraphics[width=1\linewidth]{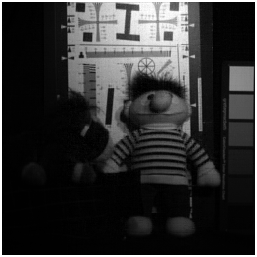}
	\end{minipage}}
	\subfloat[]{
		\begin{minipage}[b]{0.06\linewidth}
			\includegraphics[width=1\linewidth]{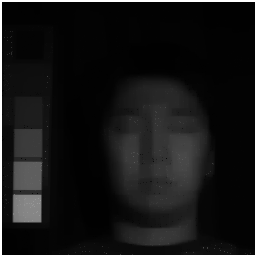}
			\includegraphics[width=1\linewidth]{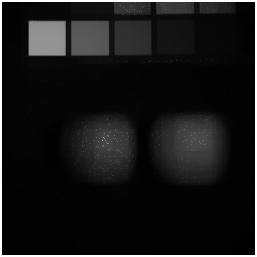}
			\includegraphics[width=1\linewidth]{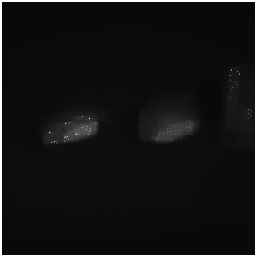}
			\includegraphics[width=1\linewidth]{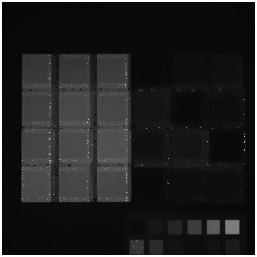}
			\includegraphics[width=1\linewidth]{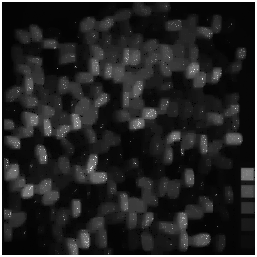}
			\includegraphics[width=1\linewidth]{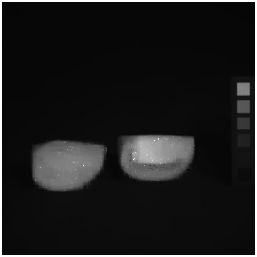}
			\includegraphics[width=1\linewidth]{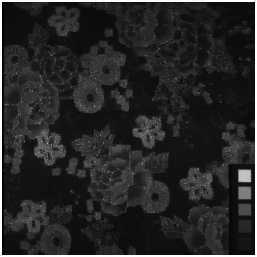}
			\includegraphics[width=1\linewidth]{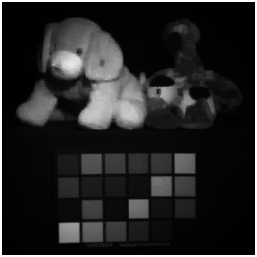}
			\includegraphics[width=1\linewidth]{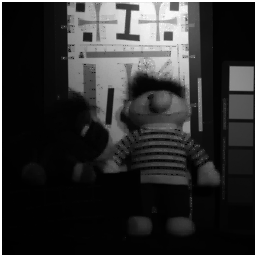}
	\end{minipage}}
	\subfloat[]{
		\begin{minipage}[b]{0.06\linewidth}
			\includegraphics[width=1\linewidth]{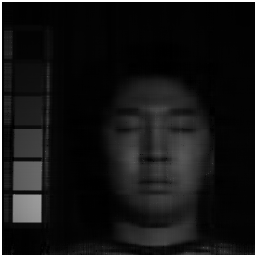}
			\includegraphics[width=1\linewidth]{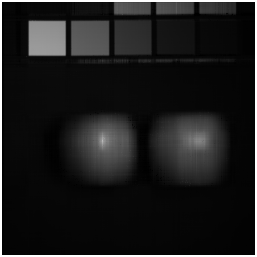}
			\includegraphics[width=1\linewidth]{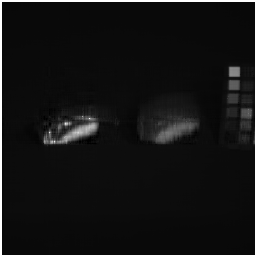}
			\includegraphics[width=1\linewidth]{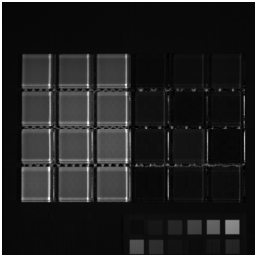}
			\includegraphics[width=1\linewidth]{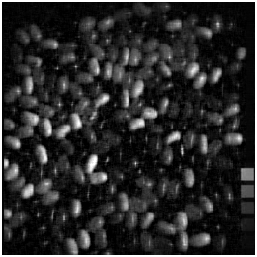}
			\includegraphics[width=1\linewidth]{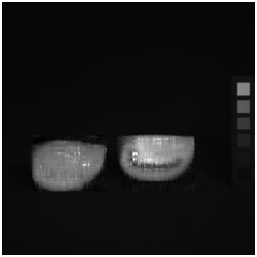}
			\includegraphics[width=1\linewidth]{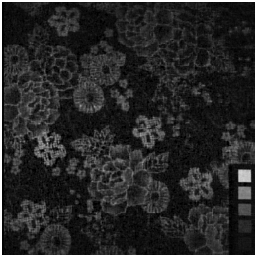}
			\includegraphics[width=1\linewidth]{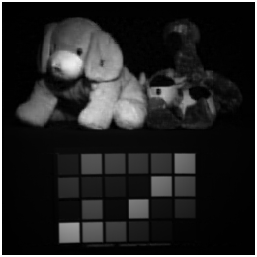}
			\includegraphics[width=1\linewidth]{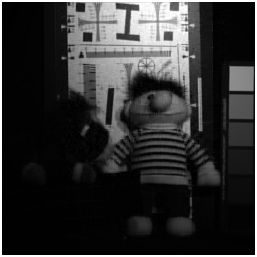}
	\end{minipage}}
	\subfloat[]{
		\begin{minipage}[b]{0.06\linewidth}
			\includegraphics[width=1\linewidth]{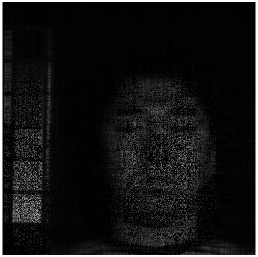}
			\includegraphics[width=1\linewidth]{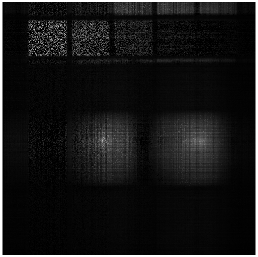}
			\includegraphics[width=1\linewidth]{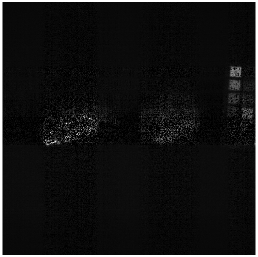}
			\includegraphics[width=1\linewidth]{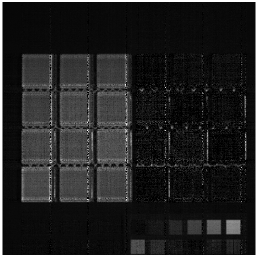}
			\includegraphics[width=1\linewidth]{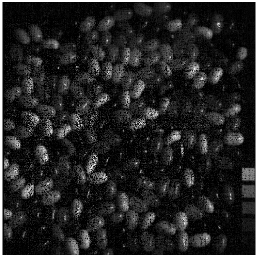}
			\includegraphics[width=1\linewidth]{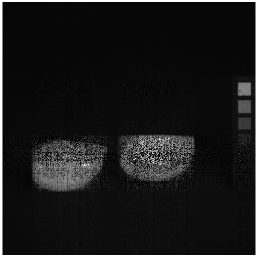}
			\includegraphics[width=1\linewidth]{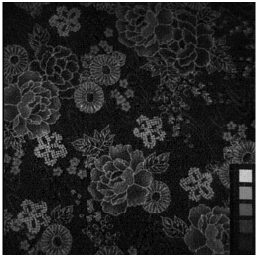}
			\includegraphics[width=1\linewidth]{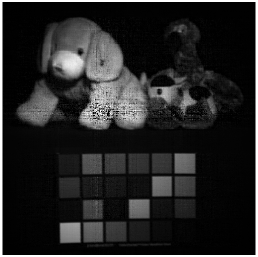}
			\includegraphics[width=1\linewidth]{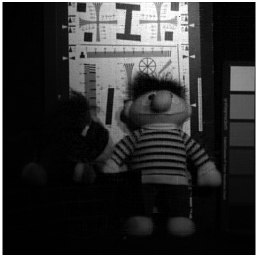}
	\end{minipage}}
	\subfloat[]{
		\begin{minipage}[b]{0.06\linewidth}
			\includegraphics[width=1\linewidth]{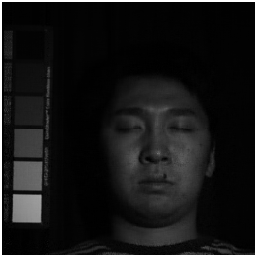}
			\includegraphics[width=1\linewidth]{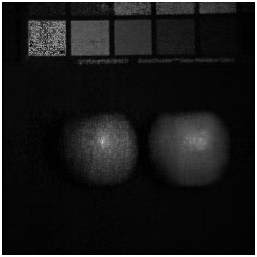}
			\includegraphics[width=1\linewidth]{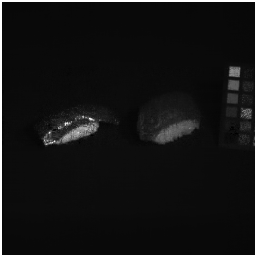}
			\includegraphics[width=1\linewidth]{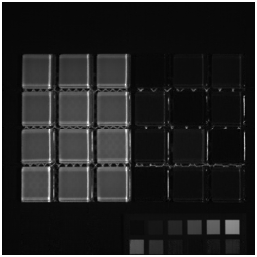}
			\includegraphics[width=1\linewidth]{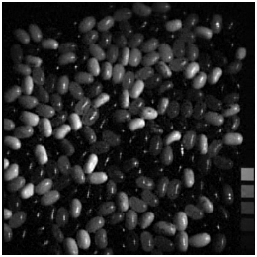}
			\includegraphics[width=1\linewidth]{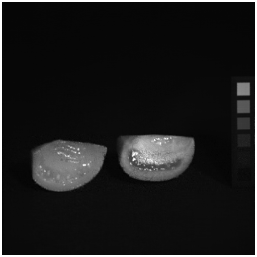}
			\includegraphics[width=1\linewidth]{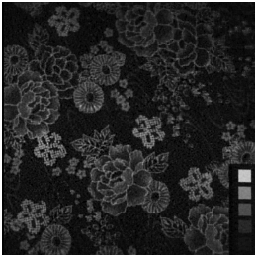}
			\includegraphics[width=1\linewidth]{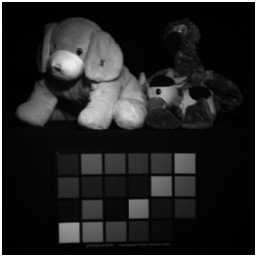}
			\includegraphics[width=1\linewidth]{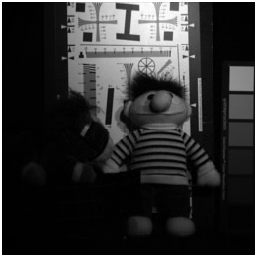}
	\end{minipage}}
	\subfloat[]{
		\begin{minipage}[b]{0.06\linewidth}
			\includegraphics[width=1\linewidth]{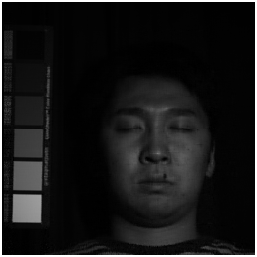}
			\includegraphics[width=1\linewidth]{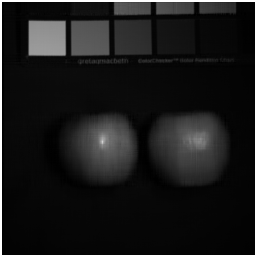}
			\includegraphics[width=1\linewidth]{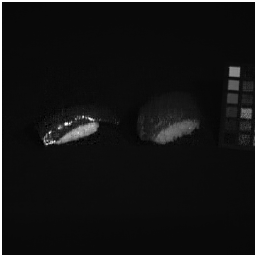}
			\includegraphics[width=1\linewidth]{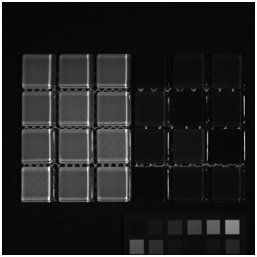}
			\includegraphics[width=1\linewidth]{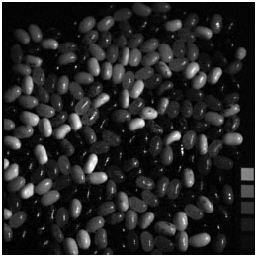}
			\includegraphics[width=1\linewidth]{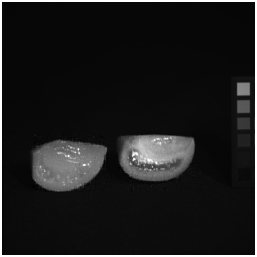}
			\includegraphics[width=1\linewidth]{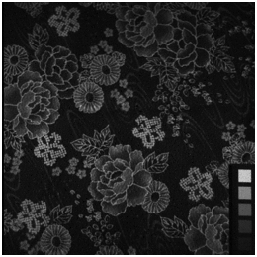}
			\includegraphics[width=1\linewidth]{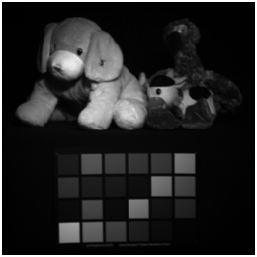}
			\includegraphics[width=1\linewidth]{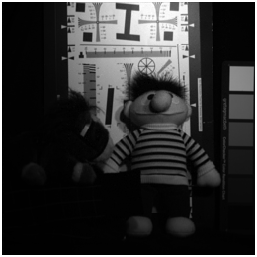}
	\end{minipage}}
	\subfloat[]{
		\begin{minipage}[b]{0.06\linewidth}
			\includegraphics[width=1\linewidth]{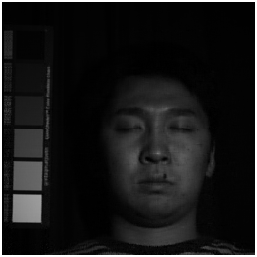}
			\includegraphics[width=1\linewidth]{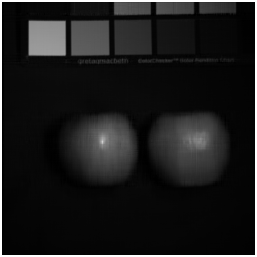}
			\includegraphics[width=1\linewidth]{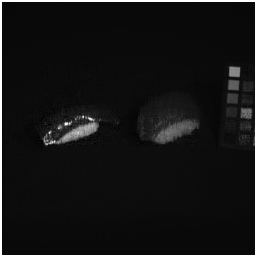}
			\includegraphics[width=1\linewidth]{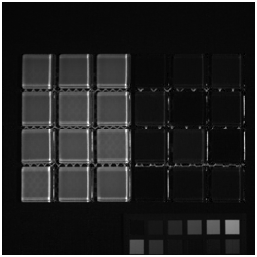}
			\includegraphics[width=1\linewidth]{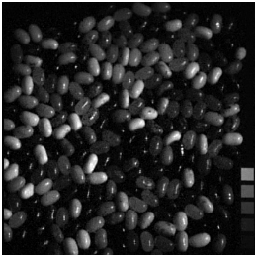}
			\includegraphics[width=1\linewidth]{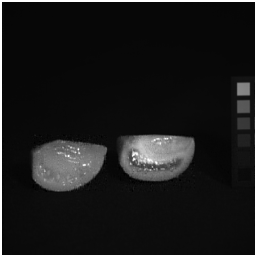}
			\includegraphics[width=1\linewidth]{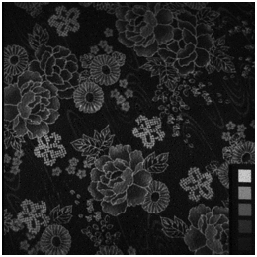}
			\includegraphics[width=1\linewidth]{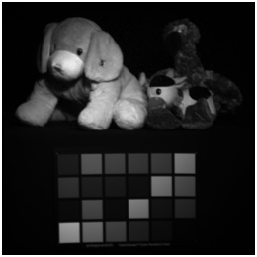}
			\includegraphics[width=1\linewidth]{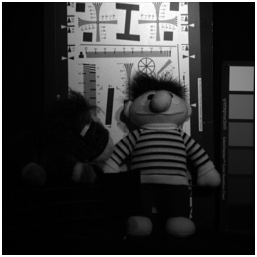}
	\end{minipage}}
	\subfloat[]{
		\begin{minipage}[b]{0.06\linewidth}
			\includegraphics[width=1\linewidth]{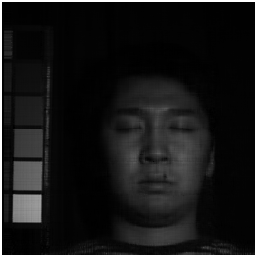}
			\includegraphics[width=1\linewidth]{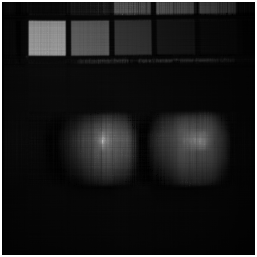}
			\includegraphics[width=1\linewidth]{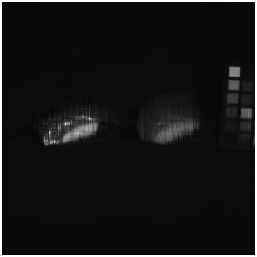}
			\includegraphics[width=1\linewidth]{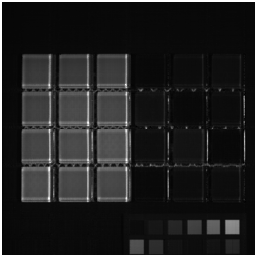}
			\includegraphics[width=1\linewidth]{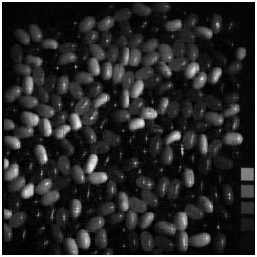}
			\includegraphics[width=1\linewidth]{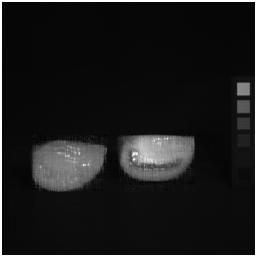}
			\includegraphics[width=1\linewidth]{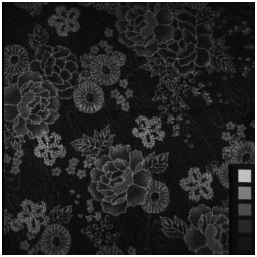}
			\includegraphics[width=1\linewidth]{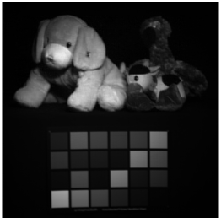}
			\includegraphics[width=1\linewidth]{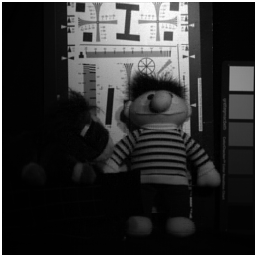}
	\end{minipage}}
	\subfloat[]{
		\begin{minipage}[b]{0.06\linewidth}
			\includegraphics[width=1\linewidth]{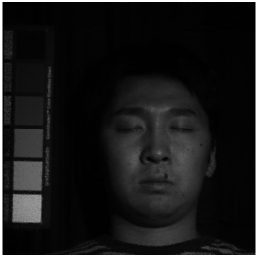}
			\includegraphics[width=1\linewidth]{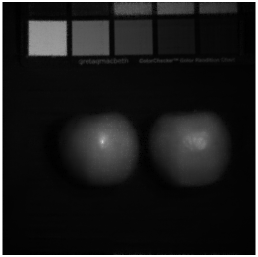}
			\includegraphics[width=1\linewidth]{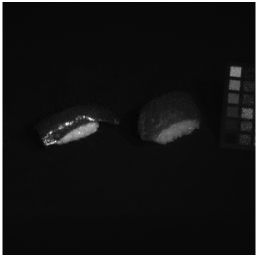}
			\includegraphics[width=1\linewidth]{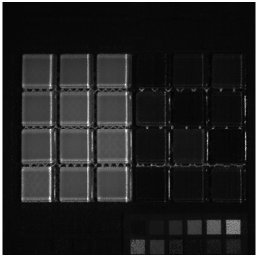}
			\includegraphics[width=1\linewidth]{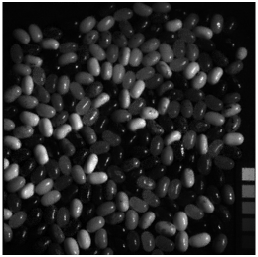}
			\includegraphics[width=1\linewidth]{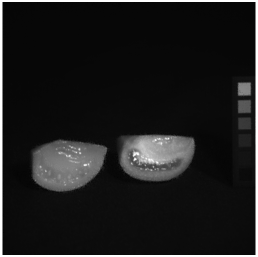}
			\includegraphics[width=1\linewidth]{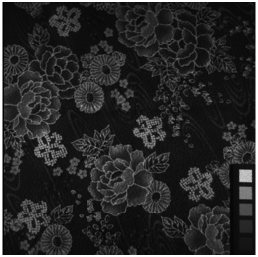}
			\includegraphics[width=1\linewidth]{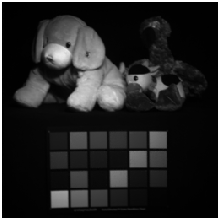}
			\includegraphics[width=1\linewidth]{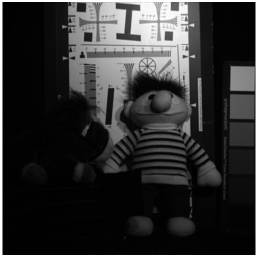}
	\end{minipage}}
	\subfloat[]{
		\begin{minipage}[b]{0.06\linewidth}
			\includegraphics[width=1\linewidth]{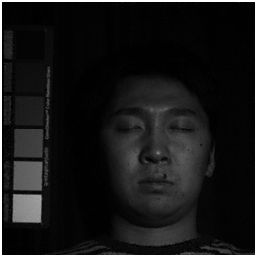}
			\includegraphics[width=1\linewidth]{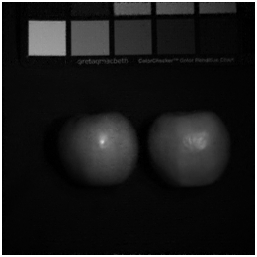}
			\includegraphics[width=1\linewidth]{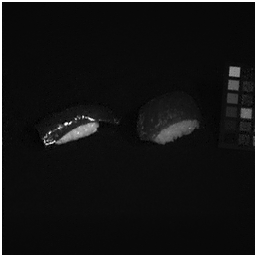}
			\includegraphics[width=1\linewidth]{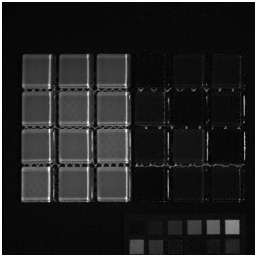}
			\includegraphics[width=1\linewidth]{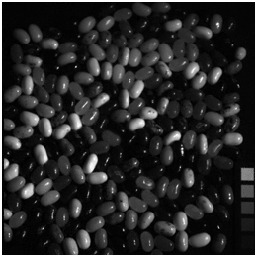}
			\includegraphics[width=1\linewidth]{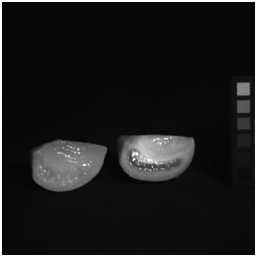}
			\includegraphics[width=1\linewidth]{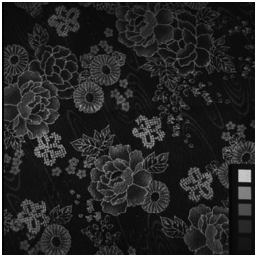}
			\includegraphics[width=1\linewidth]{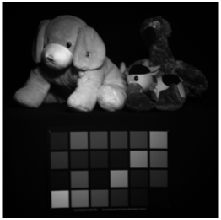}
			\includegraphics[width=1\linewidth]{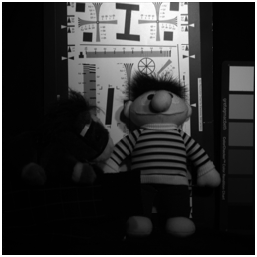}
	\end{minipage}}
	\subfloat[]{
		\begin{minipage}[b]{0.06\linewidth}
			\includegraphics[width=1\linewidth]{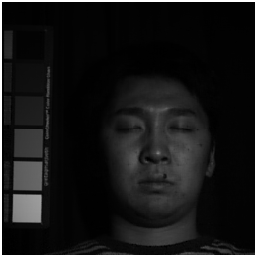}
			\includegraphics[width=1\linewidth]{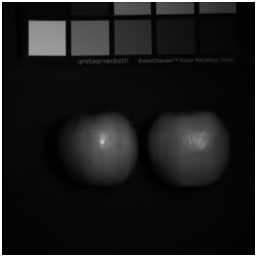}
			\includegraphics[width=1\linewidth]{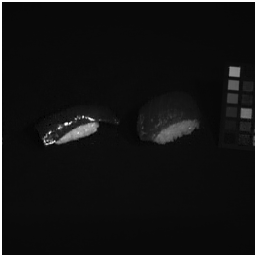}
			\includegraphics[width=1\linewidth]{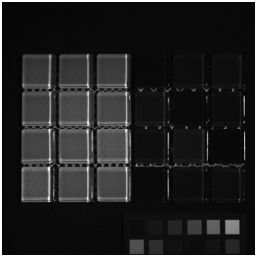}
			\includegraphics[width=1\linewidth]{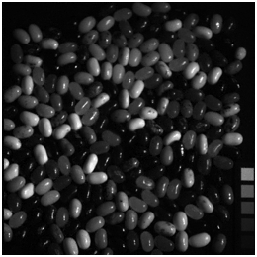}
			\includegraphics[width=1\linewidth]{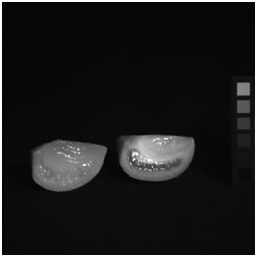}
			\includegraphics[width=1\linewidth]{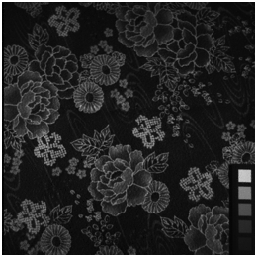}
			\includegraphics[width=1\linewidth]{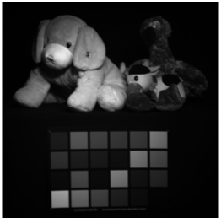}
			\includegraphics[width=1\linewidth]{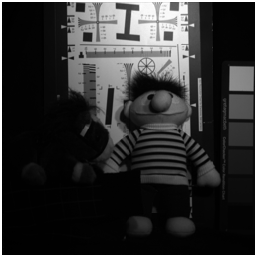}
	\end{minipage}}
	\caption{(a) Original image. (b) Obeserved image. (c) HaLRTC. (d) TNN. (e) LRTCTV-I. (f) McpTC. (g) PSTNN. (h) SMFLRTC. (i) KBRTC. (j) ESPTC. (k) FTNN. (l) WSTNN. (m) NWSTNN. (n) FFMTC. SR: top 3 rows is 5\%, middle 3 rows is 10\% and last 3 rows is 20\%. The rows of MSIs are in order: face, real\_and\_fake\_apples, fake\_and\_real\_sushi, glass\_tiles, jelly\_beans, fake\_and\_real\_tomatoes, cloth, stuffed\_toys, chart\_and\_stuffed\_toy. The corresponding bands in each row are: 10, 30, 20, 8, 18, 28, 5, 15, 25.}
	\label{MSITC}
\end{figure*}
\begin{table*}[]
	\caption{The average PSNR, SSIM, FSIM and ERGAS values for 32 MSIs tested by observed and the twelve utilized LRTC methods.}
	\resizebox{\textwidth}{!}{
		\begin{tabular}{cccccccccccccc}
			\hline
			SR        & \multicolumn{4}{c}{5\%}                                             & \multicolumn{4}{c}{10\%}                                            & \multicolumn{4}{c}{20\%}                                            & Time(s)          \\
			Method    & PSNR            & SSIM           & FSIM           & ERGAS           & PSNR            & SSIM           & FSIM           & ERGAS           & PSNR            & SSIM           & FSIM           & ERGAS           &                  \\ \hline
			Observed  & 15.438          & 0.153          & 0.644          & 845.374         & 15.672          & 0.194          & 0.646          & 822.902         & 16.185          & 0.269          & 0.651          & 775.704         & 0.000            \\
			HaLRTC    & 25.347          & 0.774          & 0.837          & 299.279         & 29.821          & 0.856          & 0.894          & 185.726         & 35.048          & 0.930          & 0.947          & 105.188         & 10.223           \\
			TNN       & 25.328          & 0.713          & 0.817          & 290.398         & 33.095          & 0.879          & 0.918          & 128.242         & 40.229          & 0.964          & 0.972          & 58.755          & 30.425           \\
			LRTCTV-I & 25.867          & 0.800          & 0.835          & 277.506         & 30.699          & 0.890          & 0.906          & 163.071         & 35.542          & 0.949          & 0.957          & 94.078          & 180.325          \\
			McpTC     & 32.476          & 0.875          & 0.909          & 133.036         & 35.974          & 0.925          & 0.943          & 91.923          & 40.522          & 0.964          & 0.971          & 56.109          & 186.442          \\
			PSTNN     & 18.710          & 0.474          & 0.650          & 575.015         & 23.186          & 0.682          & 0.782          & 353.671         & 34.388          & 0.924          & 0.941          & 115.934         & 32.702           \\
			SMFLRTC   & 31.503          & 0.887          & 0.918          & 146.535         & 39.284          & 0.964          & 0.972          & 60.816          & 44.031          & 0.980          & 0.985          & 37.596          & 533.010          \\
			KBRTC     & 36.261          & 0.930          & 0.948          & 88.631          & 42.638          & 0.977          & 0.982          & 42.779          & 48.639          & 0.993          & 0.995          & 21.691          & 113.016          \\
			ESPTC     & 36.170          & 0.925          & 0.946          & 90.600          & 41.763          & 0.971          & 0.977          & 47.085          & 47.249          & 0.989          & 0.992          & 25.717          & 190.471          \\
			FTNN      & 32.583          & 0.898          & 0.923          & 132.496         & 37.187          & 0.954          & 0.963          & 78.753          & 43.080          & 0.984          & 0.987          & 41.455          & 183.045          \\
			WSTNN     & 34.796          & 0.947          & 0.952          & 96.183          & 40.120          & 0.981          & 0.981          & 53.209          & 47.124          & 0.995          & 0.995          & 24.774          & 48.983           \\
			NWSTNN    & 37.395          & 0.945          & 0.950          & 71.581          & 43.711          & 0.985          & 0.985          & 35.748          & 51.369          & 0.997          & 0.997          & 15.519          & 85.176           \\
			FFMTC     & \textbf{40.350} & \textbf{0.971} & \textbf{0.976} & \textbf{54.134} & \textbf{45.860} & \textbf{0.990} & \textbf{0.991} & \textbf{29.272} & \textbf{52.386} & \textbf{0.998} & \textbf{0.998} & \textbf{13.990} & \textbf{107.509} \\ \hline
		\end{tabular}%
	}\label{MSITC1}
\end{table*}
\begin{figure*}[!h] 
	\centering  
	\vspace{0cm} 
	\subfloat[SR:5\%]{
		\begin{minipage}[b]{0.3\linewidth}
			\includegraphics[width=1\linewidth]{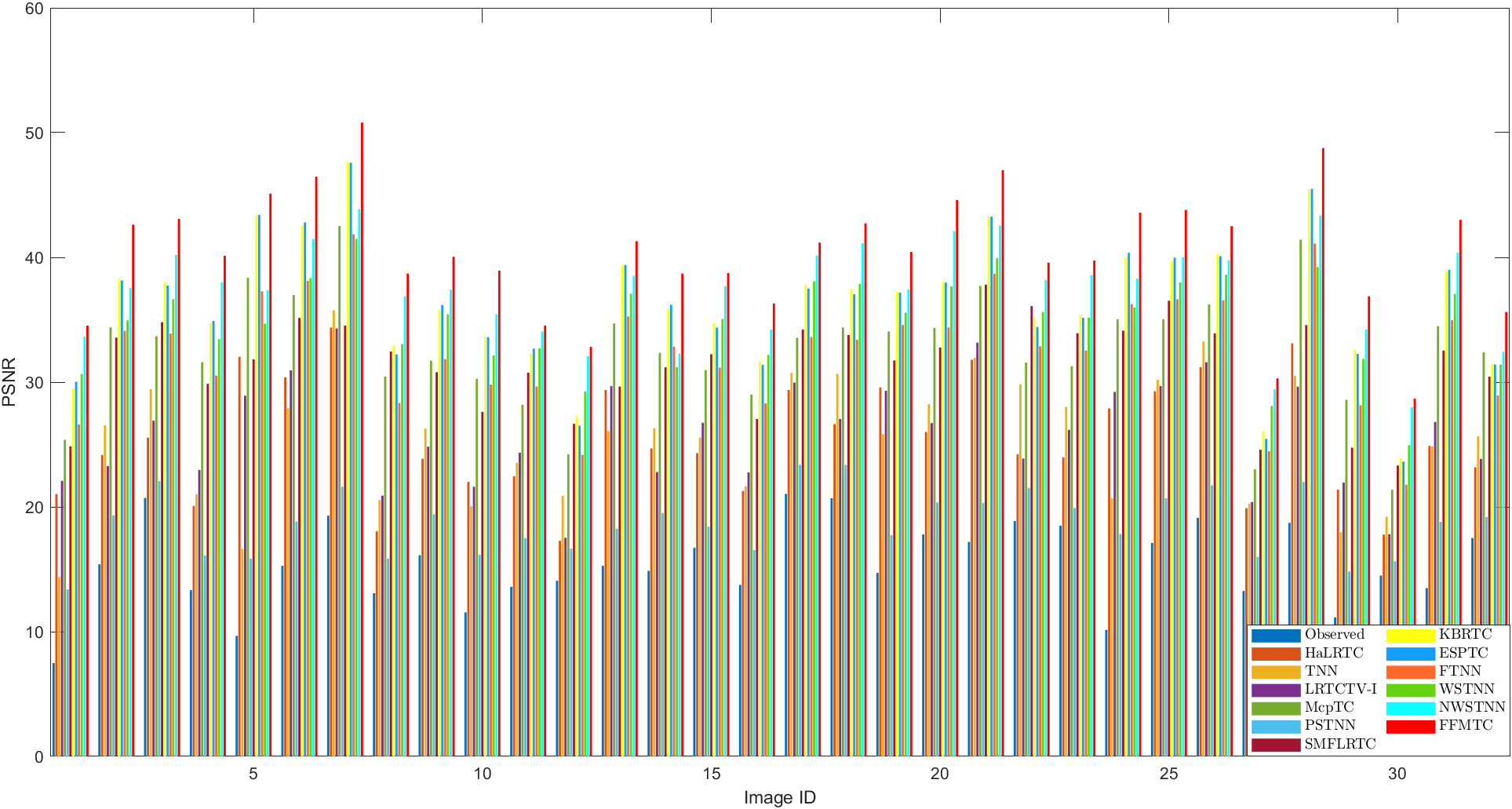}
			\includegraphics[width=1\linewidth]{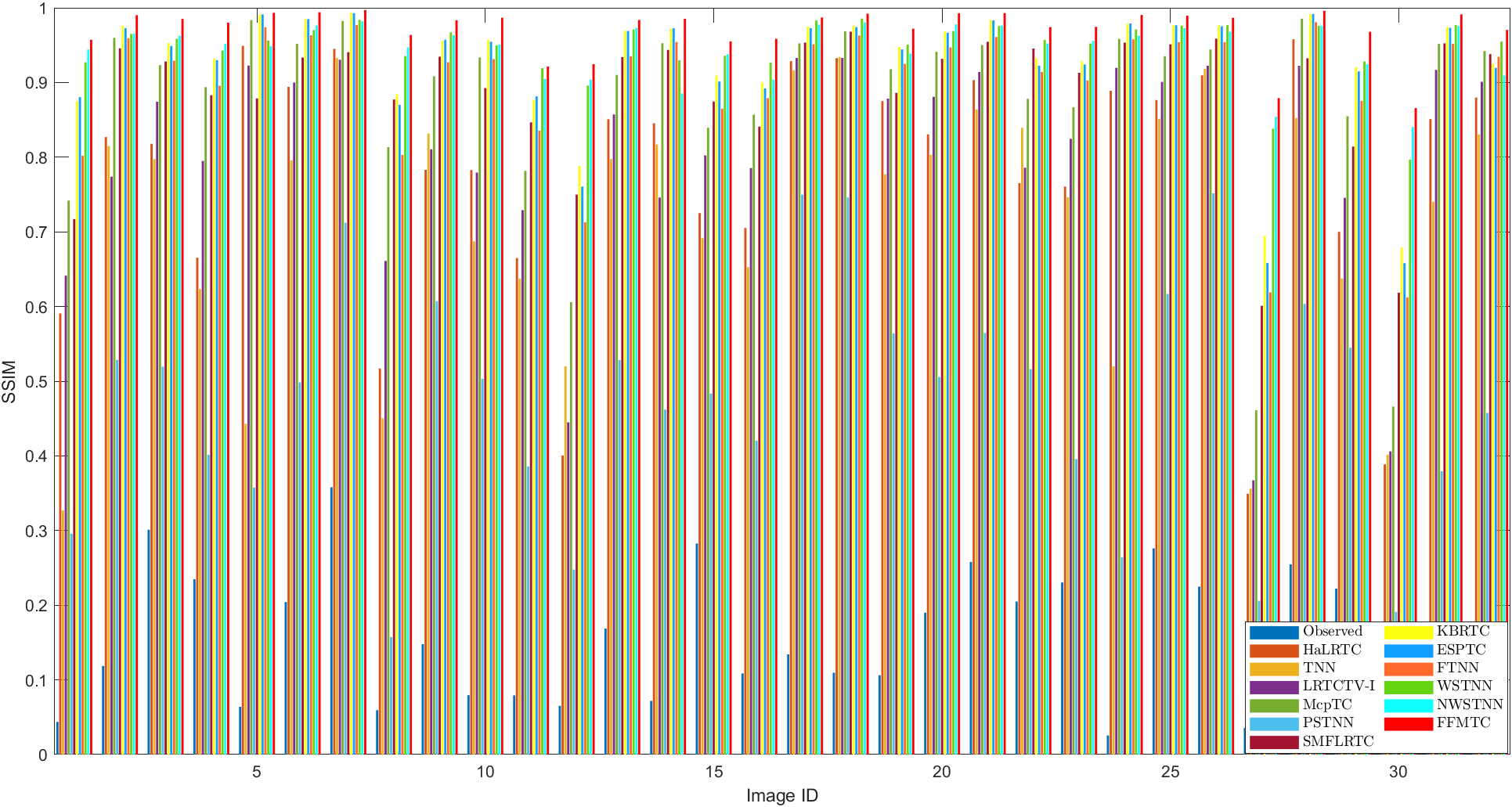}
			\includegraphics[width=1\linewidth]{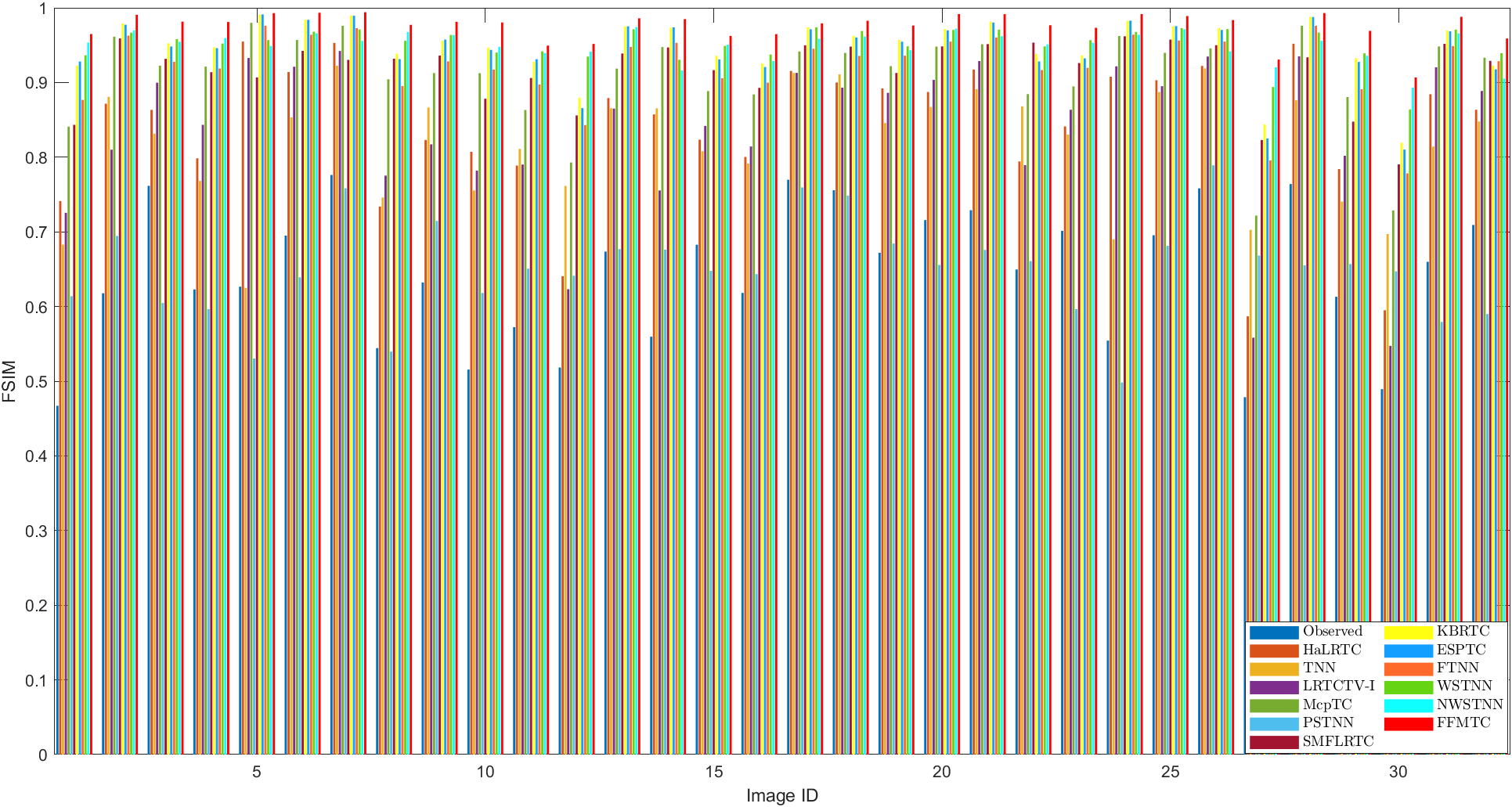}
	\end{minipage}}
	\subfloat[SR:10\%]{
		\begin{minipage}[b]{0.3\linewidth}
			\includegraphics[width=1\linewidth]{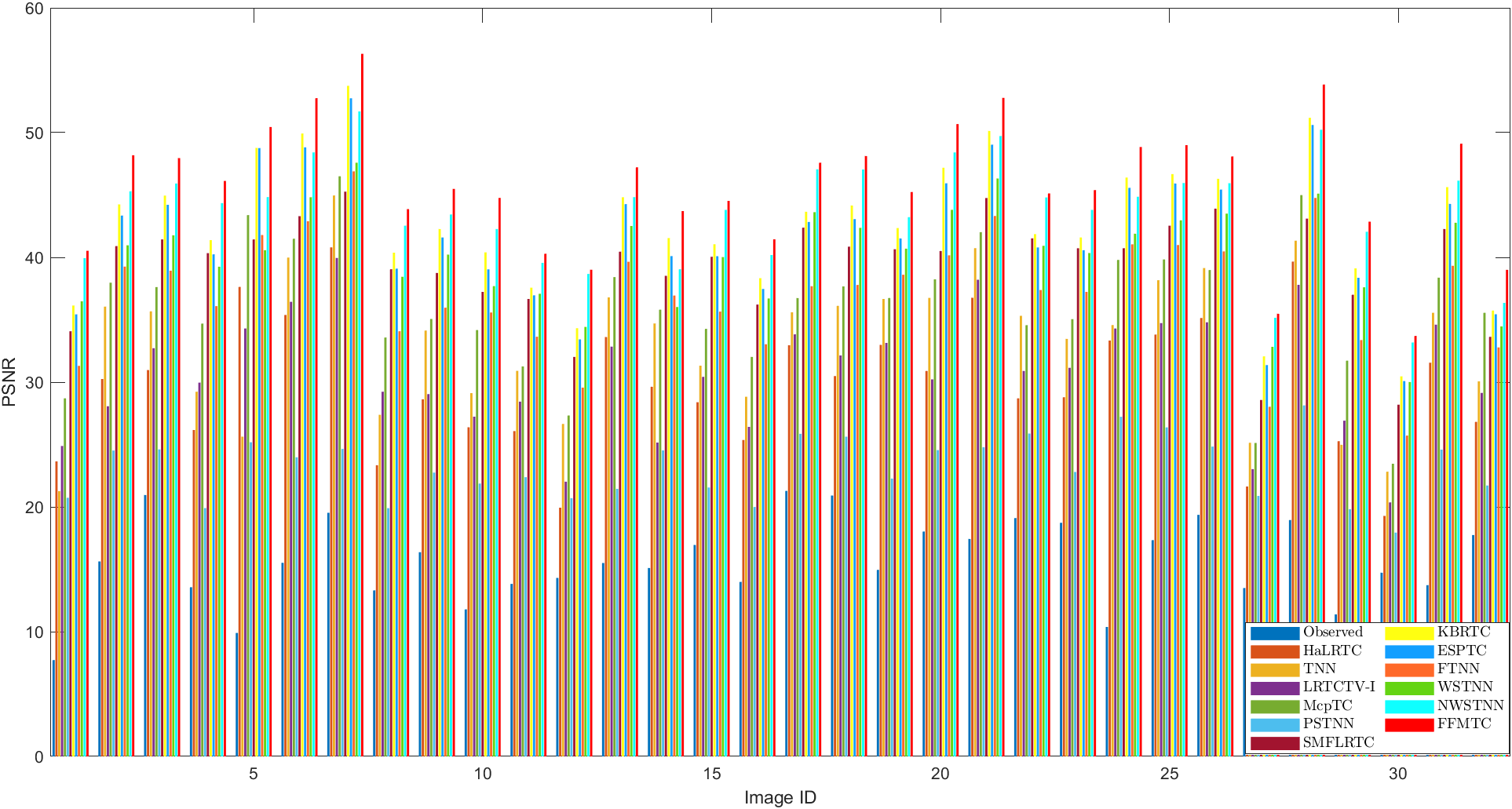}
			\includegraphics[width=1\linewidth]{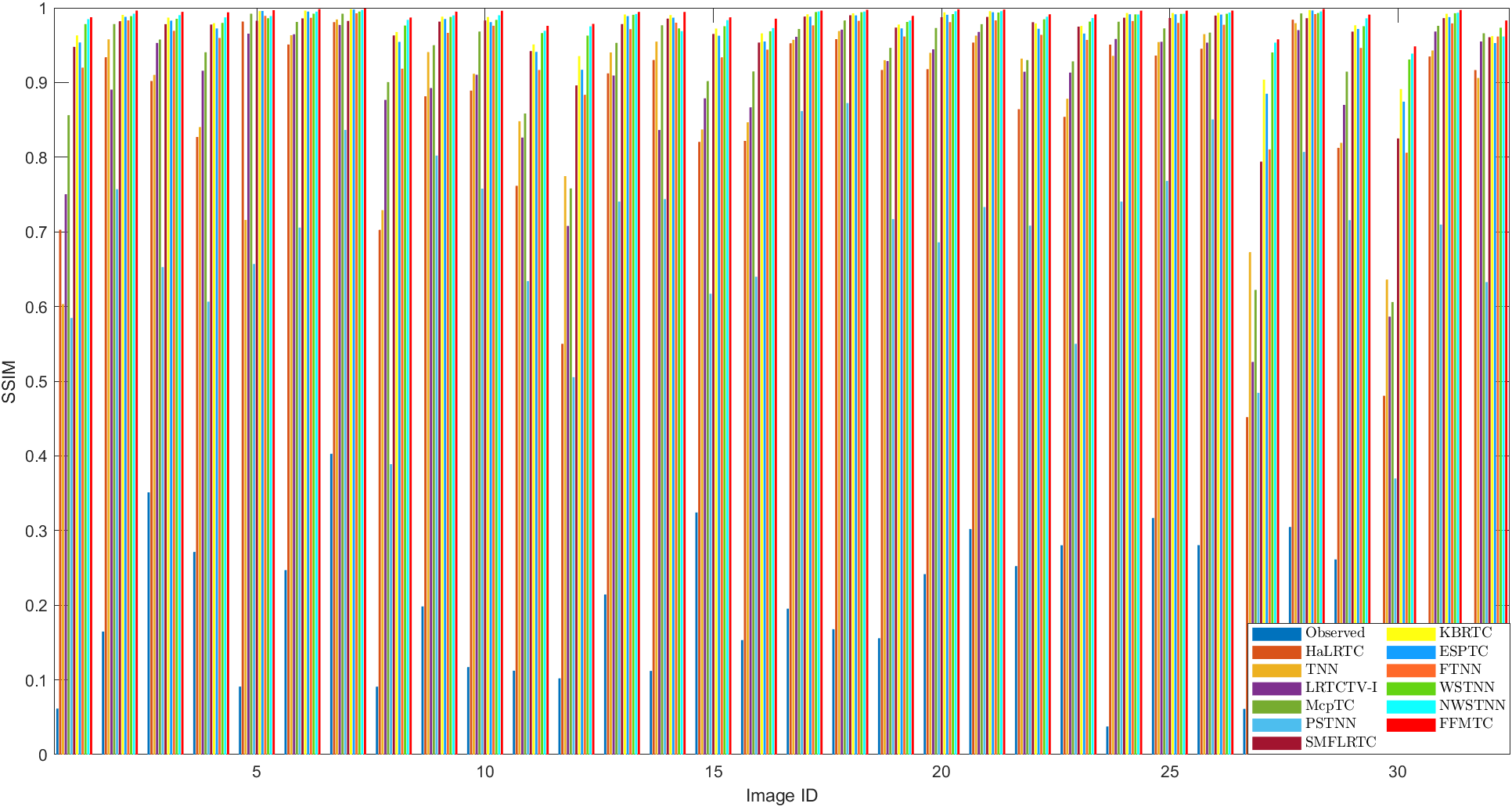}
			\includegraphics[width=1\linewidth]{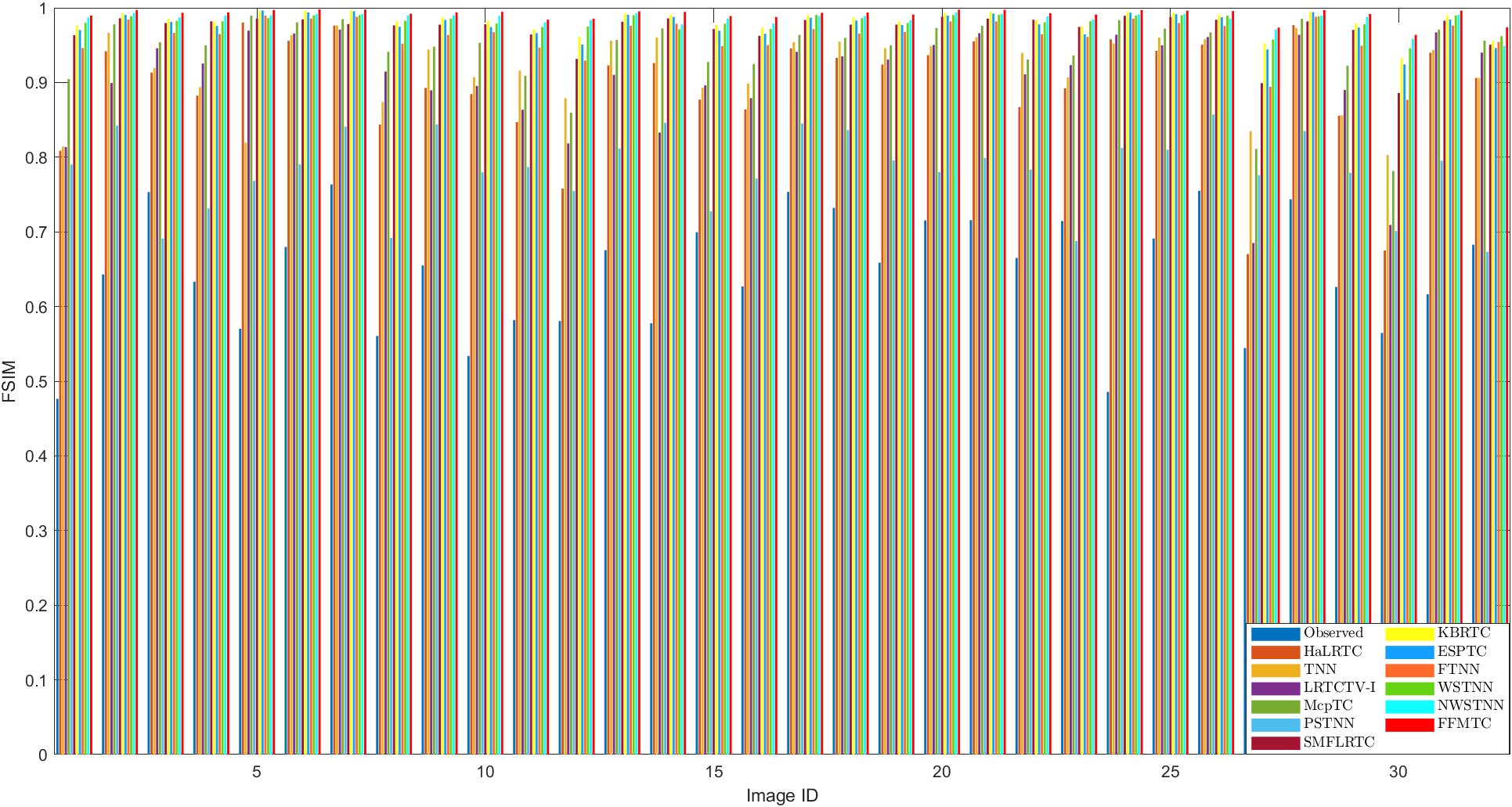}
	\end{minipage}}
	\subfloat[SR:20\%]{
		\begin{minipage}[b]{0.3\linewidth}
			\includegraphics[width=1\linewidth]{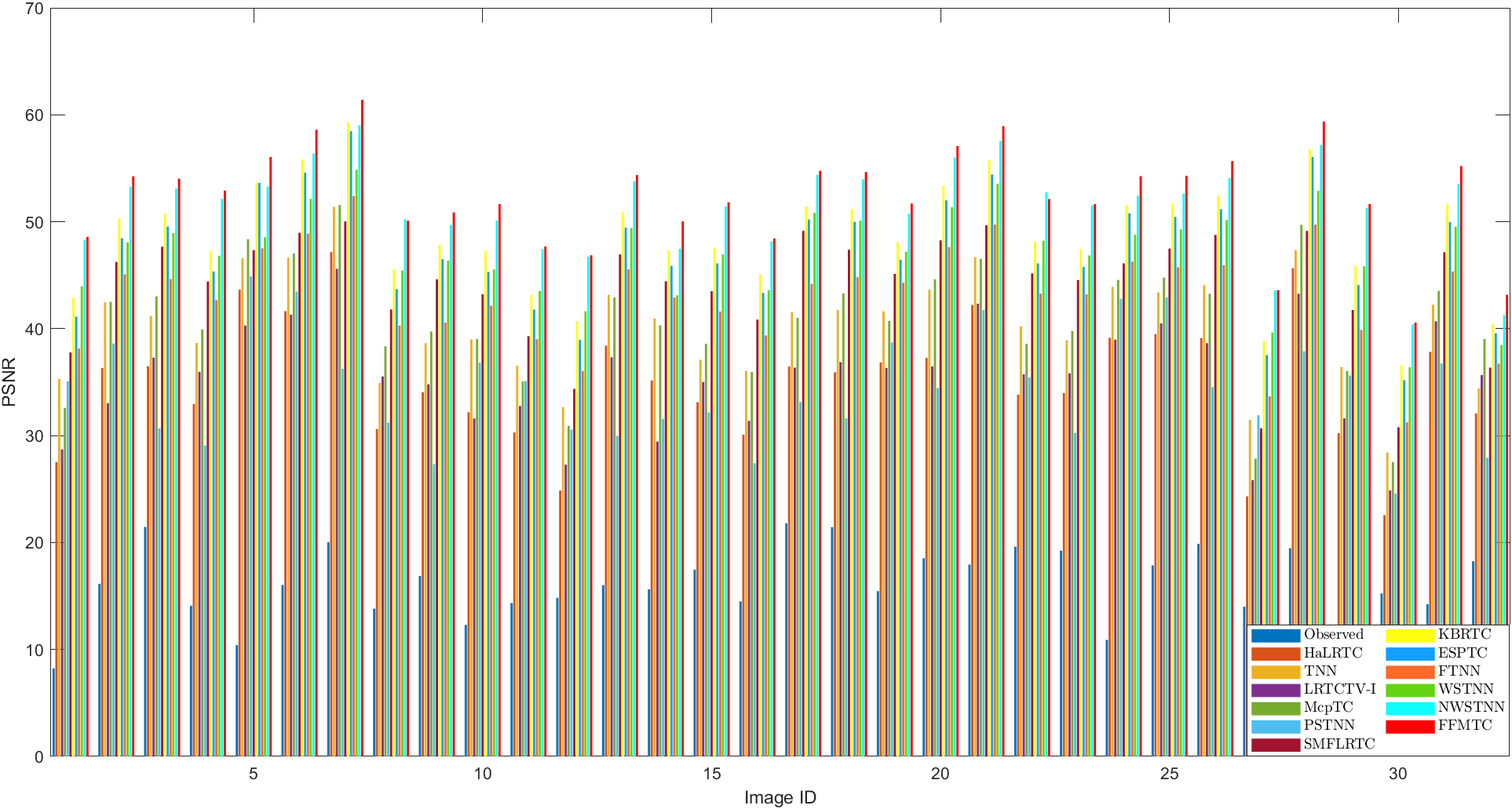}
			\includegraphics[width=1\linewidth]{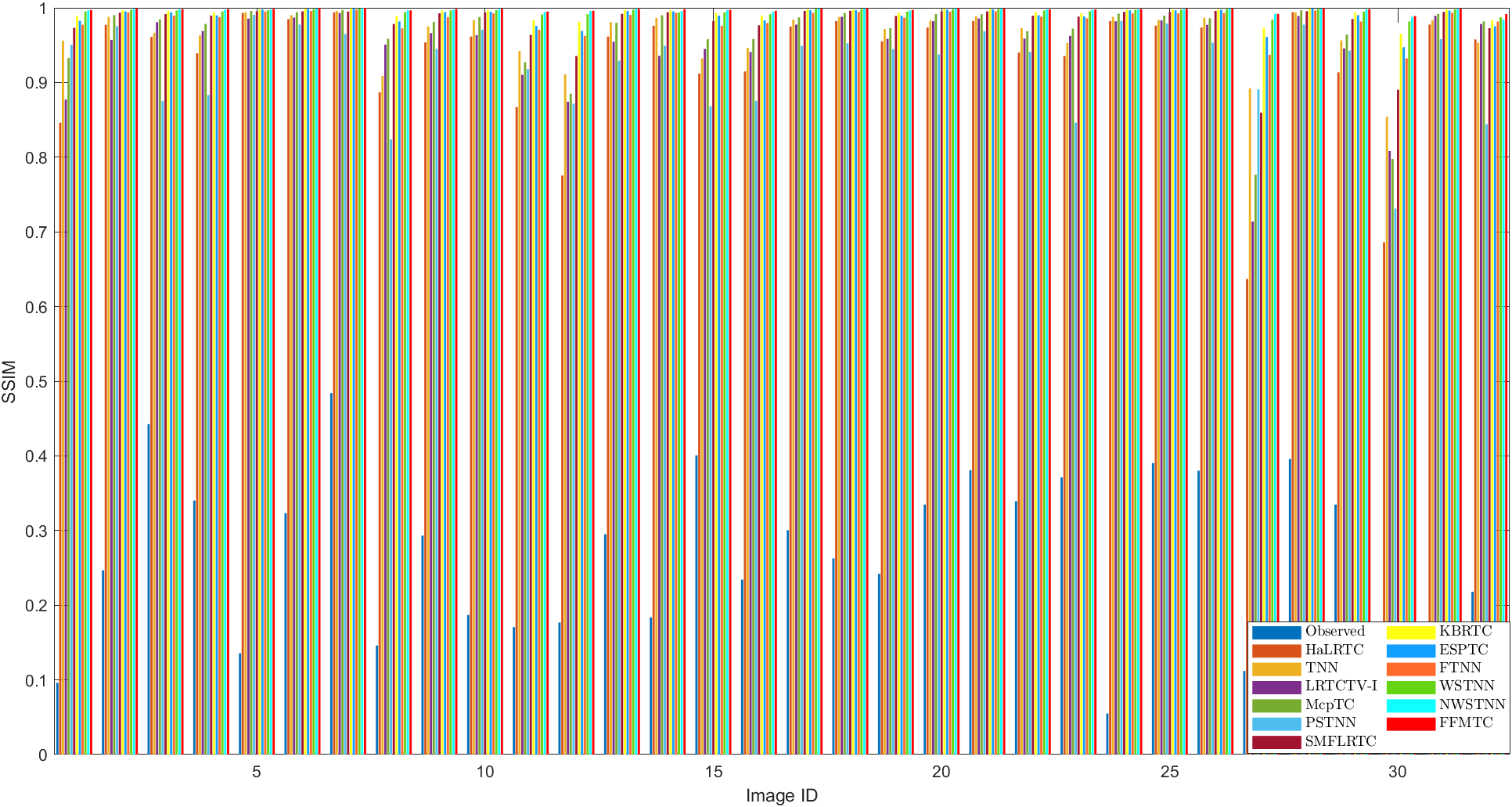}
			\includegraphics[width=1\linewidth]{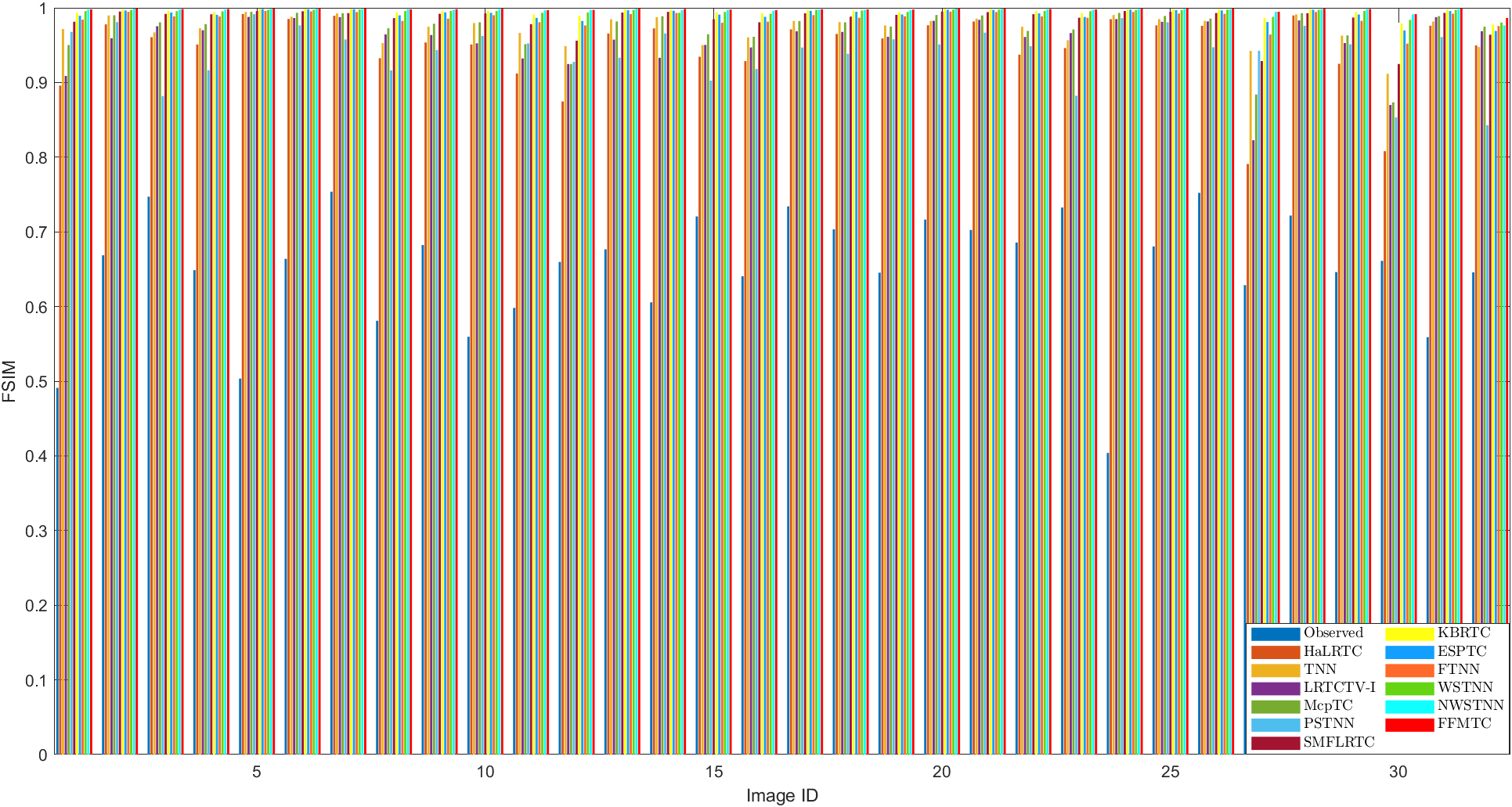}
	\end{minipage}}
	\caption{Comparison of the PSNR values (top), SSIM values (middle) and FSIM values (bottom) obtained by FFMTC and the state-of-the-art methods on 32 MSIs. SR: (a)5\%, (b)10\%, (c)20\%.}
	\label{MSITCQUANTI}
\end{figure*}
We test 32 MSIs in the dataset CAVE\footnote{http://www.cs.columbia.edu/CAVE/databases/multispectral/}. All testing data are of size $256\times256\times31$. First, we label 32 MSI in the database, and their detailed names and corresponding numbers are listed in Table \ref{MSINAME}. Then, we randomly select 9 of the 32 MSI and show their visual results at different sampling rates and different bands in Fig.\ref{MSITC}. The names of the individual MSI and their corresponding bands are reported in the caption of Fig.\ref{MSITC}. As can be seen from the Fig.\ref{MSITC}, our method is significantly superior to other methods in terms of visual results of different MSI at different sampling rates. It is especially worth mentioning that our method can still achieve excellent results even at a low sampling rate. To further highlight the superiority of our method, we list the average quantitative results of 32 MSIs in Table \ref{MSITC1}. It is not difficult to find that the PSNR value of our method at any sampling rate is at least 1db higher than that of the suboptimal method. In particular, the PSNR value obtained by our method at 5\% sampling rate is 3db higher than that of the suboptimal method, and the numerical results of SSIM, FSIM, and ERGAS are significantly better than other state-of-the-art methods. Finally, the PSNR, SSIM and FSIM values of 32 MSI obtained by different restoration methods are presented in the form of bar charts in Fig.\ref{MSITCQUANTI}. It can be seen that our method performs extremely well for each MSI. But the suboptimal method is not always the NWSTNN method in Table \ref{MSITC1}. For example, for some MSI, the KBRTC method is superior to the NWSTNN method. In summary, it can be concluded that our method is superior to other state-of-the-art methods in MSI experiments.
\subsubsection{MRI completion}
\begin{figure*}[!h] 
	\centering  
	\vspace{0cm} 
	\subfloat[]{
		\begin{minipage}[b]{0.06\linewidth}
			\includegraphics[width=1\linewidth]{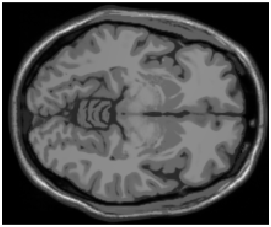}
			\includegraphics[width=1\linewidth]{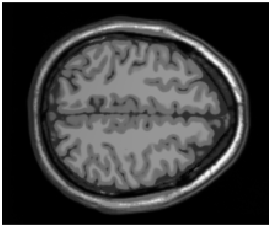}
			\includegraphics[width=1\linewidth]{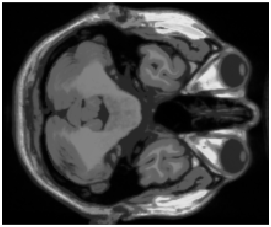}
			\includegraphics[width=1\linewidth]{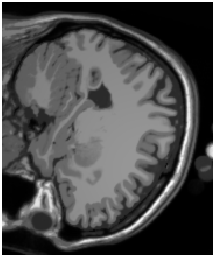}
			\includegraphics[width=1\linewidth]{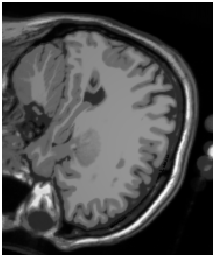}
			\includegraphics[width=1\linewidth]{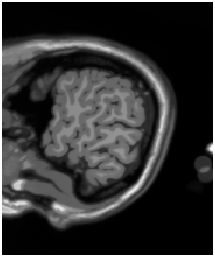}
			\includegraphics[width=1\linewidth]{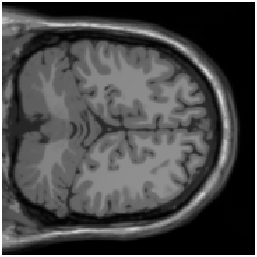}
			\includegraphics[width=1\linewidth]{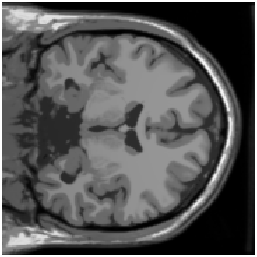}
			\includegraphics[width=1\linewidth]{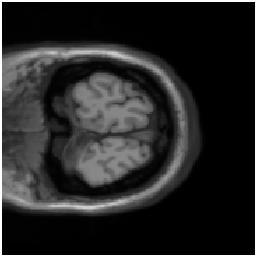}
	\end{minipage}}
	\subfloat[]{
		\begin{minipage}[b]{0.06\linewidth}
			\includegraphics[width=1\linewidth]{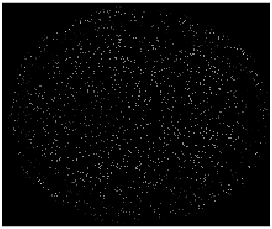}
			\includegraphics[width=1\linewidth]{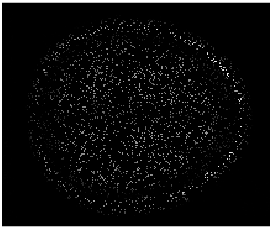}
			\includegraphics[width=1\linewidth]{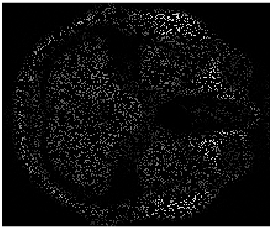}
			\includegraphics[width=1\linewidth]{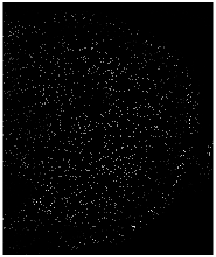}
			\includegraphics[width=1\linewidth]{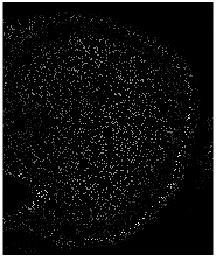}
			\includegraphics[width=1\linewidth]{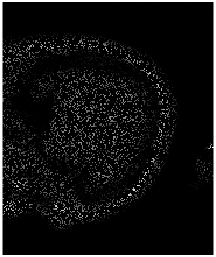}
			\includegraphics[width=1\linewidth]{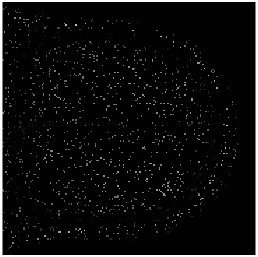}
			\includegraphics[width=1\linewidth]{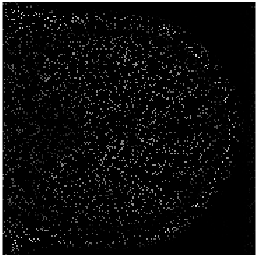}
			\includegraphics[width=1\linewidth]{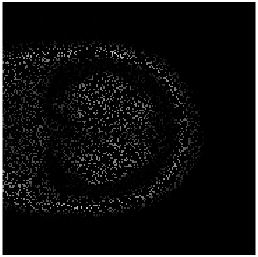}
	\end{minipage}}
	\subfloat[]{
		\begin{minipage}[b]{0.06\linewidth}
			\includegraphics[width=1\linewidth]{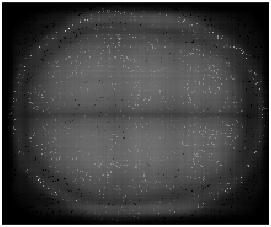}
			\includegraphics[width=1\linewidth]{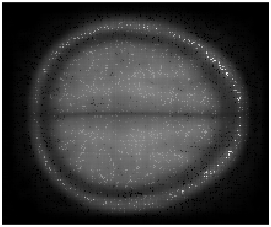}
			\includegraphics[width=1\linewidth]{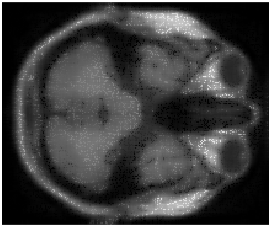}
			\includegraphics[width=1\linewidth]{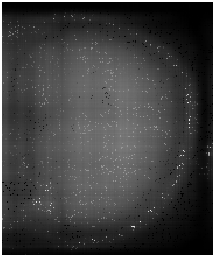}
			\includegraphics[width=1\linewidth]{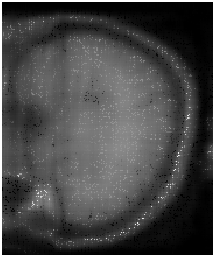}
			\includegraphics[width=1\linewidth]{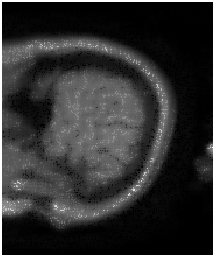}
			\includegraphics[width=1\linewidth]{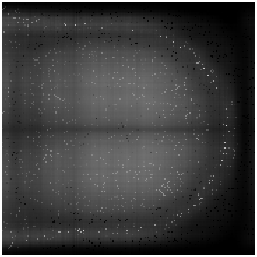}
			\includegraphics[width=1\linewidth]{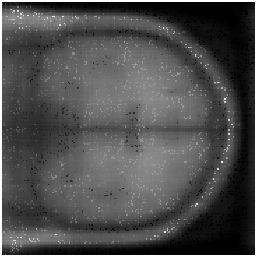}
			\includegraphics[width=1\linewidth]{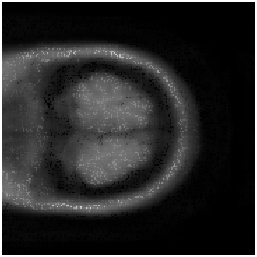}
	\end{minipage}}
	\subfloat[]{
		\begin{minipage}[b]{0.06\linewidth}
			\includegraphics[width=1\linewidth]{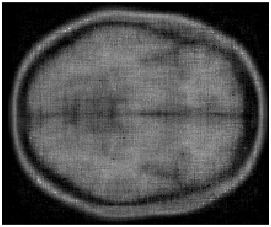}
			\includegraphics[width=1\linewidth]{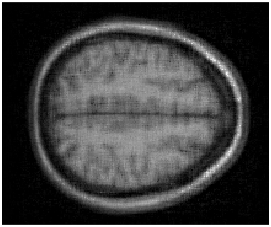}
			\includegraphics[width=1\linewidth]{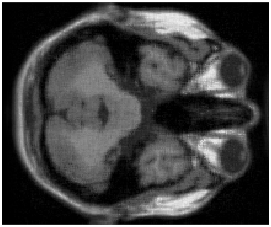}
			\includegraphics[width=1\linewidth]{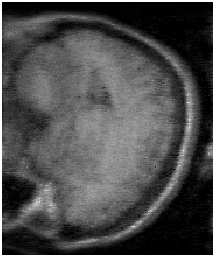}
			\includegraphics[width=1\linewidth]{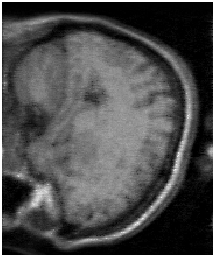}
			\includegraphics[width=1\linewidth]{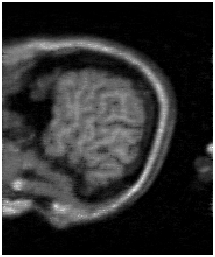}
			\includegraphics[width=1\linewidth]{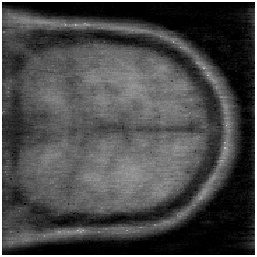}
			\includegraphics[width=1\linewidth]{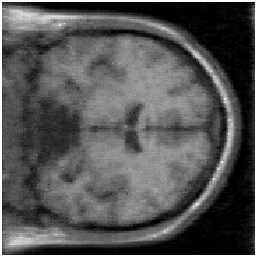}
			\includegraphics[width=1\linewidth]{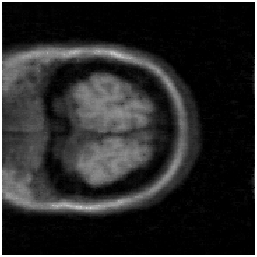}
	\end{minipage}}
	\subfloat[]{
		\begin{minipage}[b]{0.06\linewidth}
			\includegraphics[width=1\linewidth]{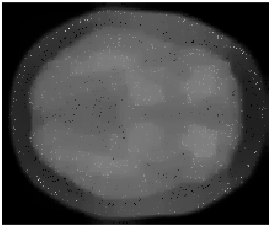}
			\includegraphics[width=1\linewidth]{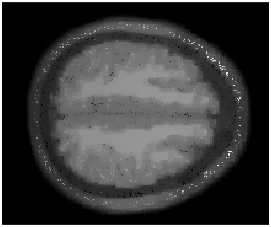}
			\includegraphics[width=1\linewidth]{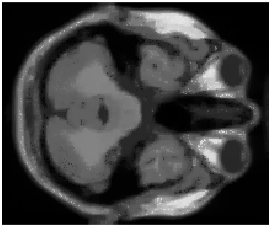}
			\includegraphics[width=1\linewidth]{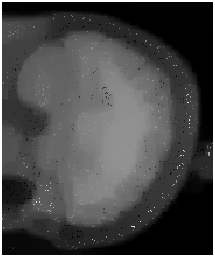}
			\includegraphics[width=1\linewidth]{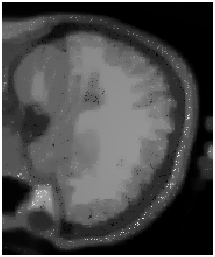}
			\includegraphics[width=1\linewidth]{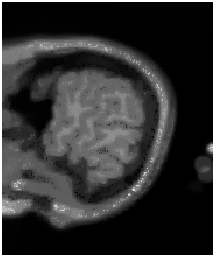}
			\includegraphics[width=1\linewidth]{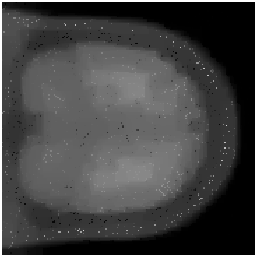}
			\includegraphics[width=1\linewidth]{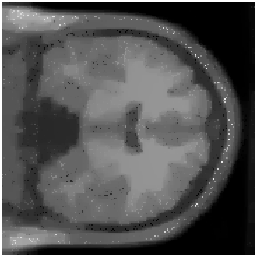}
			\includegraphics[width=1\linewidth]{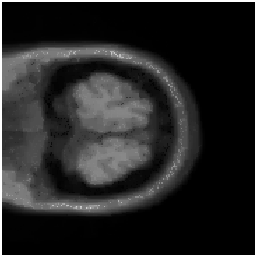}
	\end{minipage}}
	\subfloat[]{
		\begin{minipage}[b]{0.06\linewidth}
			\includegraphics[width=1\linewidth]{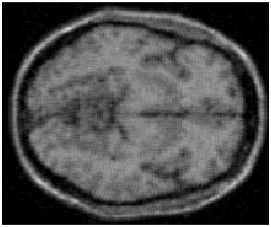}
			\includegraphics[width=1\linewidth]{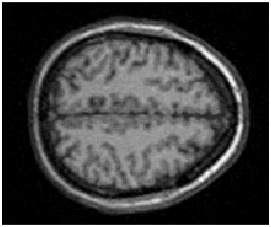}
			\includegraphics[width=1\linewidth]{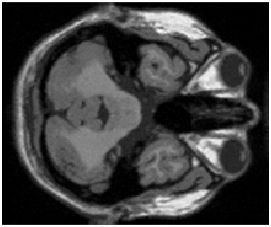}
			\includegraphics[width=1\linewidth]{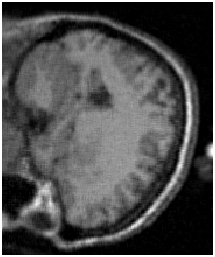}
			\includegraphics[width=1\linewidth]{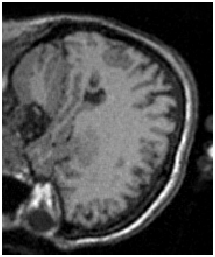}
			\includegraphics[width=1\linewidth]{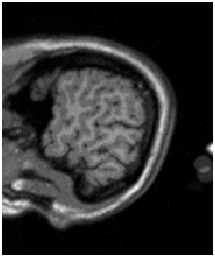}
			\includegraphics[width=1\linewidth]{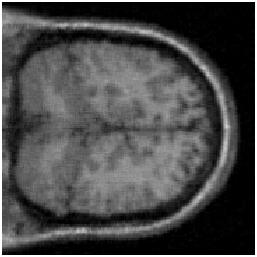}
			\includegraphics[width=1\linewidth]{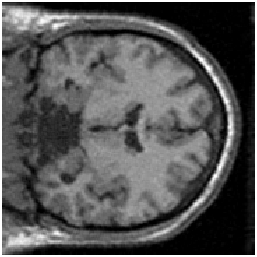}
			\includegraphics[width=1\linewidth]{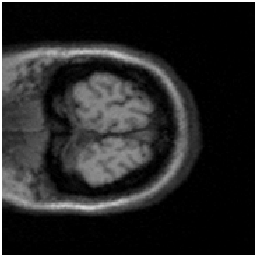}
	\end{minipage}}
	\subfloat[]{
		\begin{minipage}[b]{0.06\linewidth}
			\includegraphics[width=1\linewidth]{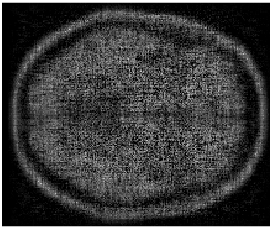}
			\includegraphics[width=1\linewidth]{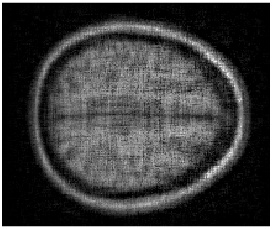}
			\includegraphics[width=1\linewidth]{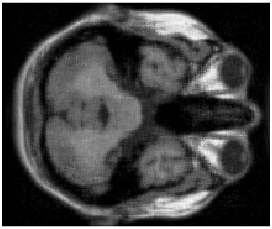}
			\includegraphics[width=1\linewidth]{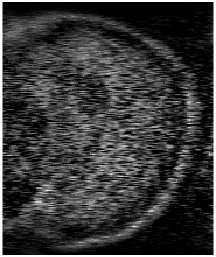}
			\includegraphics[width=1\linewidth]{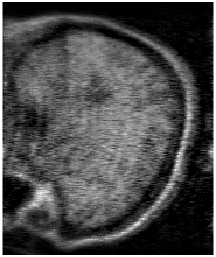}
			\includegraphics[width=1\linewidth]{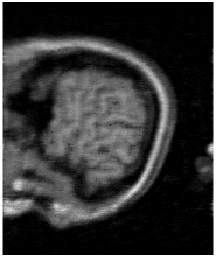}
			\includegraphics[width=1\linewidth]{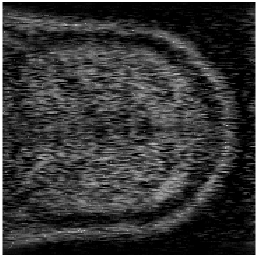}
			\includegraphics[width=1\linewidth]{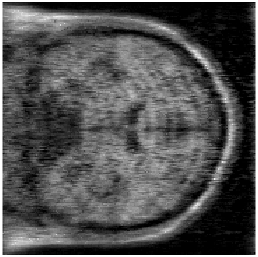}
			\includegraphics[width=1\linewidth]{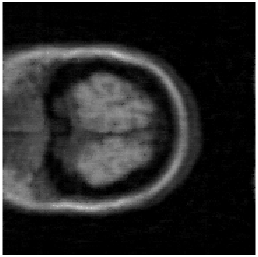}
	\end{minipage}}
	\subfloat[]{
		\begin{minipage}[b]{0.06\linewidth}
			\includegraphics[width=1\linewidth]{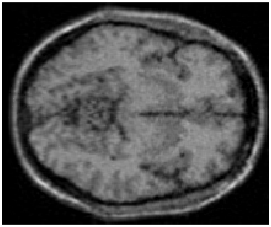}
			\includegraphics[width=1\linewidth]{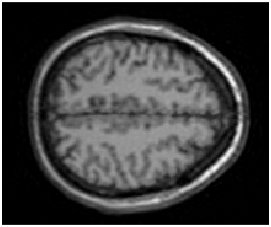}
			\includegraphics[width=1\linewidth]{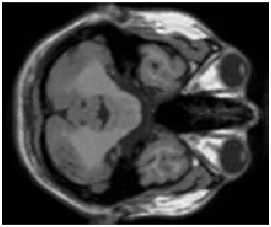}
			\includegraphics[width=1\linewidth]{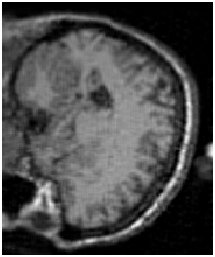}
			\includegraphics[width=1\linewidth]{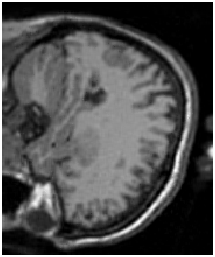}
			\includegraphics[width=1\linewidth]{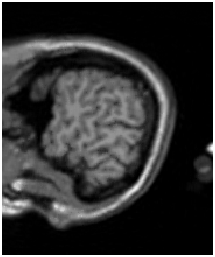}
			\includegraphics[width=1\linewidth]{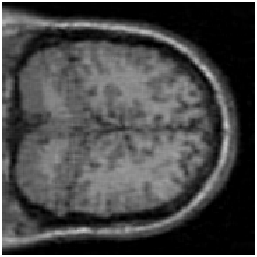}
			\includegraphics[width=1\linewidth]{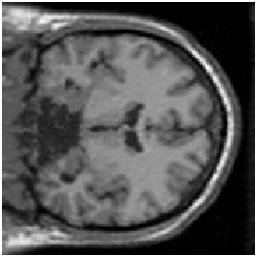}
			\includegraphics[width=1\linewidth]{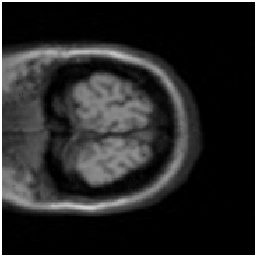}
	\end{minipage}}
	\subfloat[]{
		\begin{minipage}[b]{0.06\linewidth}
			\includegraphics[width=1\linewidth]{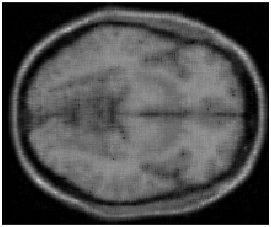}
			\includegraphics[width=1\linewidth]{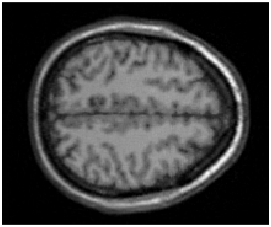}
			\includegraphics[width=1\linewidth]{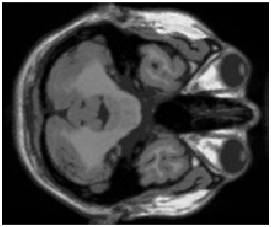}
			\includegraphics[width=1\linewidth]{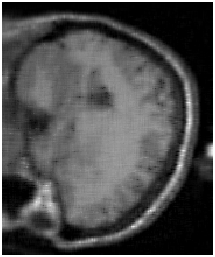}
			\includegraphics[width=1\linewidth]{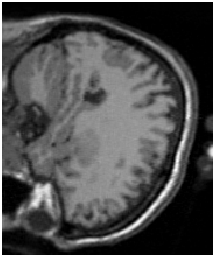}
			\includegraphics[width=1\linewidth]{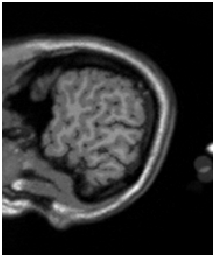}
			\includegraphics[width=1\linewidth]{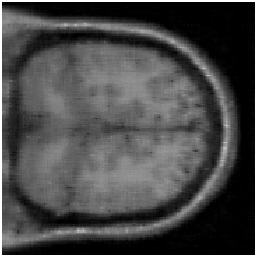}
			\includegraphics[width=1\linewidth]{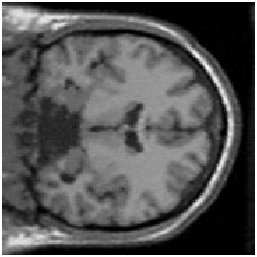}
			\includegraphics[width=1\linewidth]{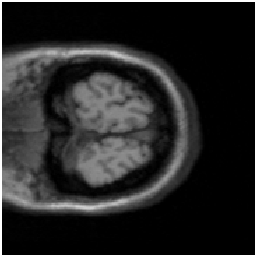}
	\end{minipage}}
	\subfloat[]{
		\begin{minipage}[b]{0.06\linewidth}
			\includegraphics[width=1\linewidth]{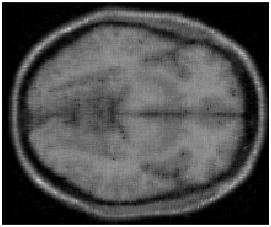}
			\includegraphics[width=1\linewidth]{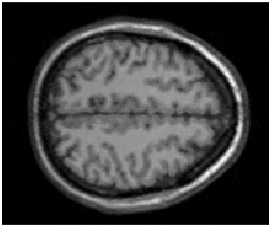}
			\includegraphics[width=1\linewidth]{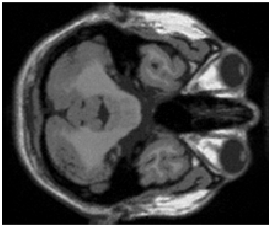}
			\includegraphics[width=1\linewidth]{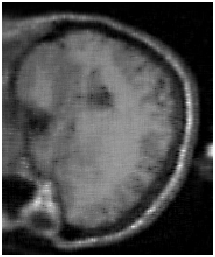}
			\includegraphics[width=1\linewidth]{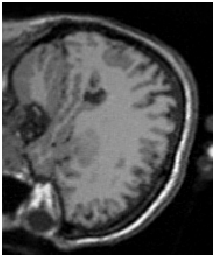}
			\includegraphics[width=1\linewidth]{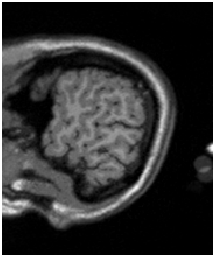}
			\includegraphics[width=1\linewidth]{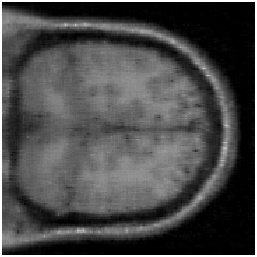}
			\includegraphics[width=1\linewidth]{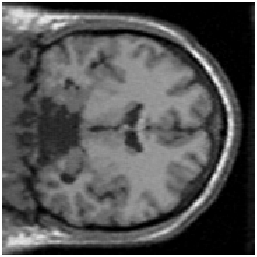}
			\includegraphics[width=1\linewidth]{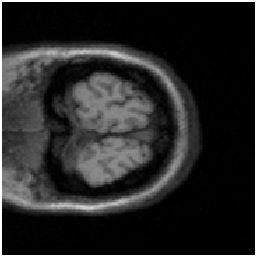}
	\end{minipage}}
	\subfloat[]{
		\begin{minipage}[b]{0.06\linewidth}
			\includegraphics[width=1\linewidth]{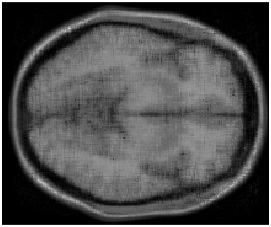}
			\includegraphics[width=1\linewidth]{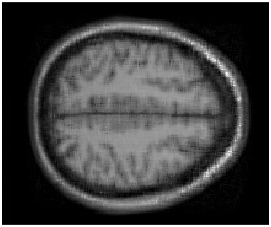}
			\includegraphics[width=1\linewidth]{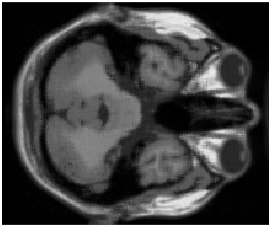}
			\includegraphics[width=1\linewidth]{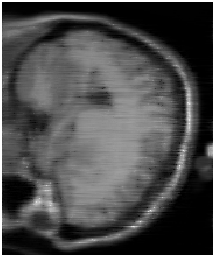}
			\includegraphics[width=1\linewidth]{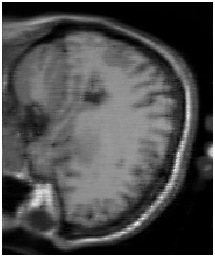}
			\includegraphics[width=1\linewidth]{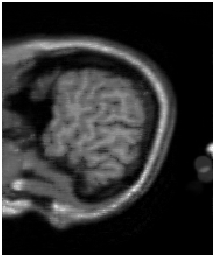}
			\includegraphics[width=1\linewidth]{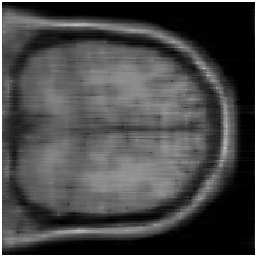}
			\includegraphics[width=1\linewidth]{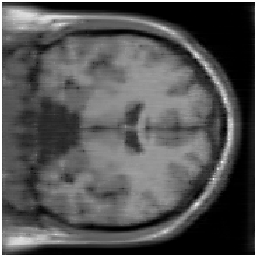}
			\includegraphics[width=1\linewidth]{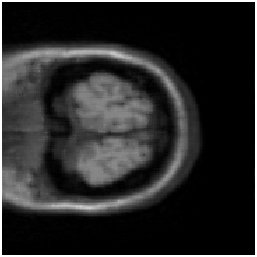}
	\end{minipage}}
	\subfloat[]{
		\begin{minipage}[b]{0.06\linewidth}
			\includegraphics[width=1\linewidth]{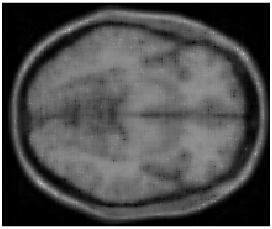}
			\includegraphics[width=1\linewidth]{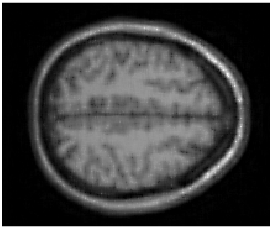}
			\includegraphics[width=1\linewidth]{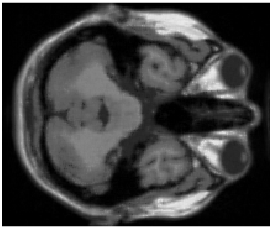}
			\includegraphics[width=1\linewidth]{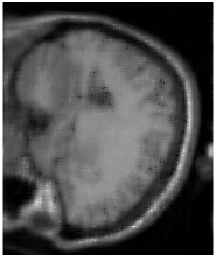}
			\includegraphics[width=1\linewidth]{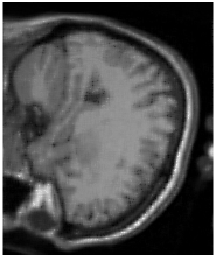}
			\includegraphics[width=1\linewidth]{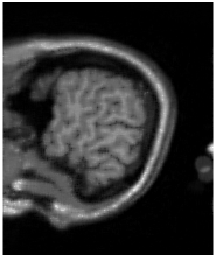}
			\includegraphics[width=1\linewidth]{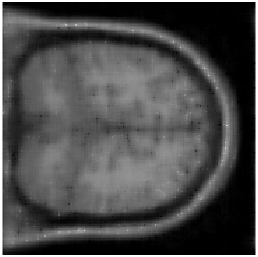}
			\includegraphics[width=1\linewidth]{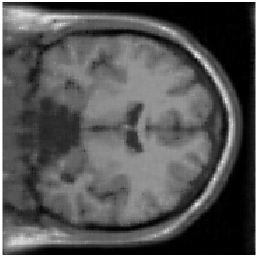}
			\includegraphics[width=1\linewidth]{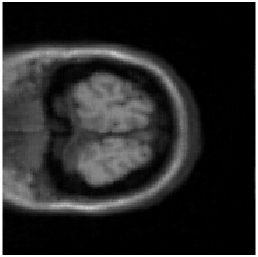}
	\end{minipage}}
	\subfloat[]{
		\begin{minipage}[b]{0.06\linewidth}
			\includegraphics[width=1\linewidth]{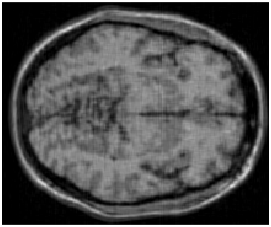}
			\includegraphics[width=1\linewidth]{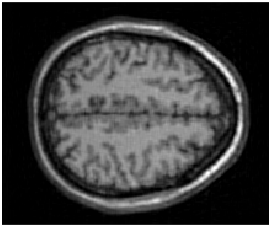}
			\includegraphics[width=1\linewidth]{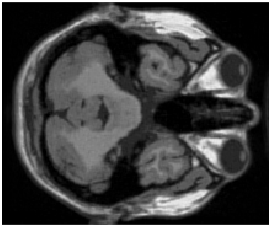}
			\includegraphics[width=1\linewidth]{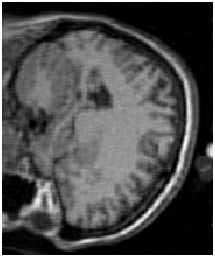}
			\includegraphics[width=1\linewidth]{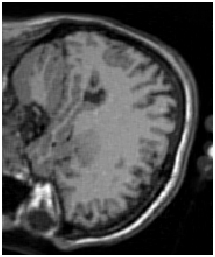}
			\includegraphics[width=1\linewidth]{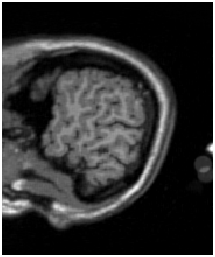}
			\includegraphics[width=1\linewidth]{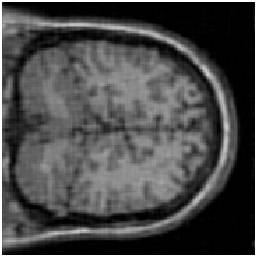}
			\includegraphics[width=1\linewidth]{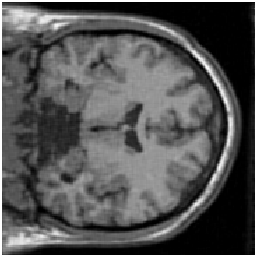}
			\includegraphics[width=1\linewidth]{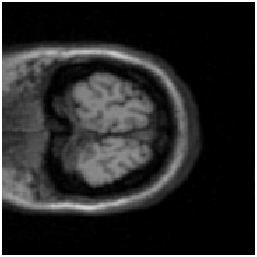}
	\end{minipage}}
	\subfloat[]{
		\begin{minipage}[b]{0.06\linewidth}
			\includegraphics[width=1\linewidth]{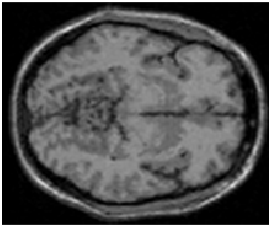}
			\includegraphics[width=1\linewidth]{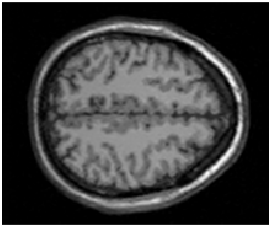}
			\includegraphics[width=1\linewidth]{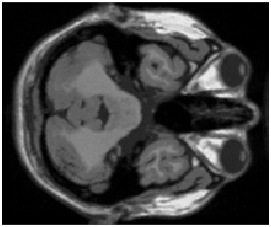}
			\includegraphics[width=1\linewidth]{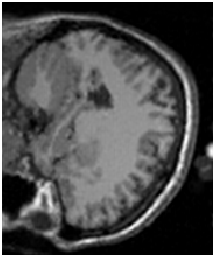}
			\includegraphics[width=1\linewidth]{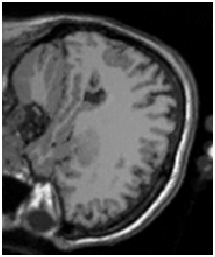}
			\includegraphics[width=1\linewidth]{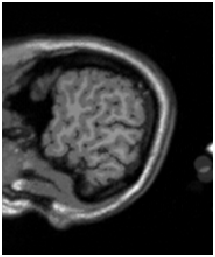}
			\includegraphics[width=1\linewidth]{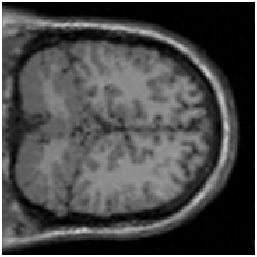}
			\includegraphics[width=1\linewidth]{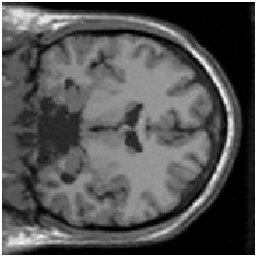}
			\includegraphics[width=1\linewidth]{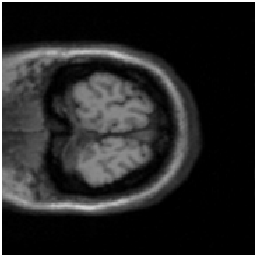}
	\end{minipage}}
	\caption{(a) Original image. (b) Obeserved image. (c) HaLRTC. (d) TNN. (e) LRTCTV-I. (f) McpTC. (g) PSTNN. (h) SMFLRTC. (i) KBRTC. (j) ESPTC. (k) FTNN. (l) WSTNN. (m) NWSTNN. (n) FFMTC. Comparison of FFMTC and the state-of-the-art methods on the MRI of the Frontal slices (top 3 row), Horizontal slices (middle 3 row) and Lateral slices (bottom 3 row). Each type of slice: the first row is the 65th slice with a sampling rate of 5\%, the second row is the 120th slice with a sampling rate of 10\%, and the third row is the 33th slice with a sampling rate of 20\%.}
	\label{MRITC}
\end{figure*}

We test the performance of the proposed method and the comparison method on MRI\footnote{http://brainweb.bic.mni.mcgill.ca/brainweb/selection\_normal.html} data with the size of $181\times217\times181$. Firstly, the visual results of MRI slices in different directions are shown in Fig.\ref{MRITC}, in which slice results of the same rank are used for the same sampling rate. Through observation, we see that the results restored by the suboptimal method, i.e., the NWSTNN method, still have certain ambiguity, while the results obtained by our method do not have this defect, and the most outstanding is that the results in all directions are the best. Therefore, we conclude that our method shows excellent comprehensiveness in the measure of tensor sparse features. Then, we list in Table \ref{MRITC1} the average quantitative results of frontal slices recovered by all methods of MRI at different sampling rates. Obviously, the PSNR value of our method is at least 0.5dB higher than that of the suboptimal method, and the numerical results of SSIM, FSIM and ERGAS are obviously better than that of the suboptimal method. Therefore, the proposed method also has high efficiency in MRI data recovery.
\begin{table*}[]
	\caption{The PSNR, SSIM, FSIM and ERGAS values output by observed and the twelve utilized LRTC methods for MRI.}
	\resizebox{\textwidth}{!}{
		\begin{tabular}{cccccccccccccc}
			\hline
			SR        & \multicolumn{4}{c}{5\%}                                                                                  & \multicolumn{4}{c}{10\%}                                                                                & \multicolumn{4}{c}{20\%}                                                                                & Time(s)  \\
			Method    & PSNR                     & SSIM                    & FSIM                    & ERGAS                     & PSNR                     & SSIM                    & FSIM                    & ERGAS                    & PSNR                     & SSIM                    & FSIM                    & ERGAS                    &          \\ \hline
			Observed  & 11.399                   & 0.310                   & 0.530                   & 1021.127                  & 11.635                   & 0.323                   & 0.565                   & 993.791                  & 12.145                   & 0.350                   & 0.612                   & 937.049                  & 0.000    \\
			HaLRTC    & 17.291                   & 0.298                   & 0.638                   & 537.831                   & 20.131                   & 0.439                   & 0.726                   & 389.565                  & 24.426                   & 0.659                   & 0.829                   & 235.793                  & 27.316   \\
			TNN       & 22.725                   & 0.471                   & 0.743                   & 302.480                   & 26.079                   & 0.642                   & 0.812                   & 205.024                  & 29.973                   & 0.799                   & 0.882                   & 130.965                  & 97.054   \\
			LRTCTV-I & 19.377                   & 0.598                   & 0.702                   & 432.845                   & 22.895                   & 0.750                   & 0.806                   & 292.166                  & 28.184                   & 0.890                   & 0.908                   & 155.939                  & 671.241  \\
			McpTC     & 27.939                   & 0.749                   & 0.843                   & 153.404                   & 31.459                   & 0.845                   & 0.888                   & 102.679                  & 35.569                   & 0.937                   & 0.941                   & 63.986                   & 622.650  \\
			PSTNN     & 16.196                   & 0.197                   & 0.589                   & 606.927                   & 22.447                   & 0.438                   & 0.723                   & 307.792                  & 29.577                   & 0.767                   & 0.870                   & 137.637                  & 113.446  \\
			SMFLRTC   & 28.527                   & 0.792                   & 0.859                   & 146.250                   & 32.187                   & 0.895                   & 0.911                   & 95.272                   & 35.119                   & 0.944                   & 0.943                   & 66.885                   & 1996.388 \\
			KBRTC     & 26.126                   & 0.728                   & 0.831                   & 192.742                   & 32.696                   & 0.912                   & 0.924                   & 89.493                   & 37.089                   & 0.967                   & 0.966                   & 53.662                   & 335.954  \\
			ESPTC     & 25.534                   & 0.695                   & 0.820                   & 208.066                   & 32.535                   & 0.904                   & 0.918                   & 91.721                   & 35.969                   & 0.949                   & 0.949                   & 61.482                   & 546.520  \\
			FTNN      & 24.884                   & 0.693                   & 0.836                   & 231.625                   & 28.348                   & 0.826                   & 0.895                   & 151.686                  & 32.738                   & 0.923                   & 0.945                   & 89.804                   & 1280.618 \\
			WSTNN     & 25.557                   & 0.709                   & 0.825                   & 211.023                   & 29.100                   & 0.837                   & 0.888                   & 138.317                  & 33.472                   & 0.928                   & 0.940                   & 83.124                   & 277.745  \\
			NWSTNN    & 30.221                   & 0.826                   & 0.883                   & 120.052                   & 33.294                   & 0.902                   & 0.924                   & 83.646                   & 36.849                   & 0.950                   & 0.956                   & 55.015                   & 458.180  \\
			FFMTC     & \textbf{30.893} & \textbf{0.880} & \textbf{0.902} & \textbf{109.815} & \textbf{34.160} & \textbf{0.934} & \textbf{0.937} & \textbf{75.110} & \textbf{37.891} & \textbf{0.969} & \textbf{0.967} & \textbf{48.770} & \textbf{434.761}  \\ \hline
	\end{tabular}}\label{MRITC1}
\end{table*}

\subsubsection{CV completion}
\begin{figure*}[!h] 
	\centering  
	\vspace{0cm} 
	\subfloat[]{
		\begin{minipage}[b]{0.06\linewidth}
			\includegraphics[width=1\linewidth]{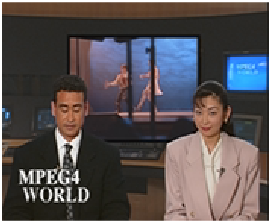}
			\includegraphics[width=1\linewidth]{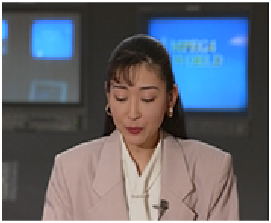}
			\includegraphics[width=1\linewidth]{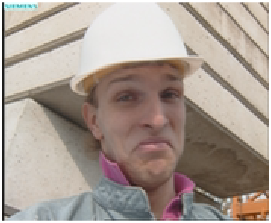}
			\includegraphics[width=1\linewidth]{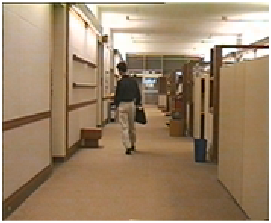}
			\includegraphics[width=1\linewidth]{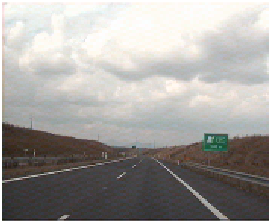}
			\includegraphics[width=1\linewidth]{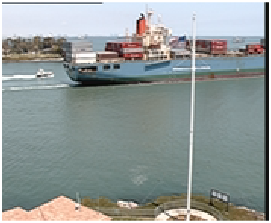}
			\includegraphics[width=1\linewidth]{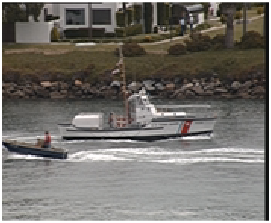}
	\end{minipage}}
	\subfloat[]{
		\begin{minipage}[b]{0.06\linewidth}
			\includegraphics[width=1\linewidth]{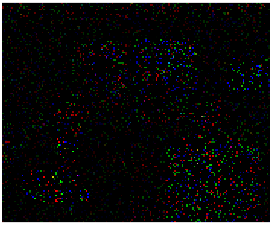}
			\includegraphics[width=1\linewidth]{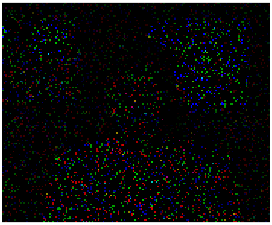}
			\includegraphics[width=1\linewidth]{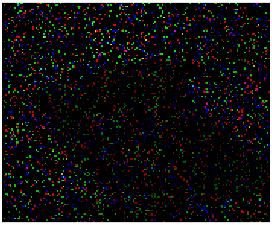}
			\includegraphics[width=1\linewidth]{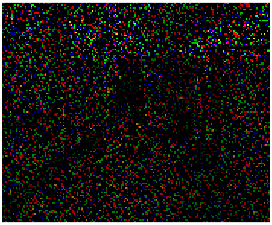}
			\includegraphics[width=1\linewidth]{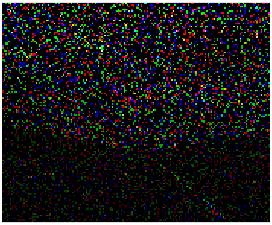}
			\includegraphics[width=1\linewidth]{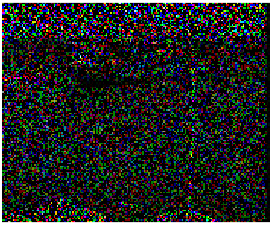}
			\includegraphics[width=1\linewidth]{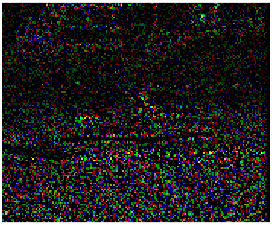}
	\end{minipage}}
	\subfloat[]{
		\begin{minipage}[b]{0.06\linewidth}
			\includegraphics[width=1\linewidth]{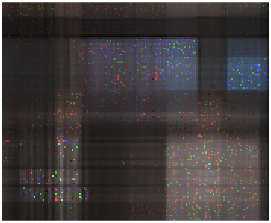}
			\includegraphics[width=1\linewidth]{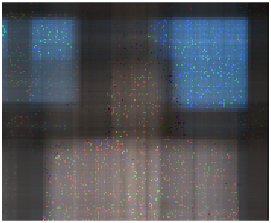}
			\includegraphics[width=1\linewidth]{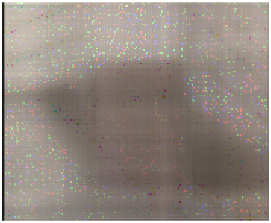}
			\includegraphics[width=1\linewidth]{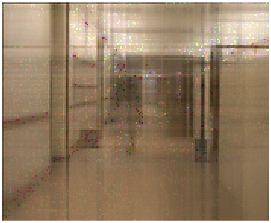}
			\includegraphics[width=1\linewidth]{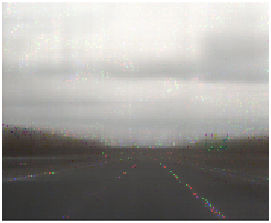}
			\includegraphics[width=1\linewidth]{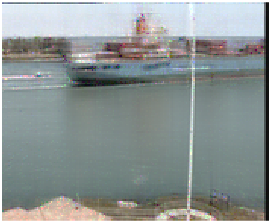}
			\includegraphics[width=1\linewidth]{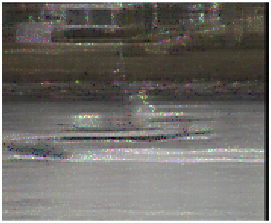}
	\end{minipage}}
	\subfloat[]{
		\begin{minipage}[b]{0.06\linewidth}
			\includegraphics[width=1\linewidth]{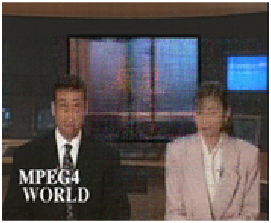}
			\includegraphics[width=1\linewidth]{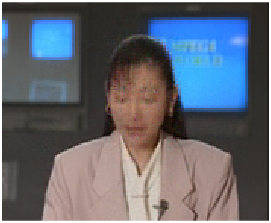}
			\includegraphics[width=1\linewidth]{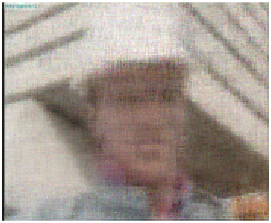}
			\includegraphics[width=1\linewidth]{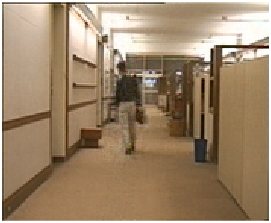}
			\includegraphics[width=1\linewidth]{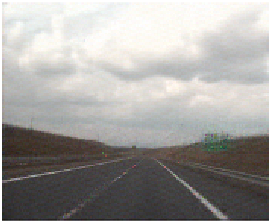}
			\includegraphics[width=1\linewidth]{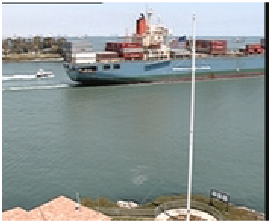}
			\includegraphics[width=1\linewidth]{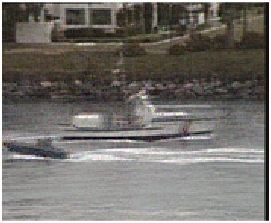}
	\end{minipage}}
	\subfloat[]{
		\begin{minipage}[b]{0.06\linewidth}
			\includegraphics[width=1\linewidth]{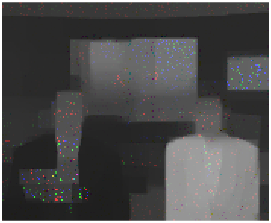}
			\includegraphics[width=1\linewidth]{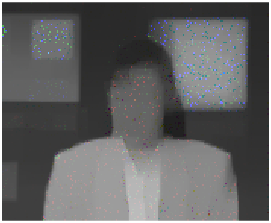}
			\includegraphics[width=1\linewidth]{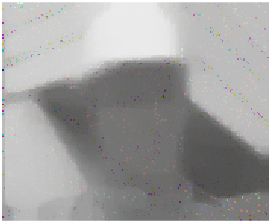}
			\includegraphics[width=1\linewidth]{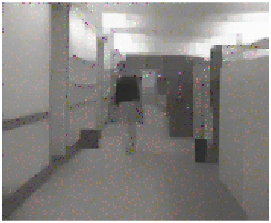}
			\includegraphics[width=1\linewidth]{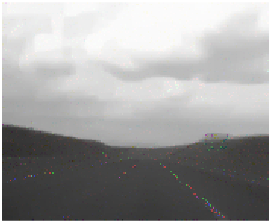}
			\includegraphics[width=1\linewidth]{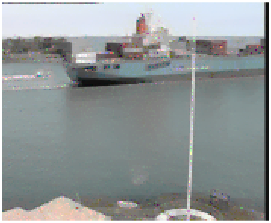}
			\includegraphics[width=1\linewidth]{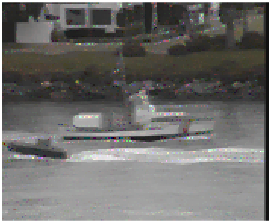}
	\end{minipage}}
	\subfloat[]{
		\begin{minipage}[b]{0.06\linewidth}
			\includegraphics[width=1\linewidth]{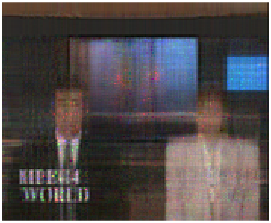}
			\includegraphics[width=1\linewidth]{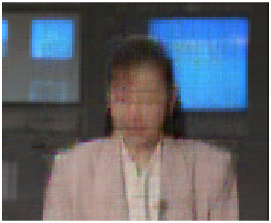}
			\includegraphics[width=1\linewidth]{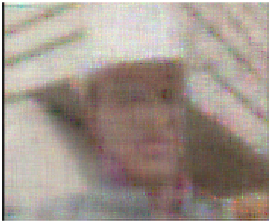}
			\includegraphics[width=1\linewidth]{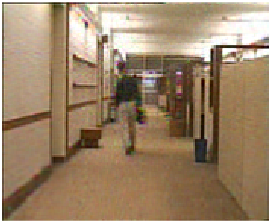}
			\includegraphics[width=1\linewidth]{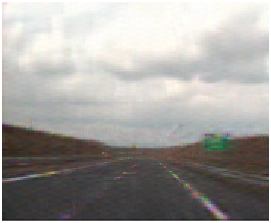}
			\includegraphics[width=1\linewidth]{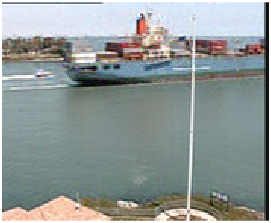}
			\includegraphics[width=1\linewidth]{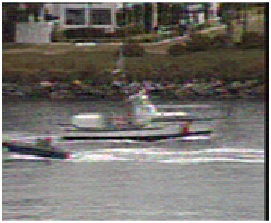}
	\end{minipage}}
	\subfloat[]{
		\begin{minipage}[b]{0.06\linewidth}
			\includegraphics[width=1\linewidth]{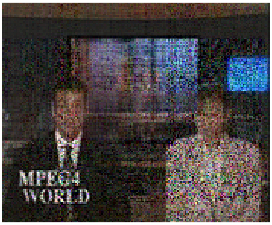}
			\includegraphics[width=1\linewidth]{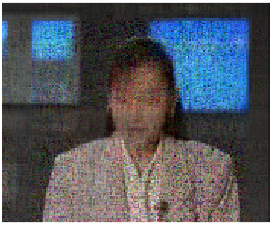}
			\includegraphics[width=1\linewidth]{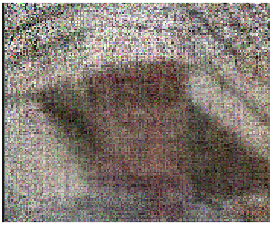}
			\includegraphics[width=1\linewidth]{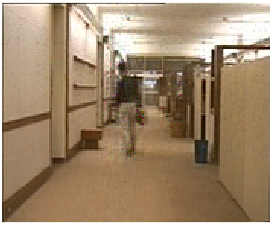}
			\includegraphics[width=1\linewidth]{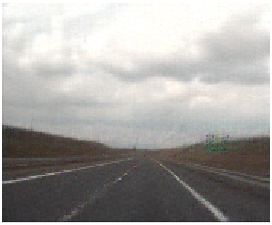}
			\includegraphics[width=1\linewidth]{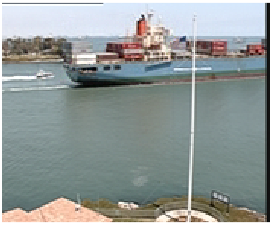}
			\includegraphics[width=1\linewidth]{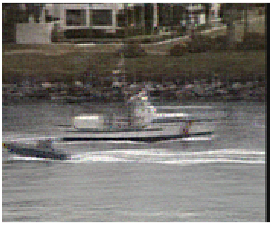}
	\end{minipage}}
	\subfloat[]{
		\begin{minipage}[b]{0.06\linewidth}
			\includegraphics[width=1\linewidth]{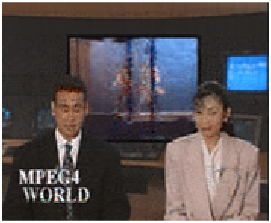}
			\includegraphics[width=1\linewidth]{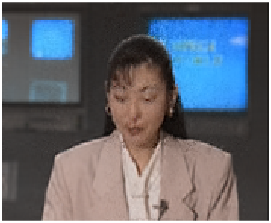}
			\includegraphics[width=1\linewidth]{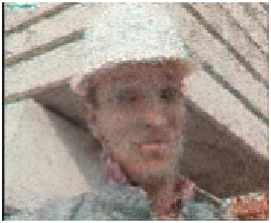}
			\includegraphics[width=1\linewidth]{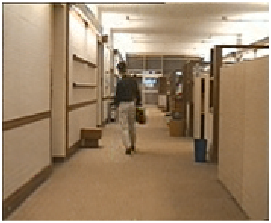}
			\includegraphics[width=1\linewidth]{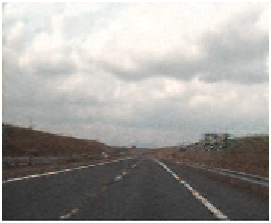}
			\includegraphics[width=1\linewidth]{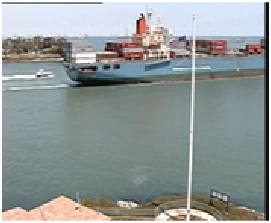}
			\includegraphics[width=1\linewidth]{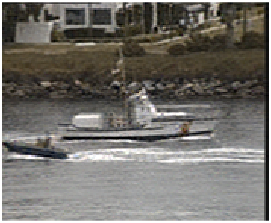}
	\end{minipage}}
	\subfloat[]{
		\begin{minipage}[b]{0.06\linewidth}
			\includegraphics[width=1\linewidth]{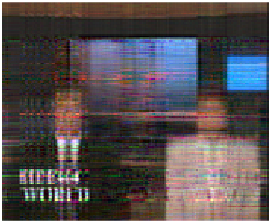}
			\includegraphics[width=1\linewidth]{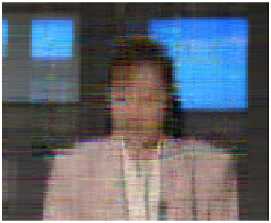}
			\includegraphics[width=1\linewidth]{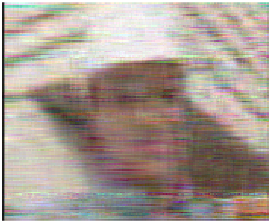}
			\includegraphics[width=1\linewidth]{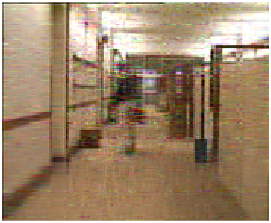}
			\includegraphics[width=1\linewidth]{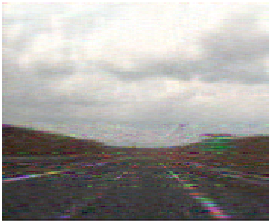}
			\includegraphics[width=1\linewidth]{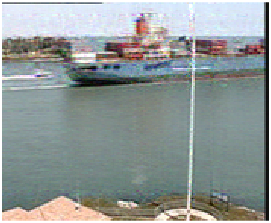}
			\includegraphics[width=1\linewidth]{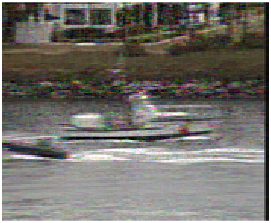}
	\end{minipage}}
	\subfloat[]{
		\begin{minipage}[b]{0.06\linewidth}
			\includegraphics[width=1\linewidth]{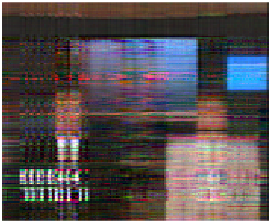}
			\includegraphics[width=1\linewidth]{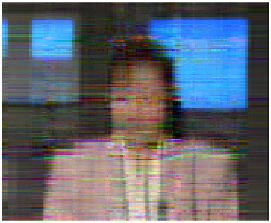}
			\includegraphics[width=1\linewidth]{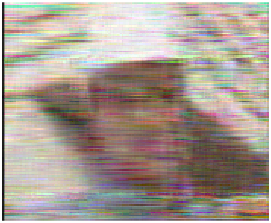}
			\includegraphics[width=1\linewidth]{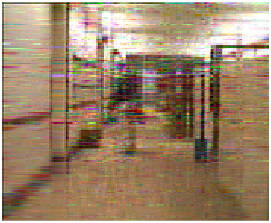}
			\includegraphics[width=1\linewidth]{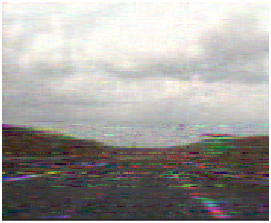}
			\includegraphics[width=1\linewidth]{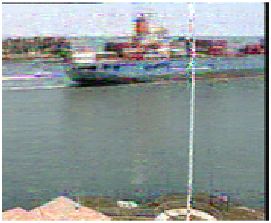}
			\includegraphics[width=1\linewidth]{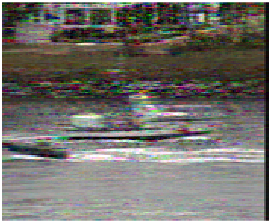}
	\end{minipage}}
	\subfloat[]{
		\begin{minipage}[b]{0.06\linewidth}
			\includegraphics[width=1\linewidth]{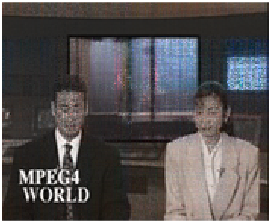}
			\includegraphics[width=1\linewidth]{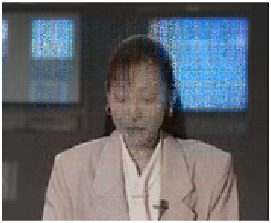}
			\includegraphics[width=1\linewidth]{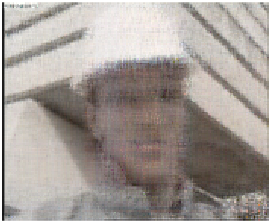}
			\includegraphics[width=1\linewidth]{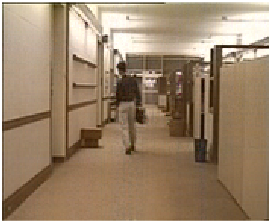}
			\includegraphics[width=1\linewidth]{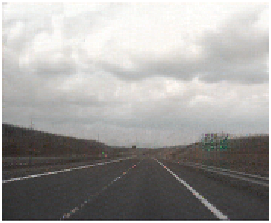}
			\includegraphics[width=1\linewidth]{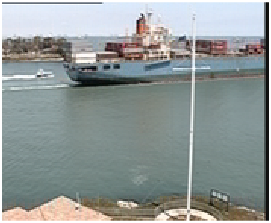}
			\includegraphics[width=1\linewidth]{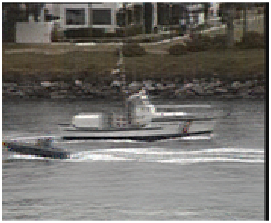}
	\end{minipage}}
	\subfloat[]{
		\begin{minipage}[b]{0.06\linewidth}
			\includegraphics[width=1\linewidth]{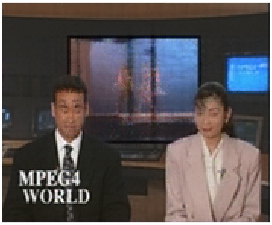}
			\includegraphics[width=1\linewidth]{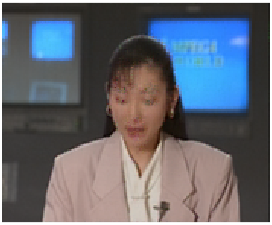}
			\includegraphics[width=1\linewidth]{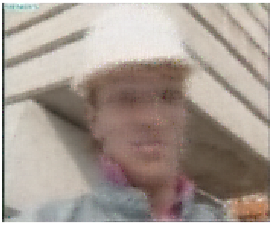}
			\includegraphics[width=1\linewidth]{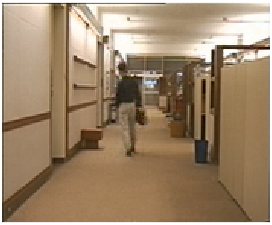}
			\includegraphics[width=1\linewidth]{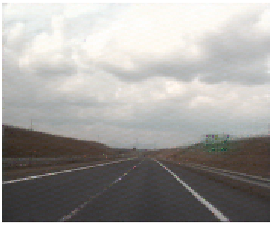}
			\includegraphics[width=1\linewidth]{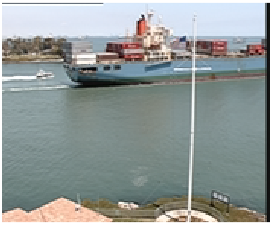}
			\includegraphics[width=1\linewidth]{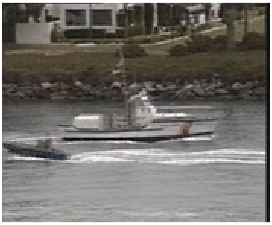}
	\end{minipage}}
	\subfloat[]{
		\begin{minipage}[b]{0.06\linewidth}
			\includegraphics[width=1\linewidth]{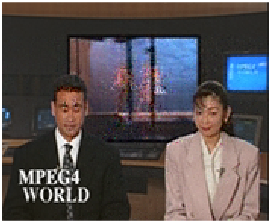}
			\includegraphics[width=1\linewidth]{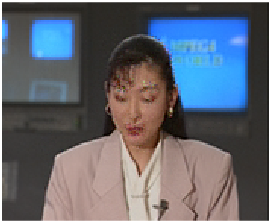}
			\includegraphics[width=1\linewidth]{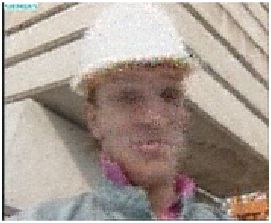}
			\includegraphics[width=1\linewidth]{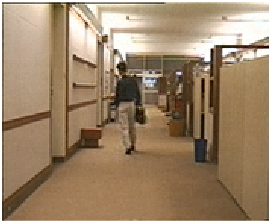}
			\includegraphics[width=1\linewidth]{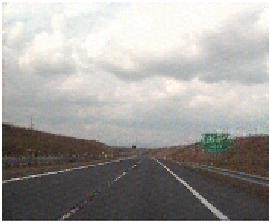}
			\includegraphics[width=1\linewidth]{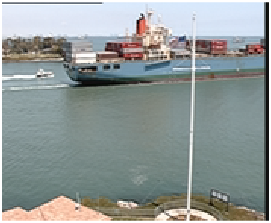}
			\includegraphics[width=1\linewidth]{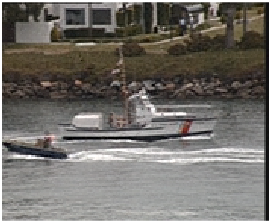}
	\end{minipage}}
	\subfloat[]{
		\begin{minipage}[b]{0.06\linewidth}
			\includegraphics[width=1\linewidth]{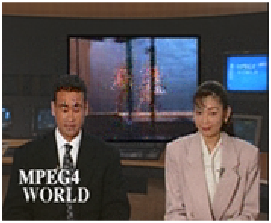}
			\includegraphics[width=1\linewidth]{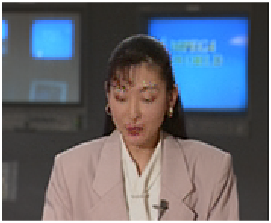}
			\includegraphics[width=1\linewidth]{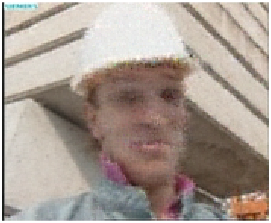}
			\includegraphics[width=1\linewidth]{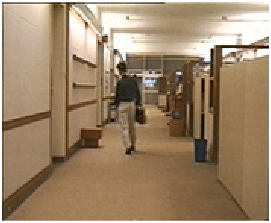}
			\includegraphics[width=1\linewidth]{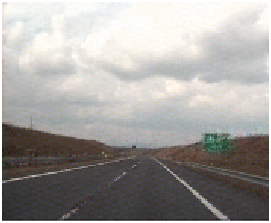}
			\includegraphics[width=1\linewidth]{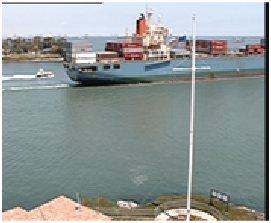}
			\includegraphics[width=1\linewidth]{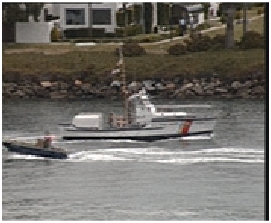}
	\end{minipage}}
	\caption{(a) Original image. (b) Obeserved image. (c) HaLRTC. (d) TNN. (e) LRTCTV-I. (f) McpTC. (g) PSTNN. (h) SMFLRTC. (i) KBRTC. (j) ESPTC. (k) FTNN. (l) WSTNN. (m) NWSTNN. (n) FFMTC.  SR: top 3 rows is 5\%, middle 2 rows is 10\% and last 2 rows is 20\%. The rows of CVs are in order: the 7th frame of news, the 14th frame of akiyo, the 21st frame of foreman, the 28th frame of hall, the 35th frame of highway, the 42nd frame of container, and the 49th frame of coastguard.}
	\label{CVTC}
\end{figure*}

\begin{figure*}[!h] 
	\centering  
	\vspace{0cm} 
	\subfloat[SR:5\%]{
		\begin{minipage}[b]{0.3\linewidth}
			\includegraphics[width=1\linewidth]{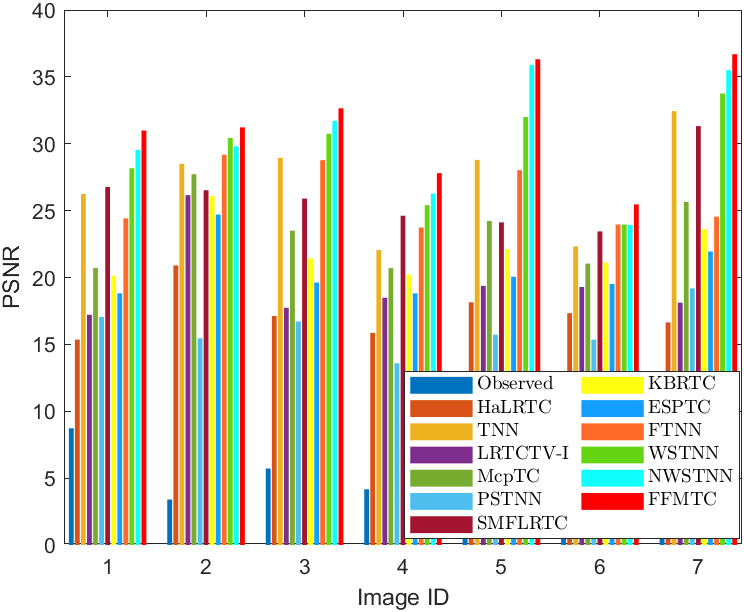}
			\includegraphics[width=1\linewidth]{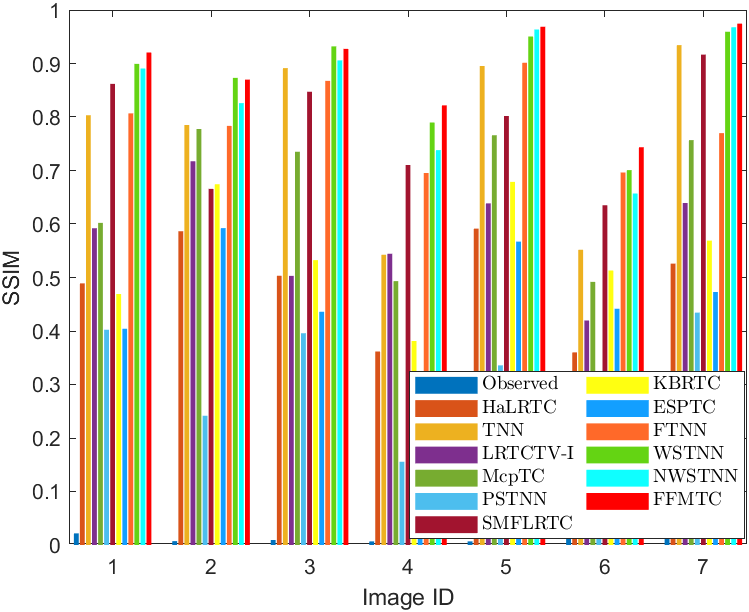}
			\includegraphics[width=1\linewidth]{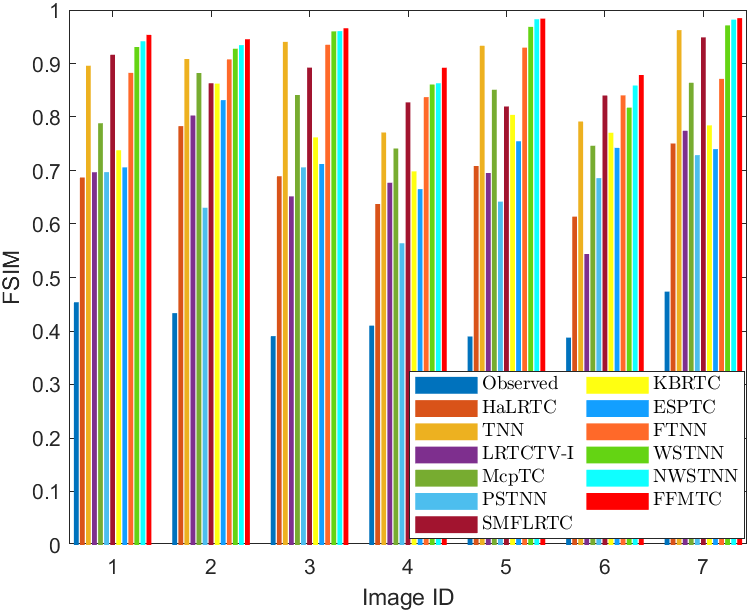}
	\end{minipage}}
	\subfloat[SR:10\%]{
		\begin{minipage}[b]{0.3\linewidth}
			\includegraphics[width=1\linewidth]{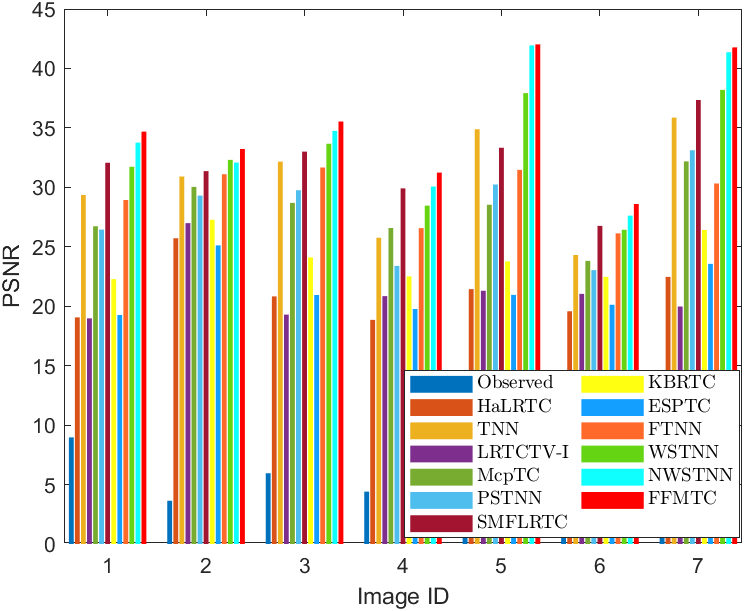}
			\includegraphics[width=1\linewidth]{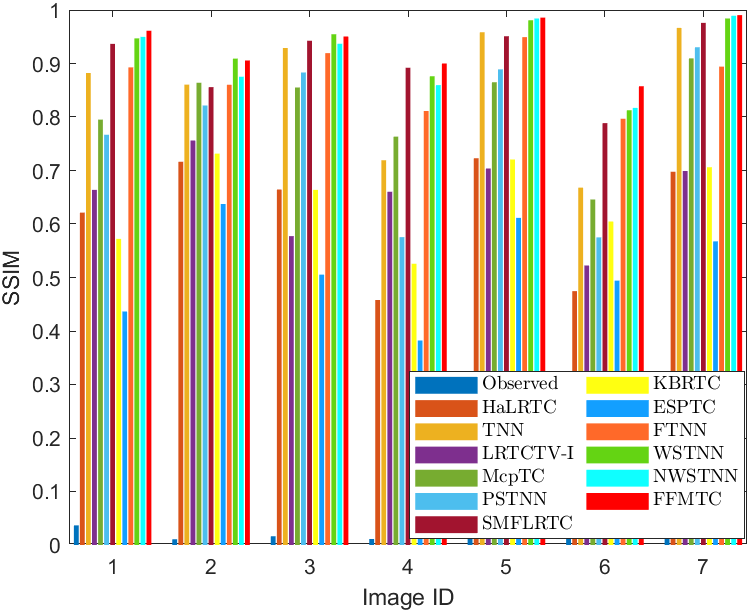}
			\includegraphics[width=1\linewidth]{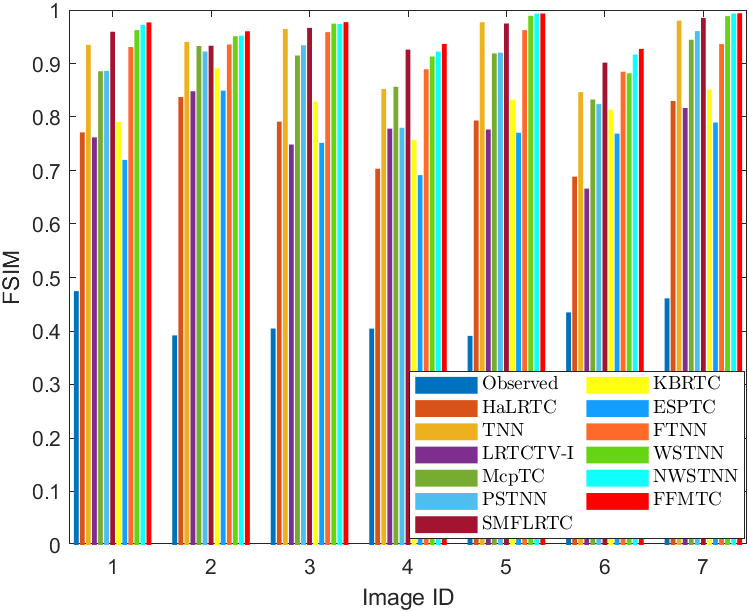}
	\end{minipage}}
	\subfloat[SR:20\%]{
		\begin{minipage}[b]{0.3\linewidth}
			\includegraphics[width=1\linewidth]{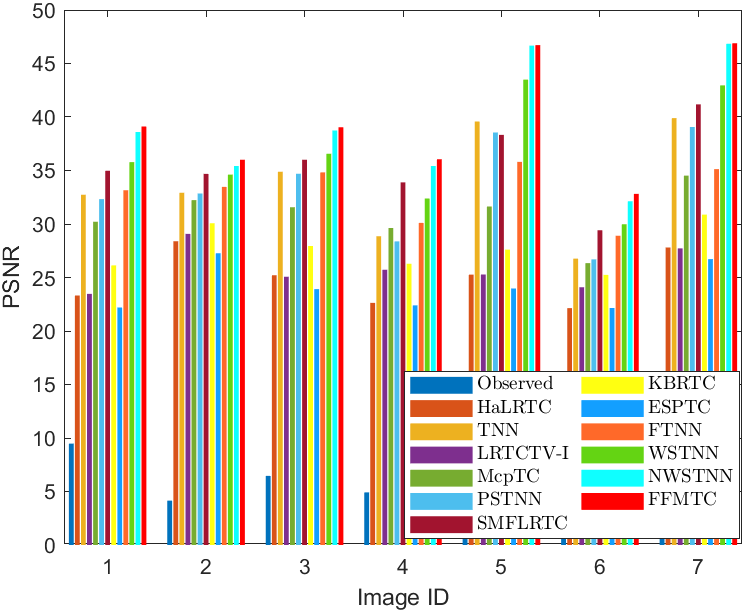}
			\includegraphics[width=1\linewidth]{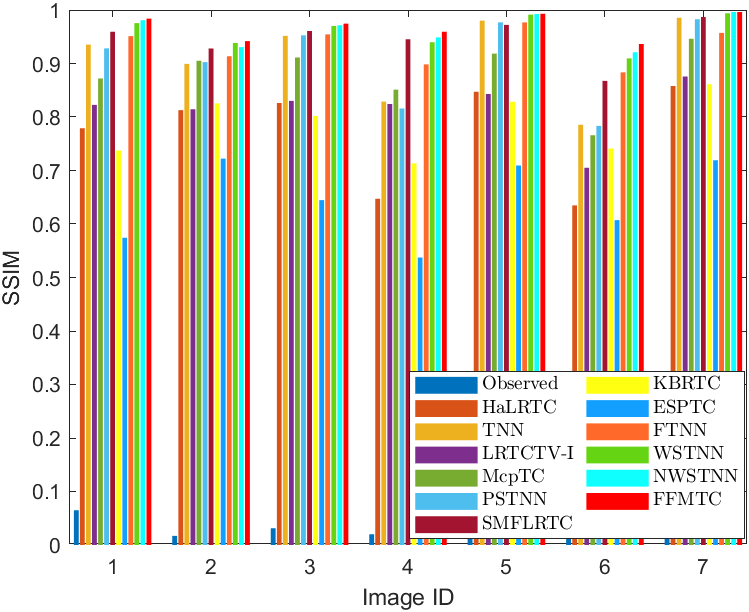}
			\includegraphics[width=1\linewidth]{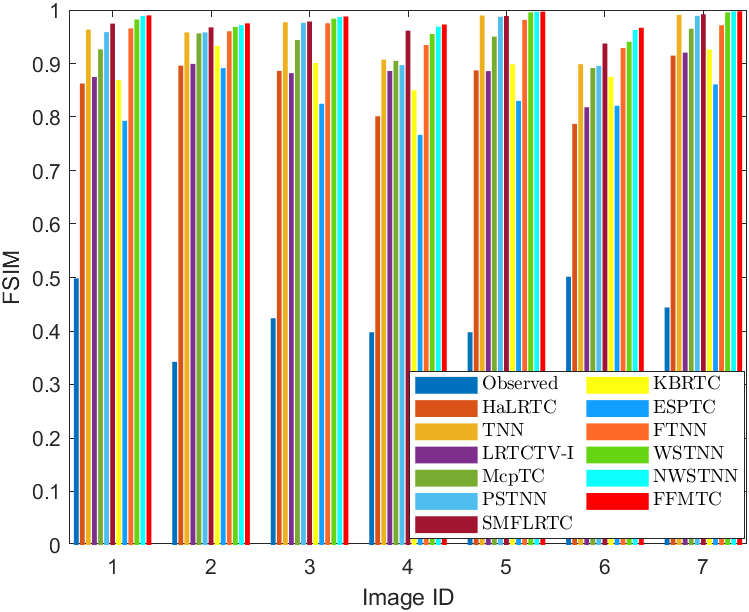}
	\end{minipage}}
	\caption{Comparison of the PSNR values (top), SSIM values (middle) and FSIM values (bottom) obtained by FFMTC and the state-of-the-art methods on 32 MSIs. SR: (a)5\%, (b)10\%, (c)20\%.}
	\label{CVTCQUANTI}
\end{figure*}

We test seven CVs\footnote{http://trace.eas.asu.edu/yuv/}(respectively named news, akiyo, foreman, hall, highway, container, coastguard) of size $144 \times 176 \times 3 \times 50$. Firstly, we report the visual results of 7 CVs in our experiment in Fig.\ref{CVTC}, in which the frame number and sampling rate corresponding to each CV are described in the annotation. Obviously, we can see from the figure that our results are optimal. Furthermore, the average quantitative results of 7 CVs are listed in Table \ref{CVTC1}. By comparison, it is found that the value of PSNR of our method is at least 0.4 dB higher than that of the suboptimal method, and the PSNR result of our method is 1.2 dB higher than that of the suboptimal method at 5\% sampling rate. Finally, the quantitative results of each CV are displayed in a bar chart in Fig.\ref{CVTCQUANTI}. We clearly see that our method still maintains great advantages for CV experiments.
\begin{table*}[]
	\caption{The average PSNR, SSIM, FSIM and ERGAS values for 7 CVs tested by observed and the twelve utilized LRTC methods.}
	\resizebox{\textwidth}{!}{
		\begin{tabular}{cccccccccccccc}
			\hline
			SR        & \multicolumn{4}{c}{5\%}                                             & \multicolumn{4}{c}{10\%}                                            & \multicolumn{4}{c}{20\%}                                            & Time(s) \\ 
			Method    & PSNR            & SSIM           & FSIM           & ERGAS           & PSNR            & SSIM           & FSIM           & ERGAS           & PSNR            & SSIM           & FSIM           & ERGAS           &         \\ \hline
			Observed  & 5.793           & 0.011          & 0.420          & 1194.943        & 6.028           & 0.019          & 0.423          & 1163.006        & 6.538           & 0.034          & 0.429          & 1096.668        & 0.000   \\
			HaLRTC    & 17.339          & 0.488          & 0.696          & 329.274         & 21.134          & 0.622          & 0.774          & 214.716         & 24.964          & 0.772          & 0.862          & 137.902         & 15.661  \\
			TNN       & 27.043          & 0.772          & 0.886          & 113.373         & 30.464          & 0.855          & 0.928          & 79.447          & 33.661          & 0.909          & 0.955          & 56.828          & 43.497  \\
			LRTCTV-I & 19.482          & 0.579          & 0.692          & 273.371         & 21.205          & 0.655          & 0.771          & 228.372         & 25.776          & 0.816          & 0.881          & 127.382         & 291.480 \\
			McpTC     & 23.363          & 0.660          & 0.816          & 168.038         & 28.081          & 0.814          & 0.898          & 97.166          & 30.878          & 0.881          & 0.934          & 70.218          & 313.965 \\
			PSTNN     & 16.152          & 0.313          & 0.665          & 364.861         & 27.901          & 0.777          & 0.890          & 102.994         & 33.223          & 0.906          & 0.952          & 59.007          & 44.006  \\
			SMFLRTC   & 26.099          & 0.777          & 0.873          & 119.374         & 31.967          & 0.906          & 0.949          & 62.964          & 35.497          & 0.946          & 0.971          & 42.931          & 856.302 \\
			KBRTC     & 22.103          & 0.545          & 0.774          & 190.439         & 24.117          & 0.646          & 0.824          & 150.466         & 27.734          & 0.787          & 0.893          & 99.394          & 158.768 \\
			ESPTC     & 20.498          & 0.461          & 0.736          & 228.808         & 21.391          & 0.519          & 0.763          & 206.860         & 24.087          & 0.645          & 0.827          & 151.579         & 305.808 \\
			FTNN      & 26.092          & 0.789          & 0.886          & 127.310         & 29.455          & 0.875          & 0.928          & 84.665          & 33.055          & 0.934          & 0.960          & 56.255          & 648.280 \\
			WSTNN     & 29.212          & 0.872          & 0.920          & 88.525          & 32.676          & 0.923          & 0.951          & 61.909          & 36.539          & 0.960          & 0.975          & 40.870          & 200.654 \\
			NWSTNN    & 30.382          & 0.850          & 0.932          & 80.843          & 34.511          & 0.916          & 0.961          & 52.606          & 39.120          & 0.963          & 0.982          & 31.234          & 319.875 \\
			FFMTC     & \textbf{31.592} & \textbf{0.889} & \textbf{0.943} & \textbf{69.471} & \textbf{35.296} & \textbf{0.936} & \textbf{0.967} & \textbf{47.310} & \textbf{39.521} & \textbf{0.969} & \textbf{0.984} & \textbf{29.423} & \textbf{341.327}\\ \hline
	\end{tabular}}\label{CVTC1}
\end{table*}

\subsubsection{HSV completion}
\begin{figure*}[!h] 
	\centering  
	\vspace{0cm} 
	\subfloat[]{
		\begin{minipage}[b]{0.06\linewidth}
			\includegraphics[width=1\linewidth]{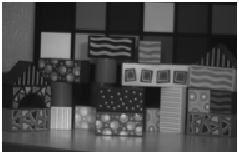}
			\includegraphics[width=1\linewidth]{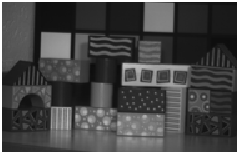}
			\includegraphics[width=1\linewidth]{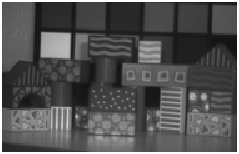}
			\includegraphics[width=1\linewidth]{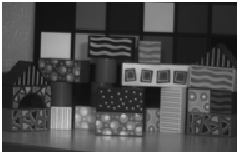}
			\includegraphics[width=1\linewidth]{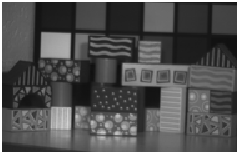}
			\includegraphics[width=1\linewidth]{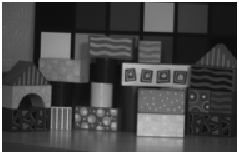}
	\end{minipage}}
	\subfloat[]{
		\begin{minipage}[b]{0.06\linewidth}
			\includegraphics[width=1\linewidth]{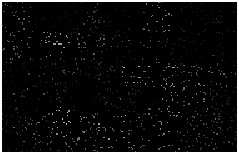}
			\includegraphics[width=1\linewidth]{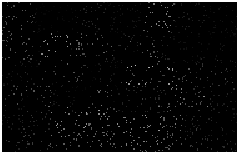}
			\includegraphics[width=1\linewidth]{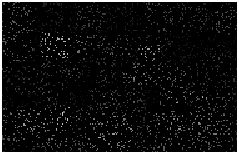}
			\includegraphics[width=1\linewidth]{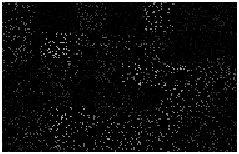}
			\includegraphics[width=1\linewidth]{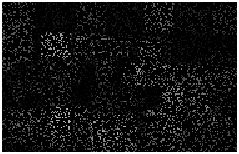}
			\includegraphics[width=1\linewidth]{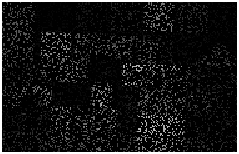}
	\end{minipage}}
	\subfloat[]{
		\begin{minipage}[b]{0.06\linewidth}
			\includegraphics[width=1\linewidth]{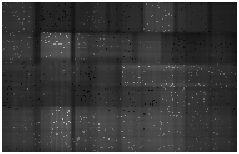}
			\includegraphics[width=1\linewidth]{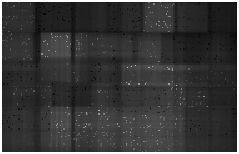}
			\includegraphics[width=1\linewidth]{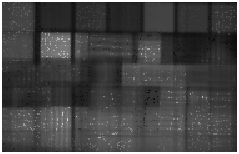}
			\includegraphics[width=1\linewidth]{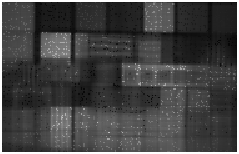}
			\includegraphics[width=1\linewidth]{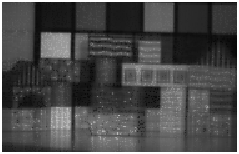}
			\includegraphics[width=1\linewidth]{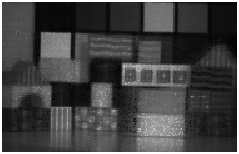}
	\end{minipage}}
	\subfloat[]{
		\begin{minipage}[b]{0.06\linewidth}
			\includegraphics[width=1\linewidth]{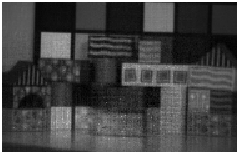}
			\includegraphics[width=1\linewidth]{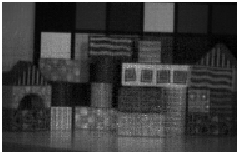}
			\includegraphics[width=1\linewidth]{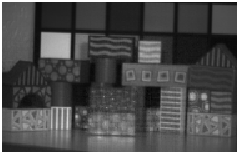}
			\includegraphics[width=1\linewidth]{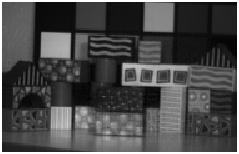}
			\includegraphics[width=1\linewidth]{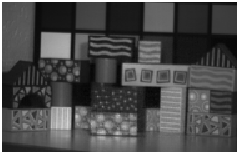}
			\includegraphics[width=1\linewidth]{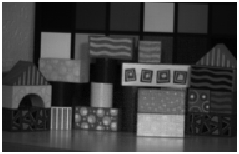}
	\end{minipage}}
	\subfloat[]{
		\begin{minipage}[b]{0.06\linewidth}
			\includegraphics[width=1\linewidth]{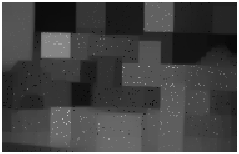}
			\includegraphics[width=1\linewidth]{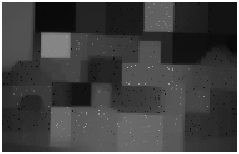}
			\includegraphics[width=1\linewidth]{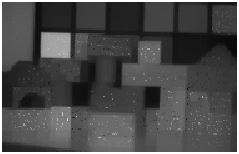}
			\includegraphics[width=1\linewidth]{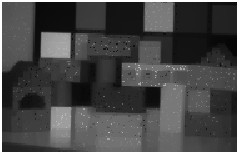}
			\includegraphics[width=1\linewidth]{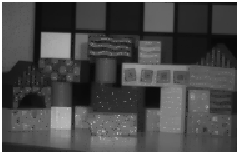}
			\includegraphics[width=1\linewidth]{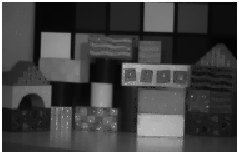}
	\end{minipage}}
	\subfloat[]{
		\begin{minipage}[b]{0.06\linewidth}
			\includegraphics[width=1\linewidth]{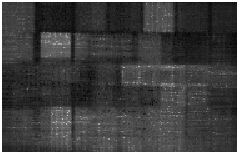}
			\includegraphics[width=1\linewidth]{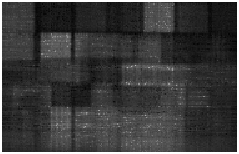}
			\includegraphics[width=1\linewidth]{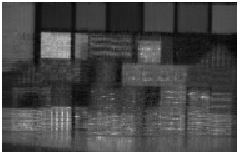}
			\includegraphics[width=1\linewidth]{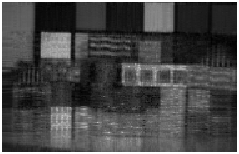}
			\includegraphics[width=1\linewidth]{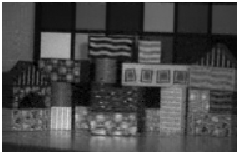}
			\includegraphics[width=1\linewidth]{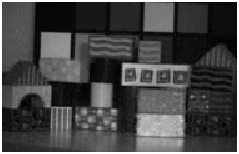}
	\end{minipage}}
	\subfloat[]{
		\begin{minipage}[b]{0.06\linewidth}
			\includegraphics[width=1\linewidth]{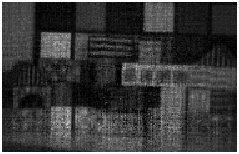}
			\includegraphics[width=1\linewidth]{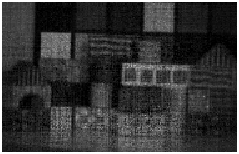}
			\includegraphics[width=1\linewidth]{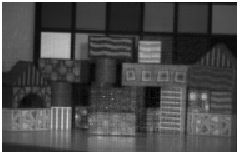}
			\includegraphics[width=1\linewidth]{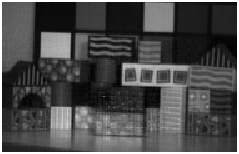}
			\includegraphics[width=1\linewidth]{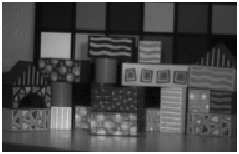}
			\includegraphics[width=1\linewidth]{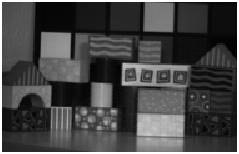}
	\end{minipage}}
	\subfloat[]{
		\begin{minipage}[b]{0.06\linewidth}
			\includegraphics[width=1\linewidth]{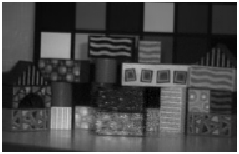}
			\includegraphics[width=1\linewidth]{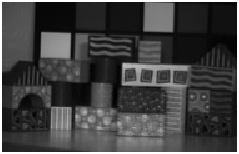}
			\includegraphics[width=1\linewidth]{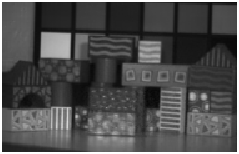}
			\includegraphics[width=1\linewidth]{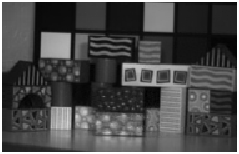}
			\includegraphics[width=1\linewidth]{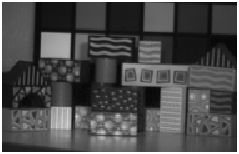}
			\includegraphics[width=1\linewidth]{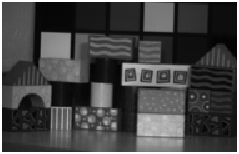}
	\end{minipage}}
	\subfloat[]{
		\begin{minipage}[b]{0.06\linewidth}
			\includegraphics[width=1\linewidth]{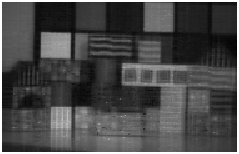}
			\includegraphics[width=1\linewidth]{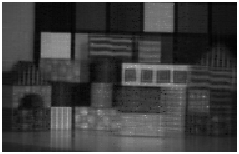}
			\includegraphics[width=1\linewidth]{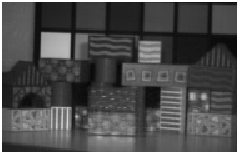}
			\includegraphics[width=1\linewidth]{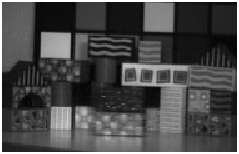}
			\includegraphics[width=1\linewidth]{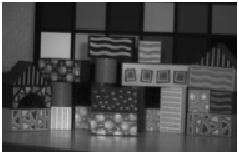}
			\includegraphics[width=1\linewidth]{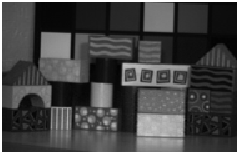}
	\end{minipage}}
	\subfloat[]{
		\begin{minipage}[b]{0.06\linewidth}
			\includegraphics[width=1\linewidth]{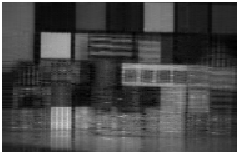}
			\includegraphics[width=1\linewidth]{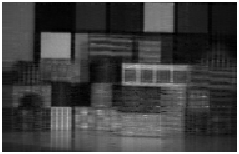}
			\includegraphics[width=1\linewidth]{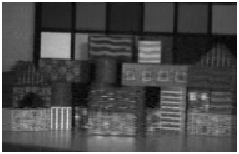}
			\includegraphics[width=1\linewidth]{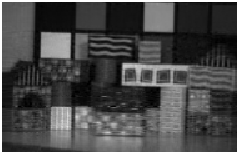}
			\includegraphics[width=1\linewidth]{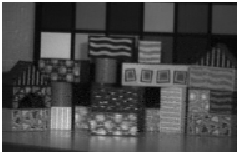}
			\includegraphics[width=1\linewidth]{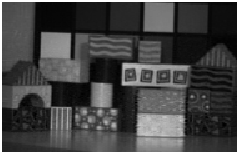}
	\end{minipage}}
	\subfloat[]{
		\begin{minipage}[b]{0.06\linewidth}
			\includegraphics[width=1\linewidth]{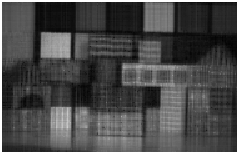}
			\includegraphics[width=1\linewidth]{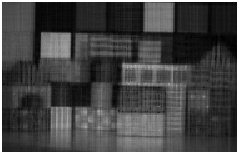}
			\includegraphics[width=1\linewidth]{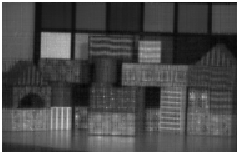}
			\includegraphics[width=1\linewidth]{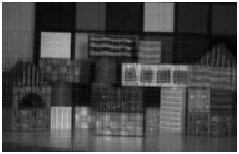}
			\includegraphics[width=1\linewidth]{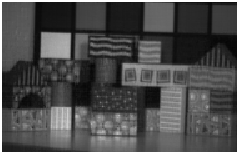}
			\includegraphics[width=1\linewidth]{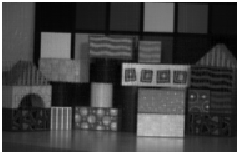}
	\end{minipage}}
	\subfloat[]{
		\begin{minipage}[b]{0.06\linewidth}
			\includegraphics[width=1\linewidth]{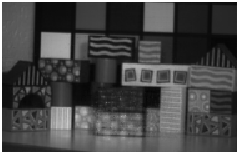}
			\includegraphics[width=1\linewidth]{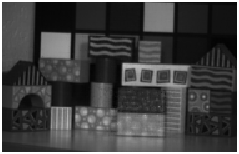}
			\includegraphics[width=1\linewidth]{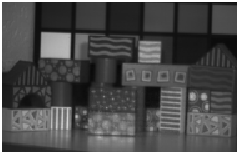}
			\includegraphics[width=1\linewidth]{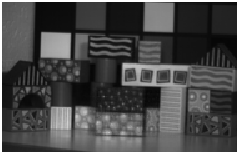}
			\includegraphics[width=1\linewidth]{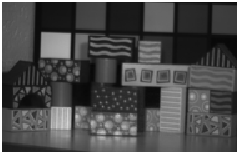}
			\includegraphics[width=1\linewidth]{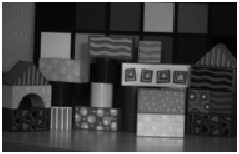}
	\end{minipage}}
	\subfloat[]{
		\begin{minipage}[b]{0.06\linewidth}
			\includegraphics[width=1\linewidth]{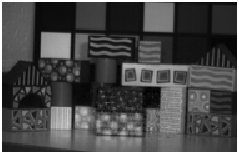}
			\includegraphics[width=1\linewidth]{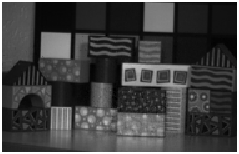}
			\includegraphics[width=1\linewidth]{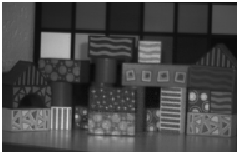}
			\includegraphics[width=1\linewidth]{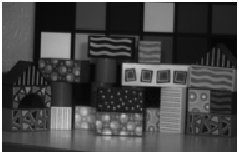}
			\includegraphics[width=1\linewidth]{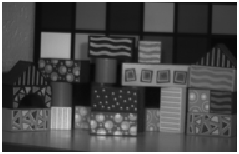}
			\includegraphics[width=1\linewidth]{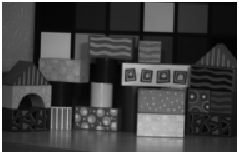}
	\end{minipage}}
	\subfloat[]{
		\begin{minipage}[b]{0.06\linewidth}
			\includegraphics[width=1\linewidth]{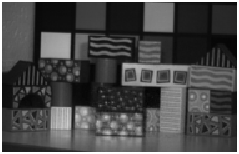}
			\includegraphics[width=1\linewidth]{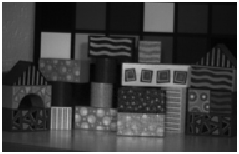}
			\includegraphics[width=1\linewidth]{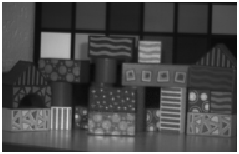}
			\includegraphics[width=1\linewidth]{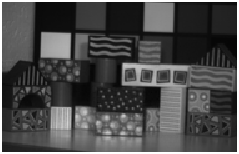}
			\includegraphics[width=1\linewidth]{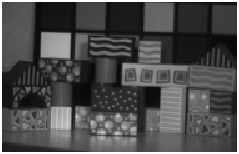}
			\includegraphics[width=1\linewidth]{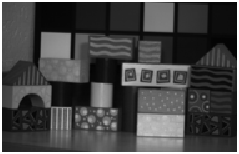}
	\end{minipage}}
	\caption{(a) Original image. (b) Obeserved image. (c) HaLRTC. (d) TNN. (e) LRTCTV-I. (f) McpTC. (g) PSTNN. (h) SMFLRTC. (i) KBRTC. (j) ESPTC. (k) FTNN. (l) WSTNN. (m) NWSTNN. (n) FFMTC. SR: top 2 rows is 5\%, middle 2 rows is 10\% and last 2 rows is 20\%. The number of frame and band corresponding to each row are as follows: 7 band and 25 frame, 11 band and 16 frame, 3 band and 28 frame, 8 band and 25 frame, 5 band and 27 frame, 14 band and 8 frame.}
	\label{HSVTC}
\end{figure*}

We test an HSV\footnote{http://openremotesensing.net/knowledgebase/hyperspectral-video/.} of size $120 \times 188 \times 33 \times 31$. Specifically, this HSV has 31 frames, and each frame has 33 bands of wavelengths of from 400 nm to 720 nm with a 10 nm step \cite{106582012673}. In Fig.\ref{HSVTC}, we show the results of different frame numbers and different bands with three sampling rates. Experimental results show that our method is significantly better than other methods in restoring texture information. Table \ref{HSVTC1} lists the values of the PSNR, SSIM, FSIM and ERGAS of the tested HSV recovered by different LRTC methods. The experimental results exhibit that the proposed method has reached the highest value in all evaluation indicators. No matter how the SR is set, compared with the sub-optimal method, the method achieves a gain of about 0.7dB in the PSNR.
\begin{table*}[]
	\caption{The PSNR, SSIM, FSIM and ERGAS values output by observed and the twelve utilized LRTC methods for HSV.}
	\resizebox{\textwidth}{!}{
		\begin{tabular}{cccccccccccccc}
			\hline
			SR        & \multicolumn{4}{c}{5\%}                                                                                 & \multicolumn{4}{c}{10\%}                                                                                & \multicolumn{4}{c}{20\%}                                                                               & Time(s)           \\
			Method    & PSNR                     & SSIM                    & FSIM                    & ERGAS                    & PSNR                     & SSIM                    & FSIM                    & ERGAS                    & PSNR                     & SSIM                    & FSIM                    & ERGAS                   &                   \\ \hline
			Observed  & 10.761                   & 0.029                   & 0.501                   & 924.271                  & 10.995                   & 0.047                   & 0.513                   & 899.681                  & 11.507                   & 0.083                   & 0.529                   & 848.164                 & 0.000             \\
			HaLRTC    & 18.213                   & 0.479                   & 0.710                   & 409.377                  & 21.565                   & 0.629                   & 0.803                   & 280.265                  & 28.091                   & 0.845                   & 0.913                   & 134.492                 & 119.097           \\
			TNN       & 26.632                   & 0.821                   & 0.909                   & 154.287                  & 36.716                   & 0.957                   & 0.976                   & 49.763                   & 41.185                   & 0.981                   & 0.989                   & 29.945                  & 284.852           \\
			LRTCTV-I & 21.521                   & 0.624                   & 0.775                   & 283.354                  & 25.288                   & 0.759                   & 0.861                   & 193.983                  & 30.189                   & 0.877                   & 0.933                   & 119.254                 & 2368.555          \\
			McpTC     & 20.178                   & 0.524                   & 0.759                   & 329.922                  & 26.444                   & 0.774                   & 0.876                   & 168.717                  & 36.690                   & 0.949                   & 0.968                   & 55.841                  & 2069.351          \\
			PSTNN     & 20.807                   & 0.561                   & 0.784                   & 292.246                  & 32.978                   & 0.916                   & 0.952                   & 75.418                   & 39.795                   & 0.974                   & 0.986                   & 35.032                  & 260.059           \\
			SMFLRTC   & 35.743                   & 0.959                   & 0.975                   & 54.765                   & 39.430                   & 0.978                   & 0.986                   & 36.030                   & 42.505                   & 0.988                   & 0.992                   & 25.403                  & 1082.246          \\
			KBRTC     & 28.898                   & 0.863                   & 0.924                   & 127.082                  & 38.852                   & 0.973                   & 0.984                   & 43.495                   & 44.365                   & 0.991                   & 0.995                   & 22.041                  & 1080.037          \\
			ESPTC     & 28.516                   & 0.828                   & 0.904                   & 132.219                  & 34.232                   & 0.924                   & 0.955                   & 70.417                   & 38.884                   & 0.968                   & 0.982                   & 41.825                  & 1792.766          \\
			FTNN      & 26.691                   & 0.777                   & 0.877                   & 157.785                  & 30.658                   & 0.877                   & 0.927                   & 101.872                  & 35.174                   & 0.942                   & 0.964                   & 62.267                  & 1749.613          \\
			WSTNN     & 37.083                   & 0.968                   & 0.981                   & 50.027                   & 42.199                   & 0.987                   & 0.992                   & 28.408                   & 47.582                   & 0.995                   & 0.997                   & 15.496                  & 1575.438          \\
			NWSTNN    & 39.234                   & 0.970                   & 0.984                   & 37.972                   & 44.997                   & 0.989                   & 0.994                   & 20.735                   & 51.224                   & 0.997                   & 0.998                   & 10.308                  & 2885.440          \\
			FFMTC     & \textit{\textbf{41.317}} & \textit{\textbf{0.982}} & \textit{\textbf{0.990}} & \textit{\textbf{29.417}} & \textit{\textbf{46.388}} & \textit{\textbf{0.993}} & \textit{\textbf{0.996}} & \textit{\textbf{17.143}} & \textit{\textbf{51.980}} & \textit{\textbf{0.997}} & \textit{\textbf{0.999}} & \textit{\textbf{8.951}} & \textbf{3045.806} \\ \hline
		\end{tabular}%
	}\label{HSVTC1}
\end{table*}

\subsection{Tensor robust principal component analysis}
In this section, we evaluate the performance of the proposed FFM-based TRPCA method by HSI denoising. The compared TRPCA methods include the SNN \cite{2252014253}, TNN \cite{8606166}, 3DTNN and 3DLogTNN \cite{7342019749}.
\subsubsection{HSI denoising}
\begin{figure*}[!h] 
	\centering  
	\vspace{0cm} 
	\subfloat[]{
		\begin{minipage}[b]{0.135\linewidth}
		\includegraphics[width=1\linewidth]{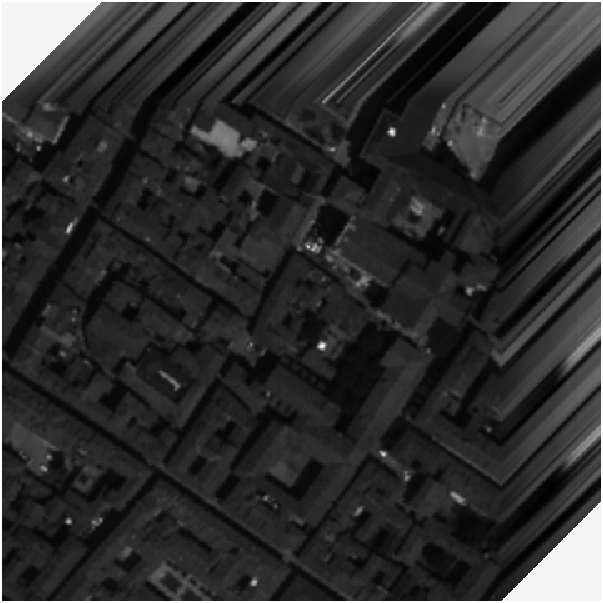}
		\includegraphics[width=1\linewidth]{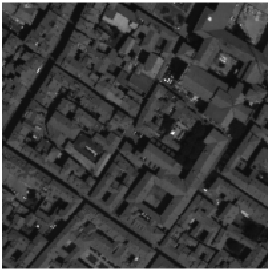}
		\includegraphics[width=1\linewidth]{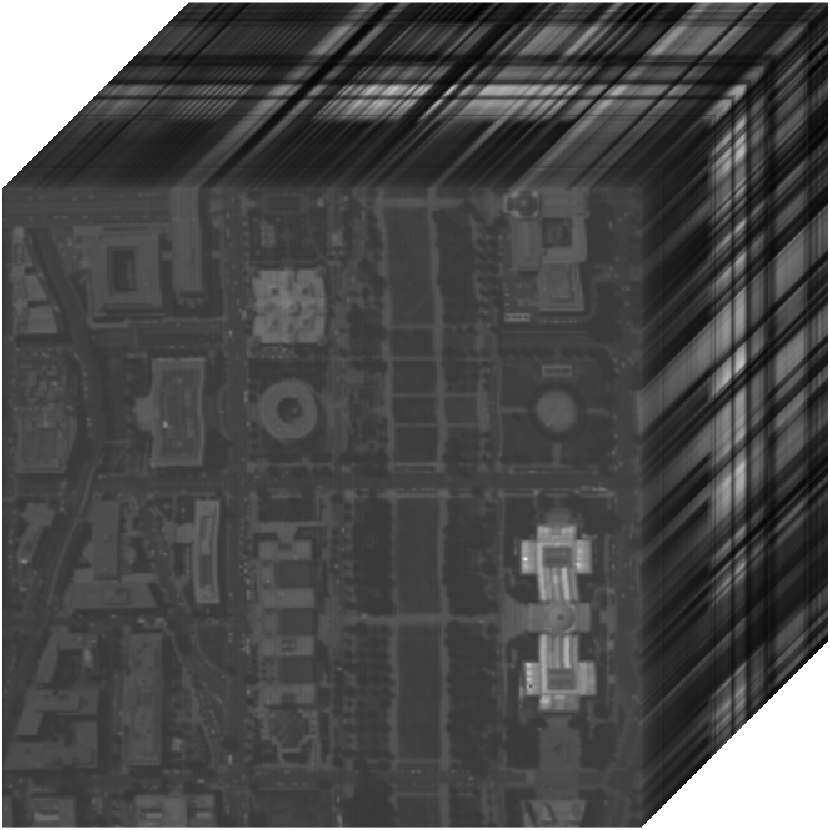}
		\includegraphics[width=1\linewidth]{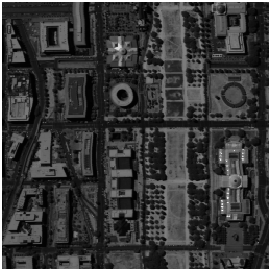}
		\includegraphics[width=1\linewidth]{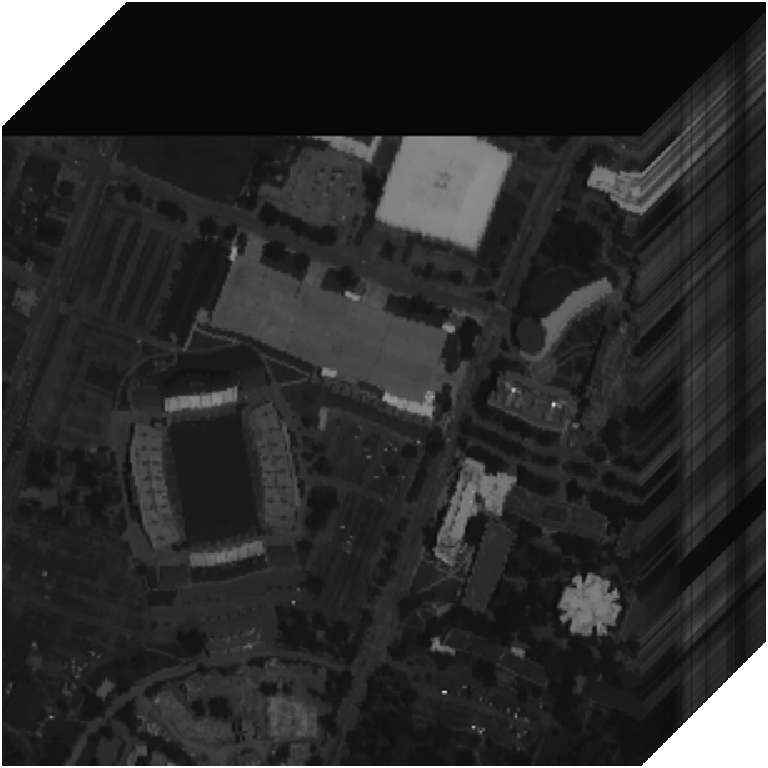}
		\includegraphics[width=1\linewidth]{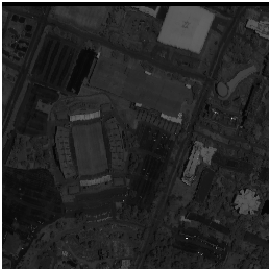}
		\end{minipage}
}
\subfloat[]{
	\begin{minipage}[b]{0.135\linewidth}
		\includegraphics[width=1\linewidth]{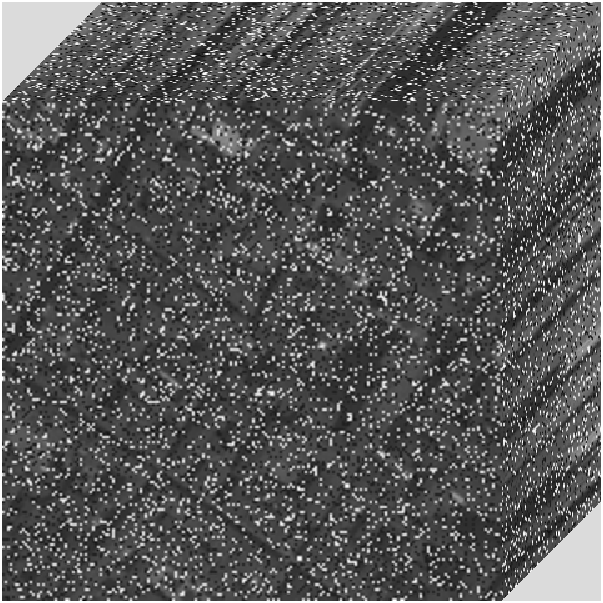}
		\includegraphics[width=1\linewidth]{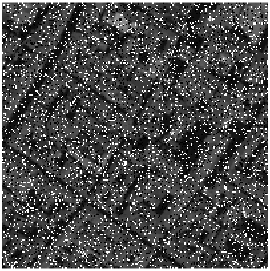}
		\includegraphics[width=1\linewidth]{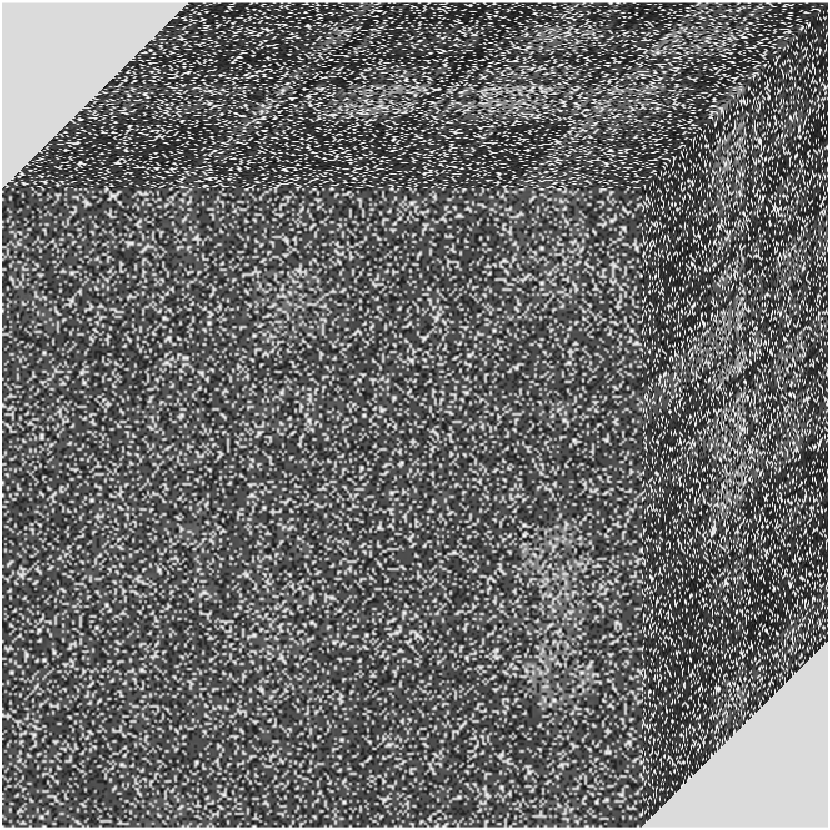}
		\includegraphics[width=1\linewidth]{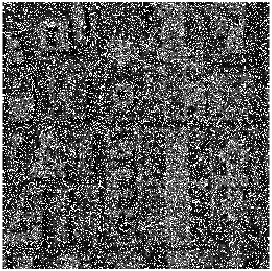}
		\includegraphics[width=1\linewidth]{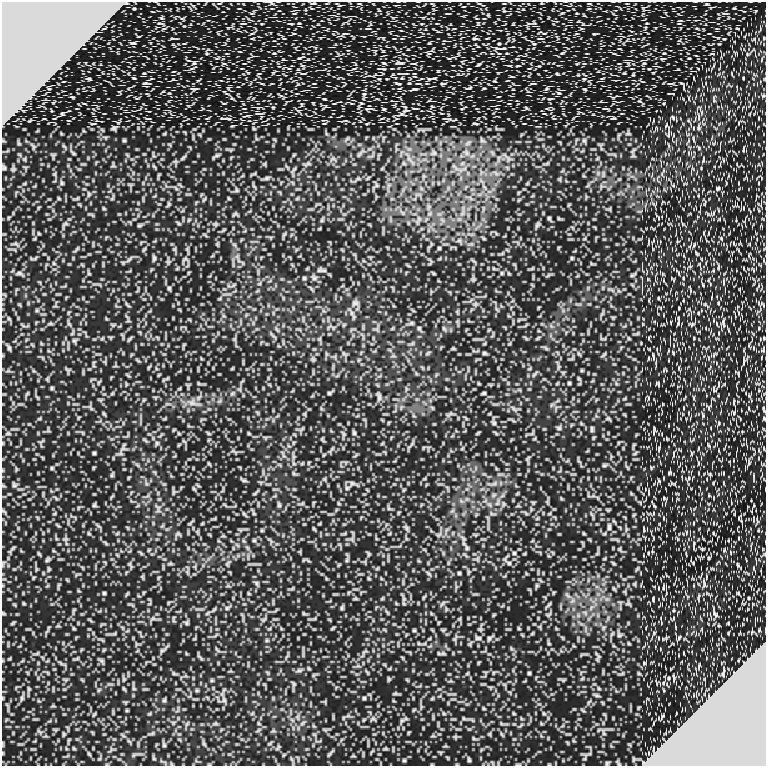}
		\includegraphics[width=1\linewidth]{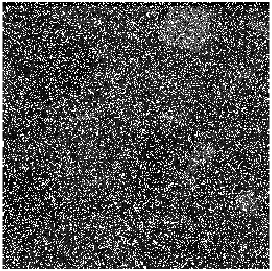}
\end{minipage}}
\subfloat[]{
	\begin{minipage}[b]{0.135\linewidth}
		\includegraphics[width=1\linewidth]{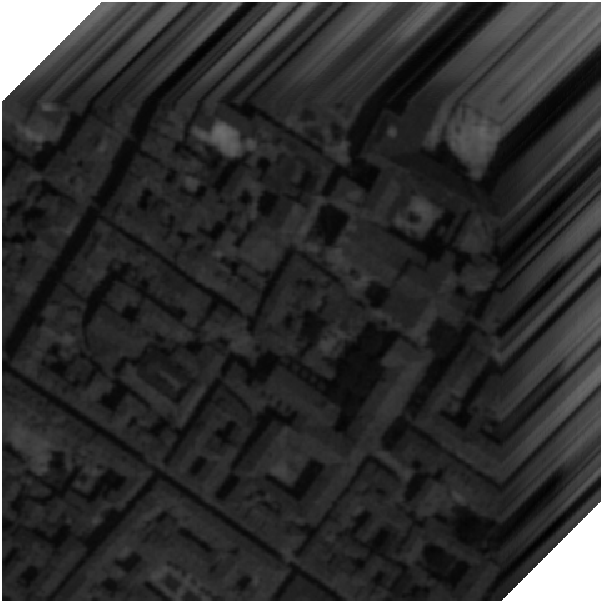}
		\includegraphics[width=1\linewidth]{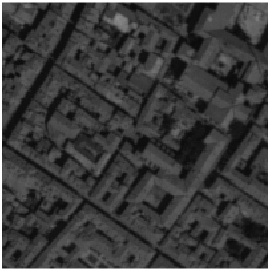}
		\includegraphics[width=1\linewidth]{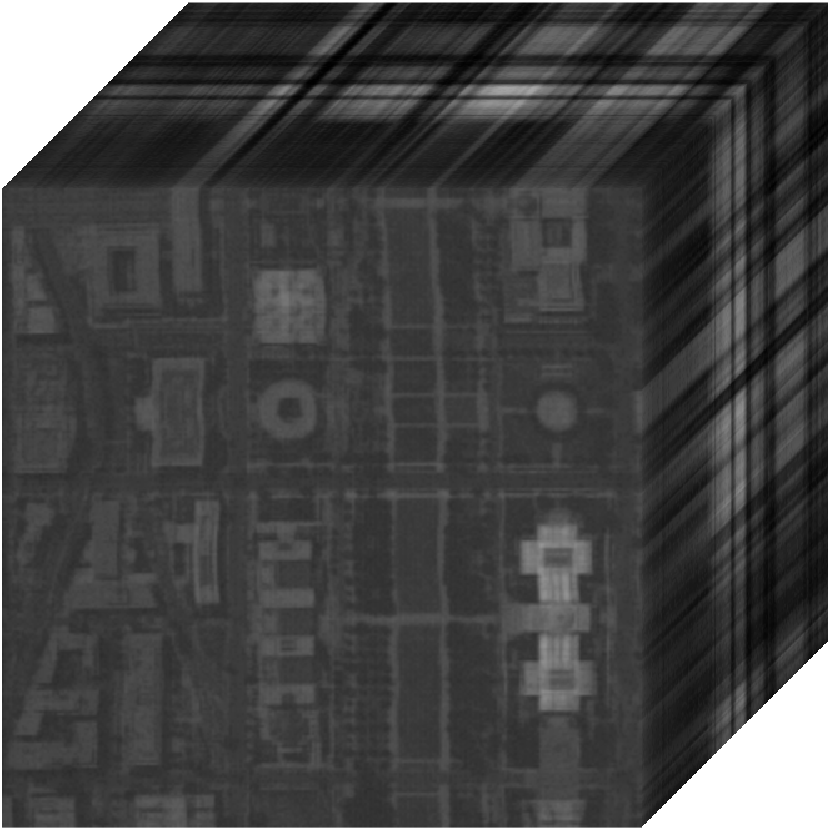}
		\includegraphics[width=1\linewidth]{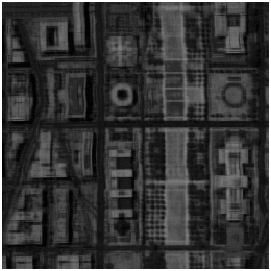}
		\includegraphics[width=1\linewidth]{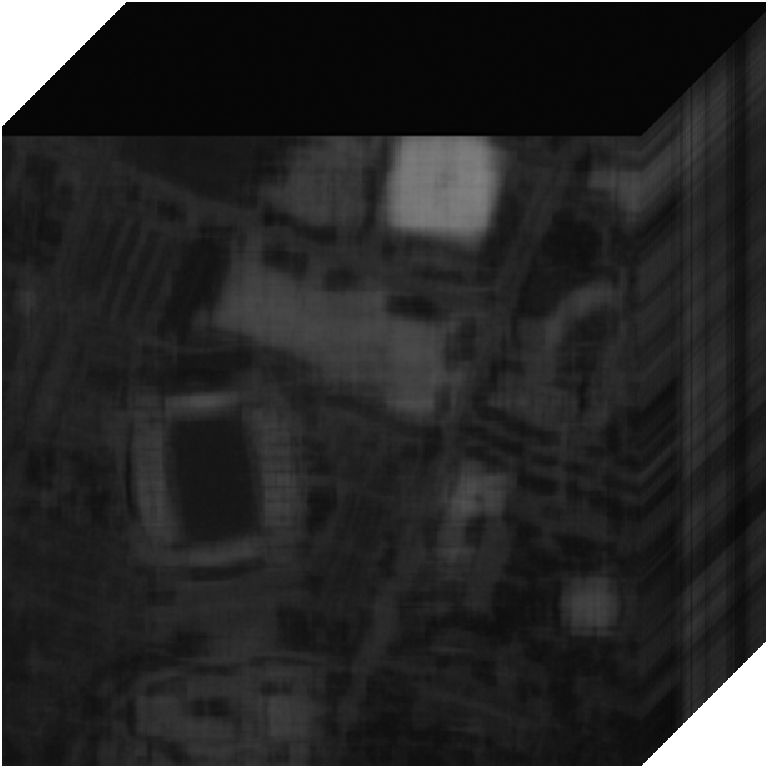}
		\includegraphics[width=1\linewidth]{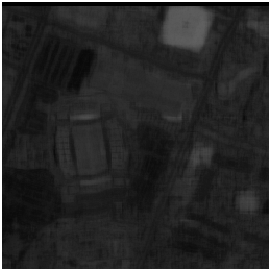}
	\end{minipage}}
\subfloat[]{
	\begin{minipage}[b]{0.135\linewidth}
		\includegraphics[width=1\linewidth]{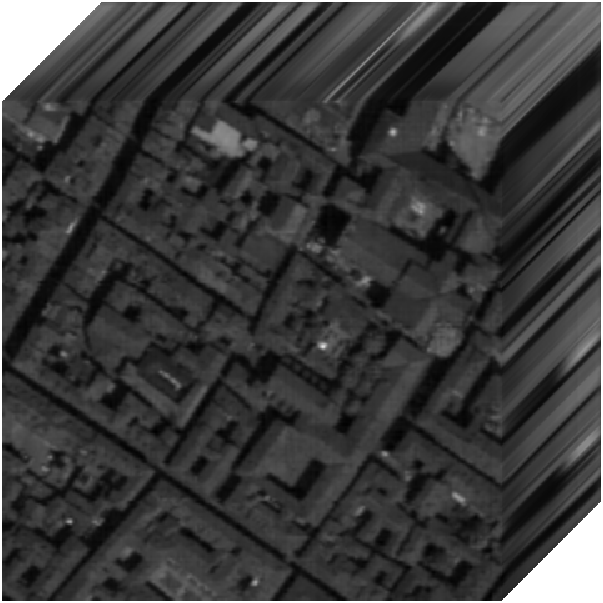}
		\includegraphics[width=1\linewidth]{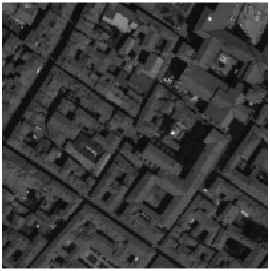}
		\includegraphics[width=1\linewidth]{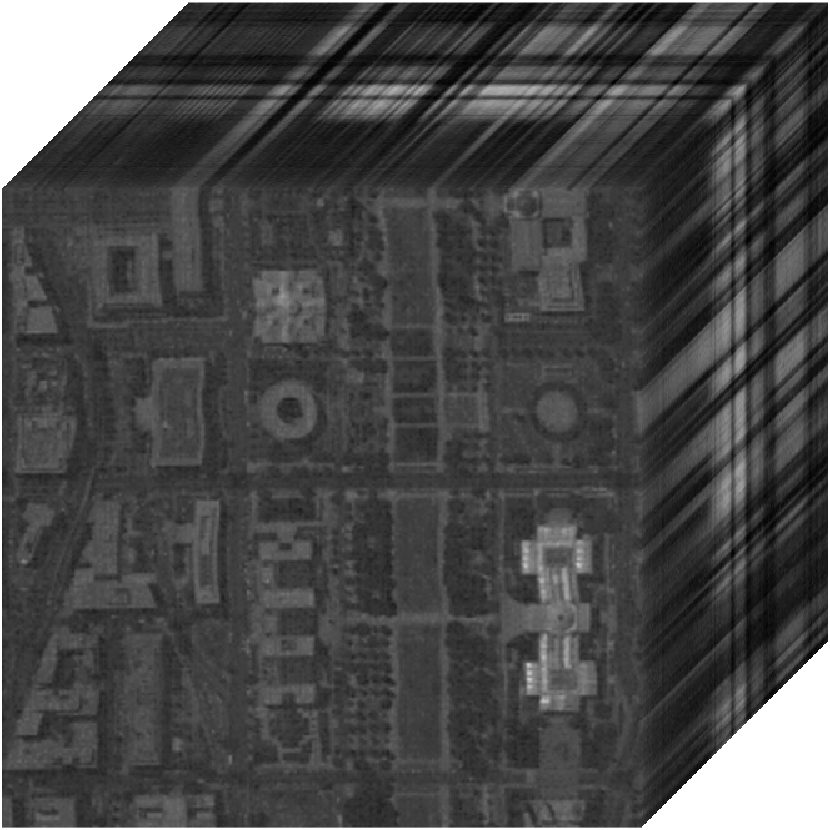}
		\includegraphics[width=1\linewidth]{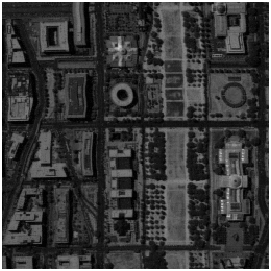}
		\includegraphics[width=1\linewidth]{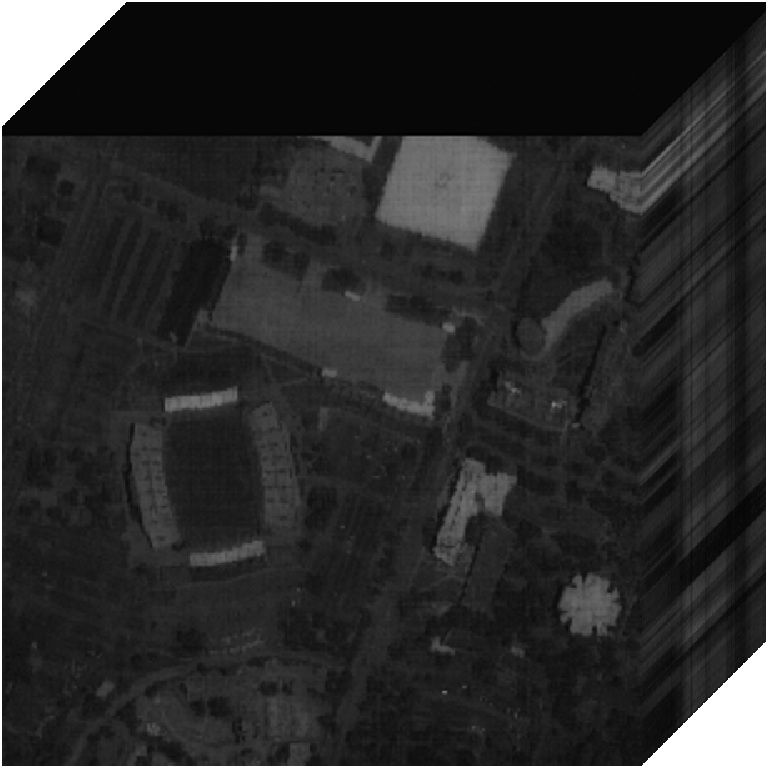}
		\includegraphics[width=1\linewidth]{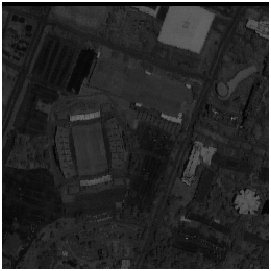}
\end{minipage}}
\subfloat[]{
	\begin{minipage}[b]{0.135\linewidth}
		\includegraphics[width=1\linewidth]{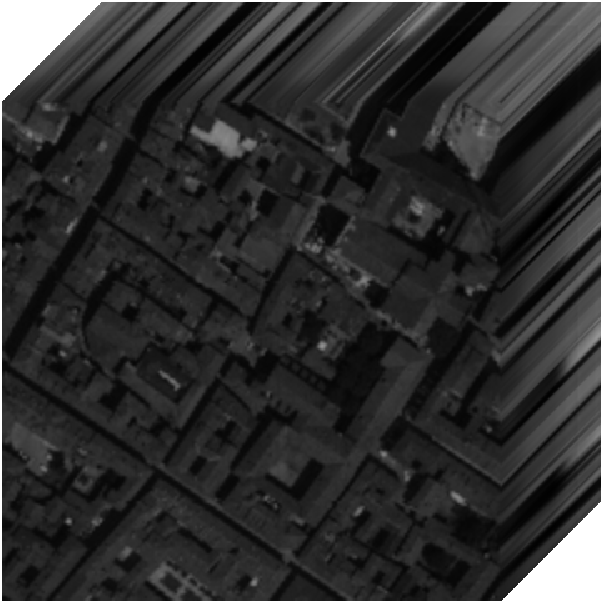}
		\includegraphics[width=1\linewidth]{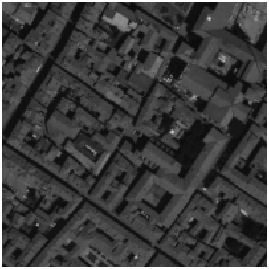}
		\includegraphics[width=1\linewidth]{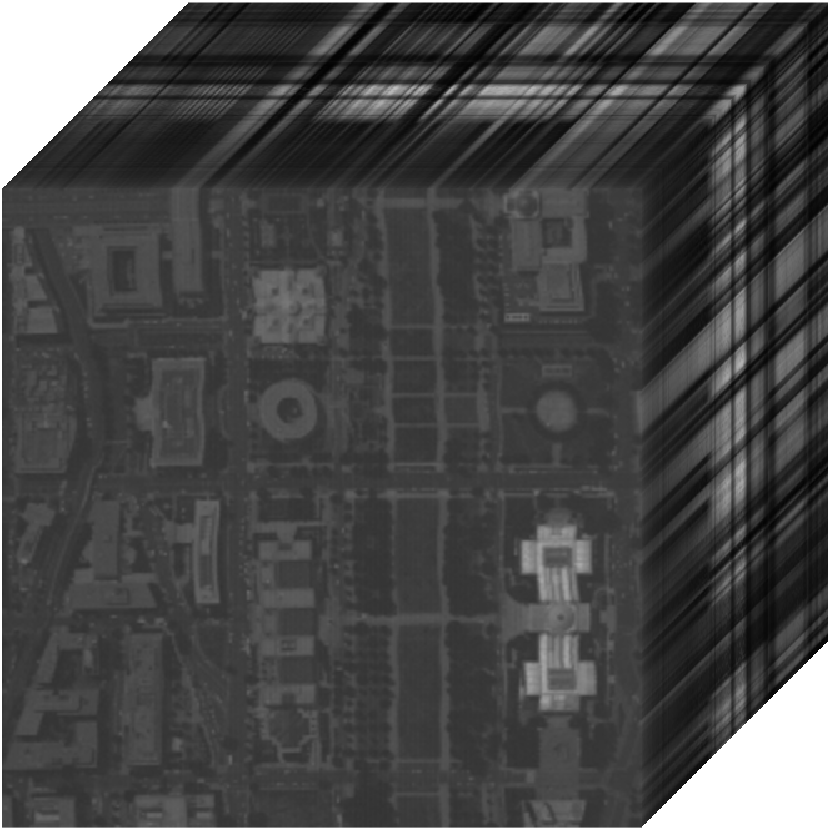}
		\includegraphics[width=1\linewidth]{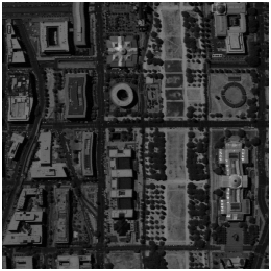}
		\includegraphics[width=1\linewidth]{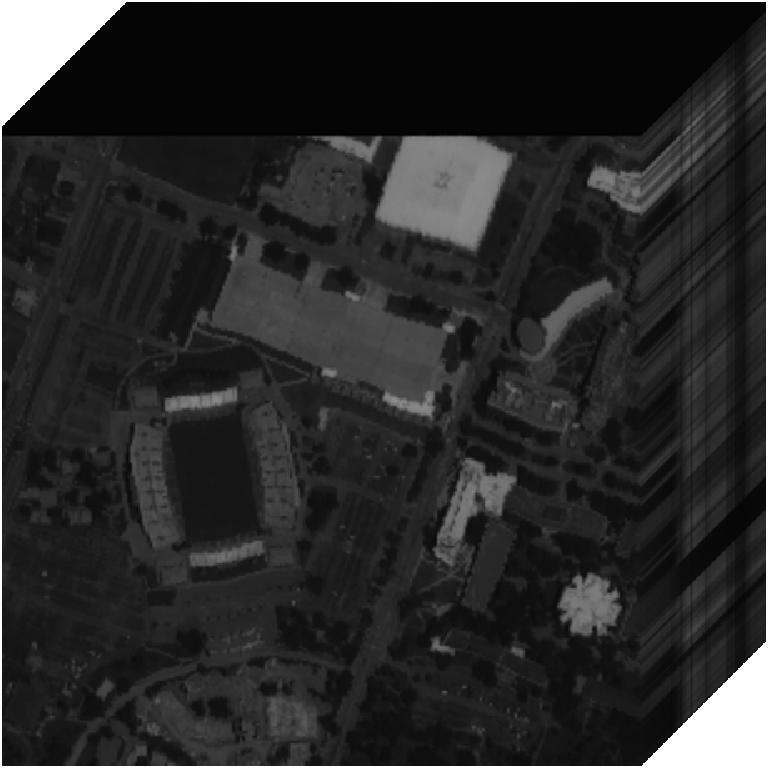}
		\includegraphics[width=1\linewidth]{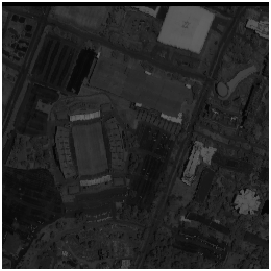}
\end{minipage}}
\subfloat[]{
	\begin{minipage}[b]{0.135\linewidth}
		\includegraphics[width=1\linewidth]{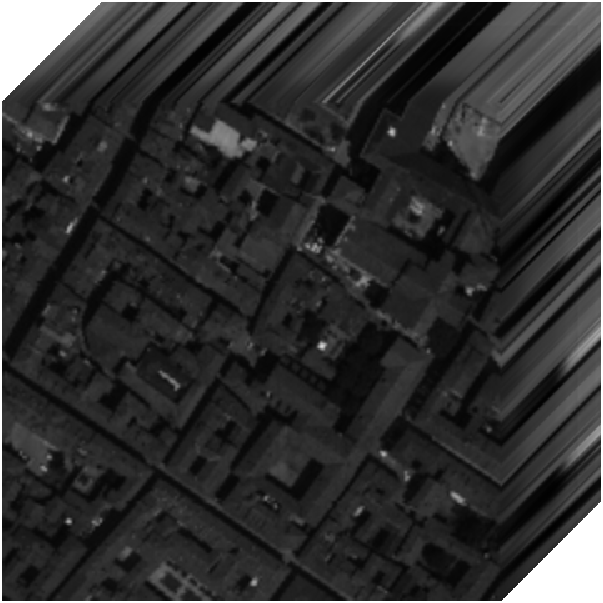}
		\includegraphics[width=1\linewidth]{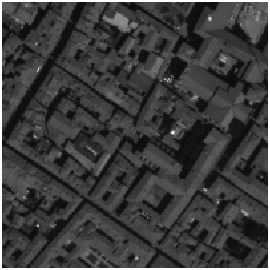}
		\includegraphics[width=1\linewidth]{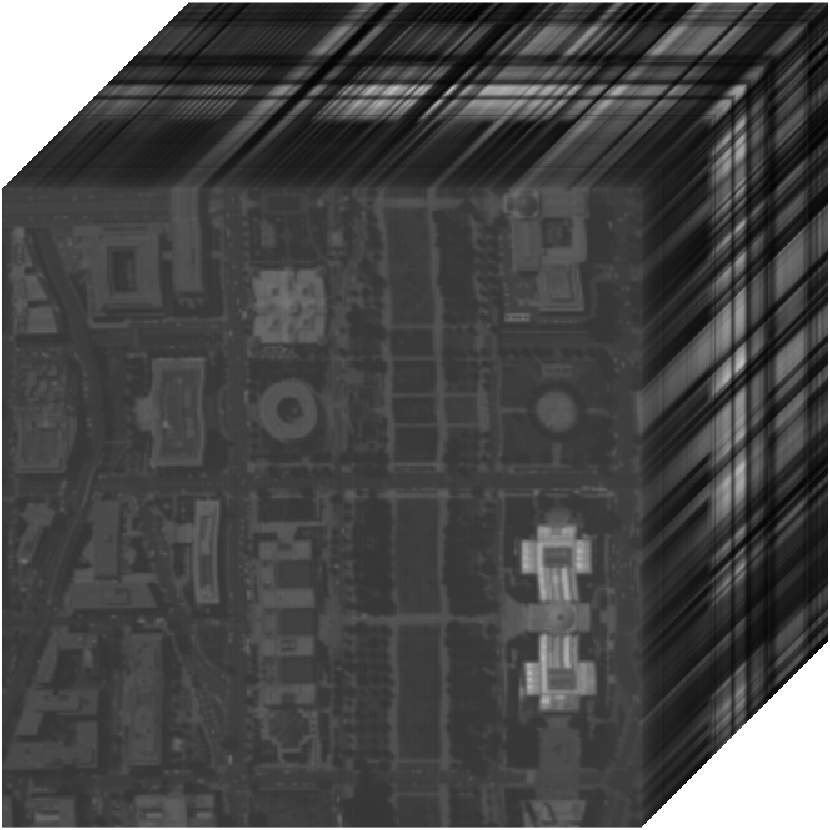}
		\includegraphics[width=1\linewidth]{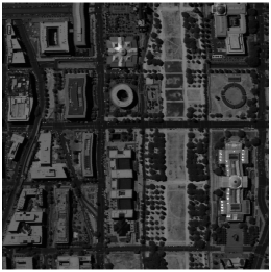}
		\includegraphics[width=1\linewidth]{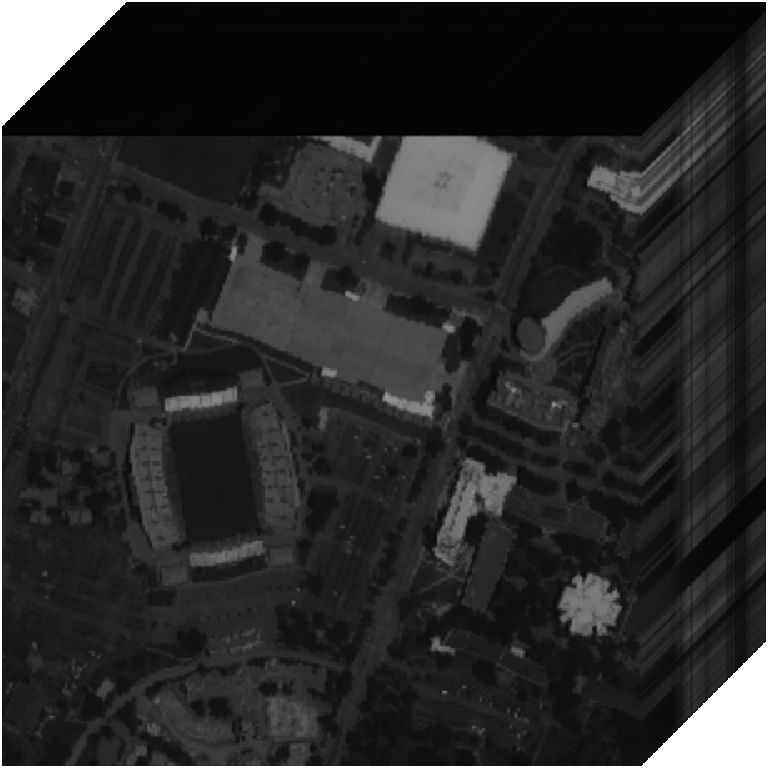}
		\includegraphics[width=1\linewidth]{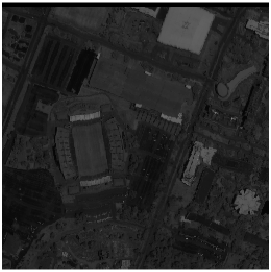}
\end{minipage}}
\subfloat[]{
	\begin{minipage}[b]{0.135\linewidth}
		\includegraphics[width=1\linewidth]{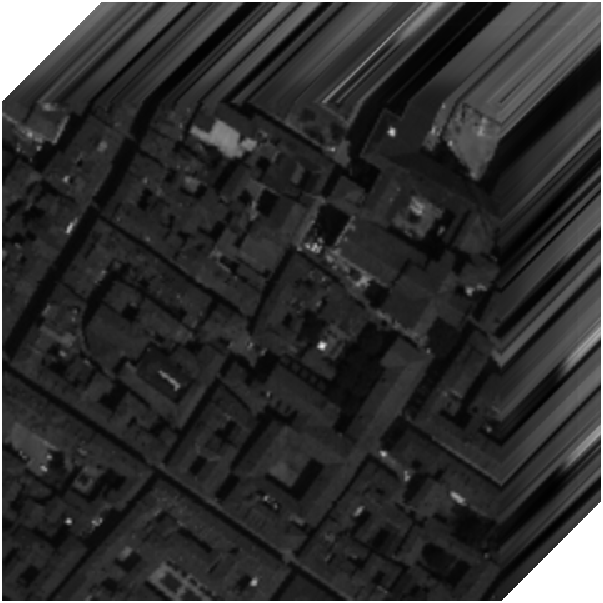}
		\includegraphics[width=1\linewidth]{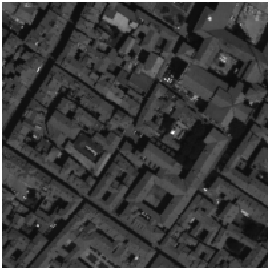}
		\includegraphics[width=1\linewidth]{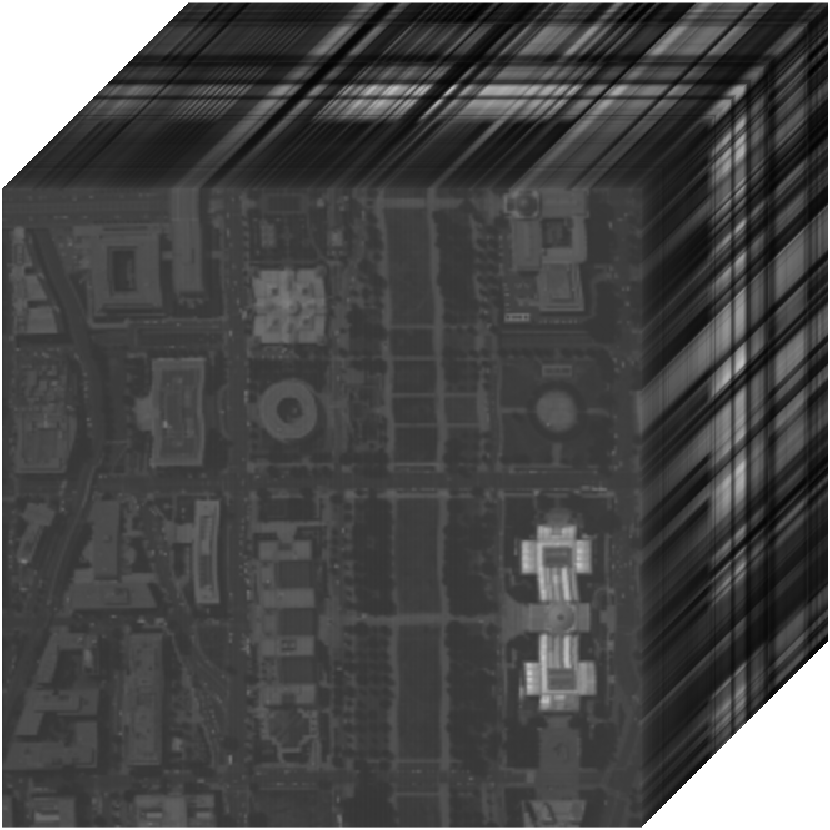}
		\includegraphics[width=1\linewidth]{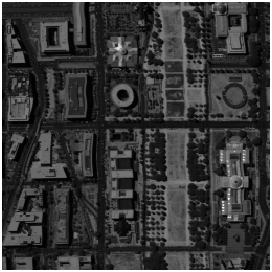}
		\includegraphics[width=1\linewidth]{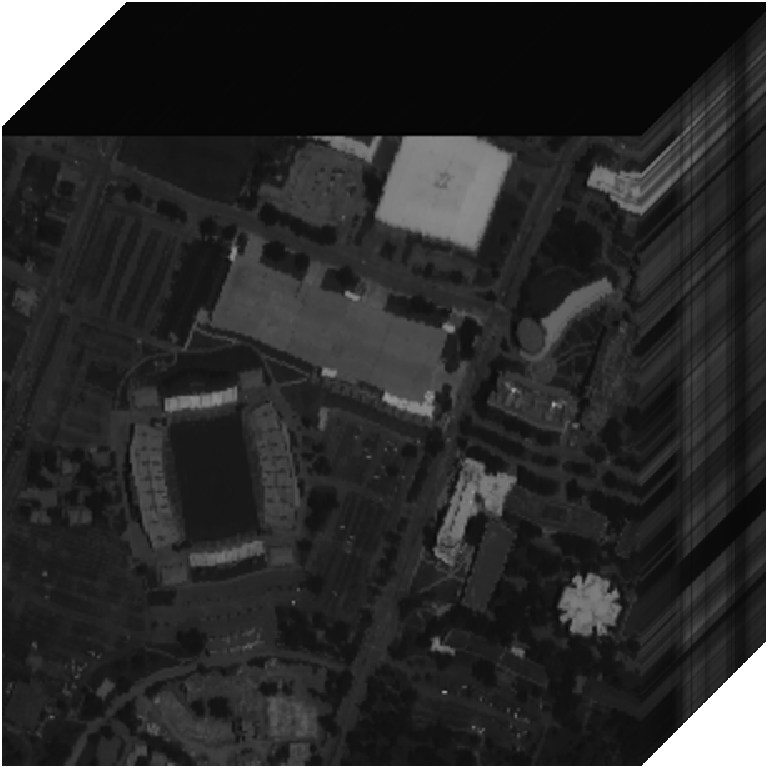}
		\includegraphics[width=1\linewidth]{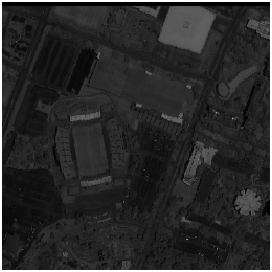}
\end{minipage}}
	\caption{(a) Original image. (b) Noise image. (c) SNN. (d) TNN. (e) 3DTNN. (f) 3DLogTNN. (g) FFMTRPCA. NL: top 2 rows is 0.2, bottom 4 rows is 0.4. HSIs are as follows: Pavia, Washington, Houston. The bands are in order: 30 band of Pavia, 110 band of Washington, 40 band of Houston.}
	\label{HSITRPCA}
\end{figure*}

We test the Pavia University and Washington DC Mall and Houston HSI data sets, where Pavia data size is $200\times200\times80$ and Washington DC Mall data size is $256\times256\times150$ and Houston data size is $256\times256\times100$. The random salt-pepper noise level (NL) is set to 0.2 and 0.4. In Fig.\ref{HSITRPCA}, we show 3-D visualization of the denoising results and one band in these three HSIs. The results demonstrate that among the five comparative methods, the FFM-based TRPCA method achieves the best visual effects in terms of denoising and detail protection. Table \ref{HSITRPCA1} lists the PSNR, SSIM, and FSIM values of three tested HSIs recovered by different methods. It is not difficult to see that the PSNR of our method is at least 1dB higher than that of the suboptimal method in any case. In particular, when NL is 0.2, the SSIM and FSIM obtained by our method are almost perfect.
\begin{table*}[]
	\caption{The PSNR, SSIM, and FSIM values output by observed and the five utilized TRPCA methods for HSI.}
	\resizebox{\textwidth}{!}{
	\centering
	\begin{tabular}{ccccccccc}
		\hline
		HSI & NL       & \multicolumn{3}{c}{0.2}   & \multicolumn{3}{c}{0.4}   & Time(s)  \\
		& Method   & PSNR    & SSIM   & FSIM   & PSNR    & SSIM   & FSIM   &          \\ \hline
		& Observed & 11.8104 & 0.1254 & 0.5656 & 8.7975  & 0.0480 & 0.4223 & 0.0000   \\
		& SNN      & 30.7029 & 0.9324 & 0.9501 & 27.5276 & 0.8406 & 0.8878 & 18.3582  \\
		& TNN      & 46.2184 & 0.9895 & 0.9923 & 38.4297 & 0.9744 & 0.9830 & 45.8733  \\
		& 3DTNN    & 46.2144 & 0.9986 & 0.9989 & 40.8729 & 0.9954 & 0.9961 & 65.3882  \\
		& 3DLogTNN & 57.4007 & 0.9998 & 0.9998 & 44.6775 & 0.9973 & 0.9980 & 108.9973 \\
		\multirow{-6}{*}{Pavia City Center} &
		FFMTRPCA &
		\textbf{60.3023} & \textbf{0.9999} & \textbf{0.9999} & \textbf{45.9924} & \textbf{0.9983} & \textbf{0.9986} & \textbf{166.6786}\\ \hline
		& Observed & 11.4303 & 0.1222 & 0.5527 & 8.4179  & 0.0478 & 0.4117 & 0.0000   \\
		& SNN      & 31.4787 & 0.9279 & 0.9509 & 28.2191 & 0.8489 & 0.9022 & 41.6687  \\
		& TNN      & 43.8989 & 0.9922 & 0.9943 & 35.8265 & 0.9533 & 0.9737 & 104.9230 \\
		& 3DTNN    & 50.7842 & 0.9996 & 0.9996 & 42.2630 & 0.9933 & 0.9952 & 214.2376 \\
		& 3DLogTNN & 51.3655 & 0.9995 & 0.9996 & 47.5463 & 0.9986 & 0.9989 & 269.4383 \\
		\multirow{-6}{*}{Washington DC} &
		FFMTRPCA &
		\textbf{54.1785} & \textbf{0.9998} & \textbf{0.9998} & \textbf{49.9167} & \textbf{0.9994} & \textbf{0.9994} & \textbf{368.1501} \\ \hline
		& Observed & 10.9201 & 0.0412 & 0.4359 & 7.9205  & 0.0136 & 0.2824 & 0.0000   \\
		& SNN      & 34.7662 & 0.9114 & 0.9312 & 32.1612 & 0.8531 & 0.8866 & 35.3139  \\
		& TNN      & 46.5744 & 0.9915 & 0.9943 & 40.4294 & 0.9639 & 0.9815 & 100.3306 \\
		& 3DTNN    & 49.0261 & 0.9980 & 0.9988 & 46.7518 & 0.9963 & 0.9973 & 181.8898 \\
		& 3DLogTNN & 57.2123 & 0.9994 & 0.9996 & 48.8178 & 0.9972 & 0.9981 & 246.6725 \\
		\multirow{-6}{*}{Houston} &
		FFMTRPCA &
		\textbf{59.5381} & \textbf{0.9994} & \textbf{0.9996} & \textbf{51.2677} & \textbf{0.9981} & \textbf{0.9986} & \textbf{414.5798} \\ \hline
\end{tabular}}\label{HSITRPCA1}
\end{table*}
\subsection{Convergency Behaviours}
\begin{figure}
	\subfloat[TC]{
			\includegraphics[width=0.45\linewidth]{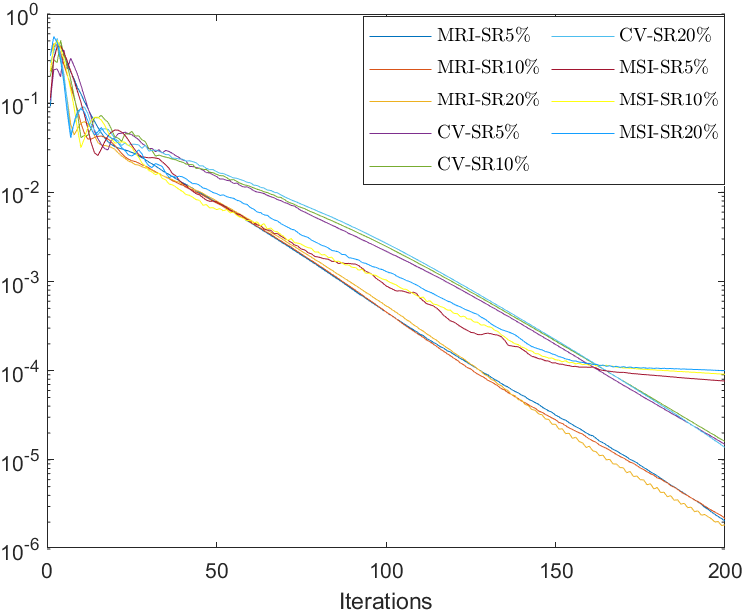}
}
	\subfloat[TRPCA]{
	\includegraphics[width=0.45\linewidth]{images/TCconvergence}	
}
	\caption{The convergence behaviours of Algorithm 2, with respect to different sampling rates. The convergence behaviours of Algorithm 3, with respect to different noise level.}
	\label{convergence}
\end{figure}

We take the completeness of MRI, CV, MSI data and the robust principal component analysis of HSI as examples to illustrate the convergence behavior of the algorithm under different sampling rates and different noise levels. We have drawn $\|\mathcal{X}^{k+1}-\mathcal{X}^{k}\|_{\infty}$ and $\|\mathcal{L}^{k+1}-\mathcal{L}^{k}\|_{\infty}$ for each iteration in the figure \ref{convergence}. It can be seen that our algorithm converges stably, and the convergence speed is also quite fast.
\section{Conclusion}
In this paper, a new tensor sparse measure is proposed, and FFM-based LRTC model and FFM-based TRPCA model are presented for LRTC and TRPCA problems. This sparse measure is more comprehensive in describing the features of tensors, which is a basic guarantee for our method to get ideal results. A large number of real data experiments verify that the proposed method is quite effective and highly efficient. Compared with other state-of-the-art methods, our method is optimal in the experiment of multiple data. How to construct a more appropriate tensor sparse measure is still the direction we need to think about in the future.


%





\ifCLASSOPTIONcaptionsoff
  \newpage
\fi



%
%
%
\bibliography{bibtex/Fm}
%








\end{document}